\pdfoutput=1 

\documentclass[acmsmall]{acmart}

\AtBeginDocument{%
  \providecommand\BibTeX{{%
    \normalfont B\kern-0.5em{\scshape i\kern-0.25em b}\kern-0.8em\TeX}}}

\usepackage{soul}

\usepackage{amssymb}
\usepackage{algorithm}
\usepackage{algorithmic}
\usepackage{multirow}

\def\ie{\textit{i.e.}}
\def\eg{\textit{e.g.}}

\begin{document}

\title{Smart Scribbles for Image Matting}


\author{Xin Yang}
\authornote{Both authors contributed equally to this research.}
\email{xinyang@dlut.edu.cn}
\affiliation{%
	\institution{Dalian University of Technologyy}
	\streetaddress{2 linggong road}
	\city{Dalian}
	\state{Liaoning}
	\postcode{116024}
}
\affiliation{%
	\institution{Beijing Technology and Business University}
	\streetaddress{11 Fucheng Road}
	\city{Beijing} 
	\country{China}
	\postcode{100048}
}

\author{Yu Qiao}
\authornotemark[1]
\email{qiaoyu2017@mail.dlut.edu.cn}
\author{Shaozhe Chen}
\email{csz@mail.dlut.edu.cn}
\affiliation{%
	\institution{Dalian University of Technologyy}
	\streetaddress{2 linggong road}
	\city{Dalian}
	\state{Liaoning}
	\country{China}
	\postcode{116024}
}

\author{Shengfeng He}
\authornote{Corresponding authors.}
\affiliation{%
	\institution{South China University of Technology}
	\city{Guangzhou}
	\state{Guangdong}
	\country{China}
}
\email{hesfe@scut.edu.cn}

\author{Baocai Yin}
\email{ybc@dlut.edu.cn}
\affiliation{%
	\institution{Dalian University of Technology}
	\streetaddress{2 linggong road}
	\city{Dalian}
	\state{Liaoning}
	\postcode{116024}
}
\affiliation{%
	\institution{Peng Cheng Laboratory}
	\streetaddress{2 linggong road}
	\city{Shenzhen}
	\state{Guangdong}
	\country{China}
	\postcode{518055}
}

\author{Qiang Zhang}
\email{zhangq@dlut.edu.cn}
\affiliation{%
	\institution{Dalian University of Technology}
	\streetaddress{2 linggong road}
	\city{Dalian}
	\state{Liaoning}
	\country{China}
	\postcode{116024}
}

\author{Xiaopeng Wei}
\authornotemark[2]
\email{weixp@dlut.edu.cn}
\affiliation{%
	\institution{Dalian University of Technology}
	\streetaddress{2 linggong road}
	\city{Dalian}
	\state{Liaoning}
	\country{China}
	\postcode{116024}
}

\author{Rynson W.H. Lau}
\affiliation{%
	\institution{City University of Hong Kong}
	\city{Hong Kong}
	\country{China}
}
\email{rynson.lau@cityu.edu.hk}

\renewcommand{\shortauthors}{Xin, et al.}

\begin{abstract}
  Image matting is an ill-posed problem that usually requires additional user input, such as trimaps or scribbles. Drawing a fine trimap requires a large amount of user effort, while using scribbles can hardly obtain satisfactory alpha mattes for non-professional users. Some recent deep learning based matting networks rely on large-scale composite datasets for training to improve performance, resulting in the occasional appearance of obvious artifacts when processing natural images.
  In this paper, we explore the intrinsic relationship between user input and alpha mattes, and strike a balance between user effort and the quality of alpha mattes. In particular, we propose an interactive framework, referred to as smart scribbles, to guide users to draw few scribbles on the input images to produce high-quality alpha mattes. It first infers the most informative regions of an image for drawing scribbles to indicate different categories (foreground, background or unknown), then spreads these scribbles (\ie, the category labels) to the rest of the image via our well-designed two-phase propagation. Both neighboring low-level affinities and high-level semantic features are considered during the propagation process. Our method can be optimized without large-scale matting datasets, and exhibits more universality in real situations. Extensive experiments demonstrate that smart scribbles can produce more accurate alpha mattes with reduced additional input, compared to the state-of-the-art matting methods.
\end{abstract}



\maketitle

\section{Introduction}
\label{sec:intro}

\begin{figure*}[htp]
	\setlength{\tabcolsep}{1pt}\small{
		\begin{tabular}{ccccc}
			\includegraphics[scale=0.134]{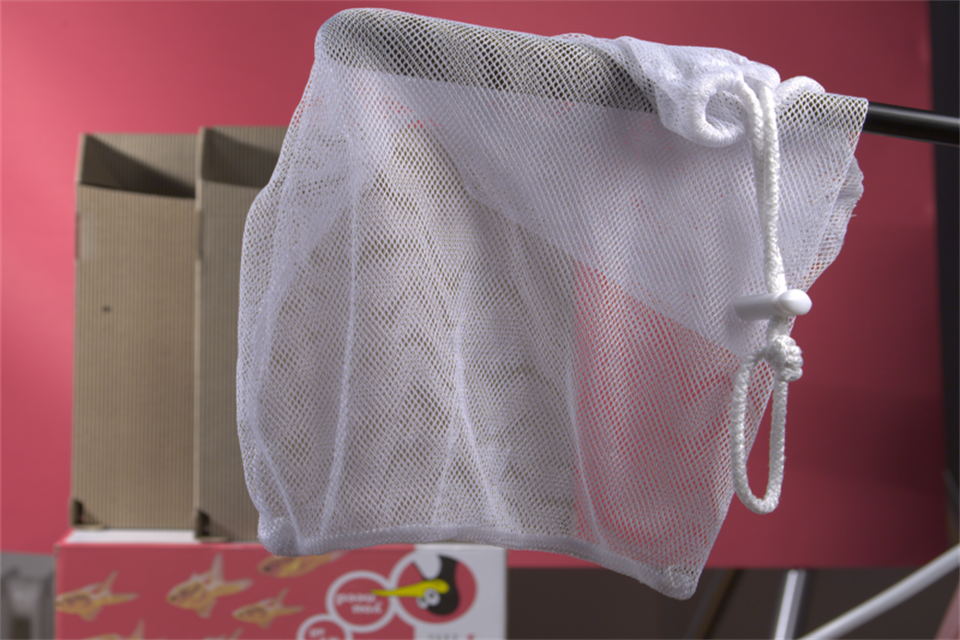} &
			\includegraphics[scale=0.1]{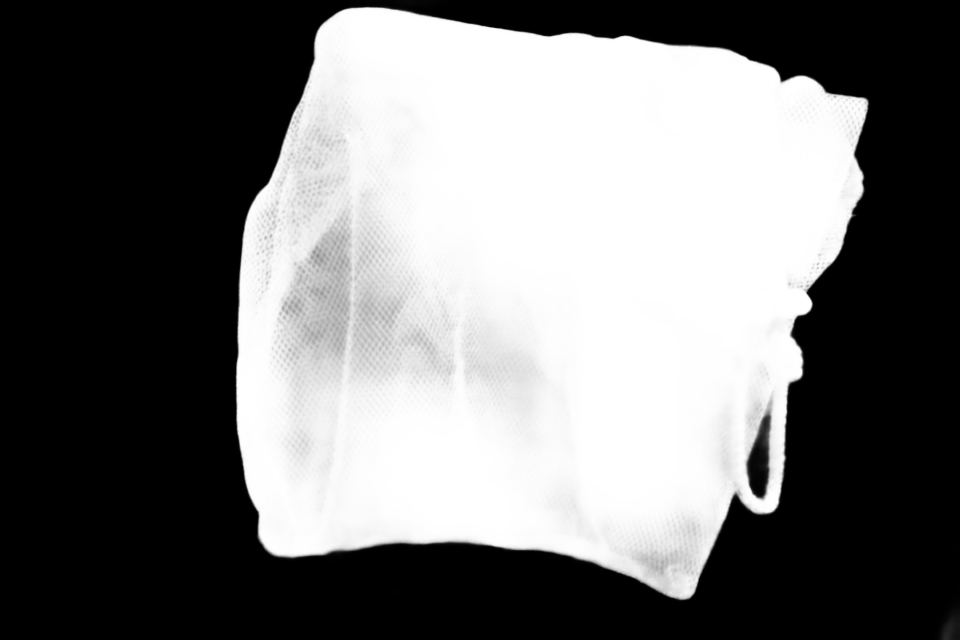} &
			\includegraphics[scale=0.134]{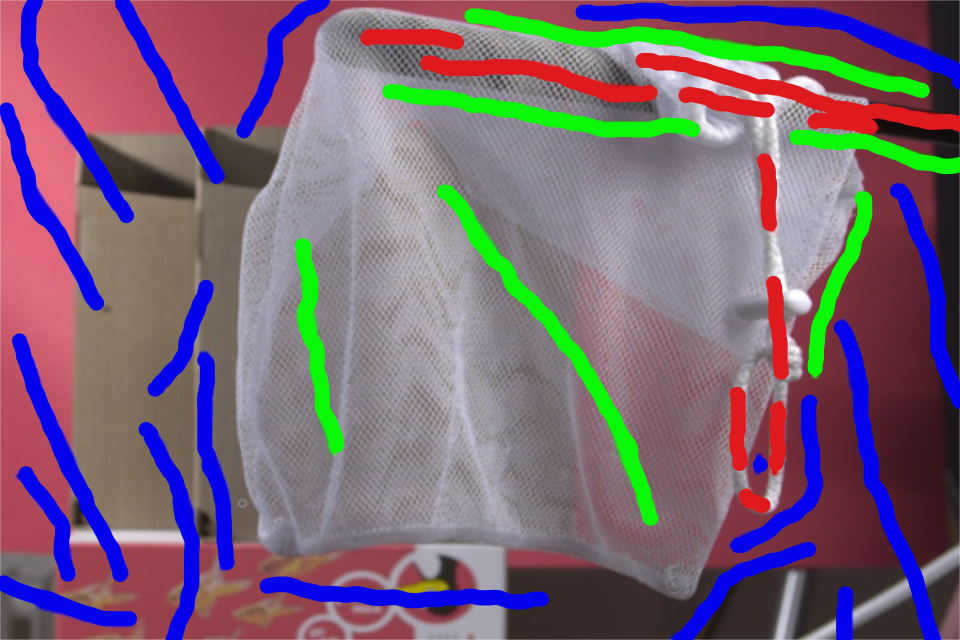} &
			\includegraphics[scale=0.134]{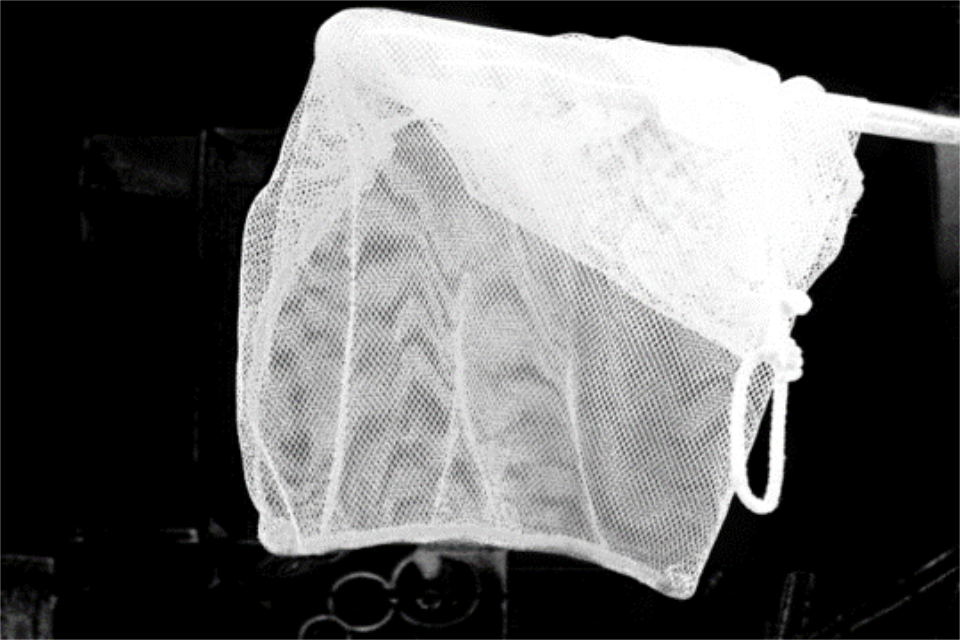} \\
			
			(a) Input Image & (b) Late Fusion~\cite{Zhang_2019_CVPR} & (c) Traditional Scribbles & (d) Matte from (c)  \\
			
			\includegraphics[scale=0.134]{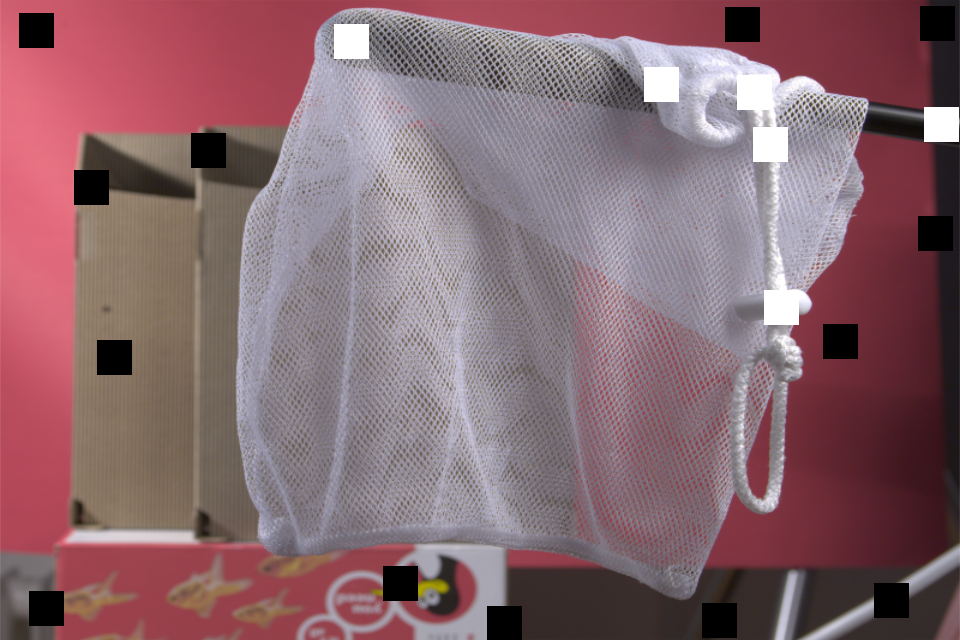} &
			\includegraphics[scale=0.134]{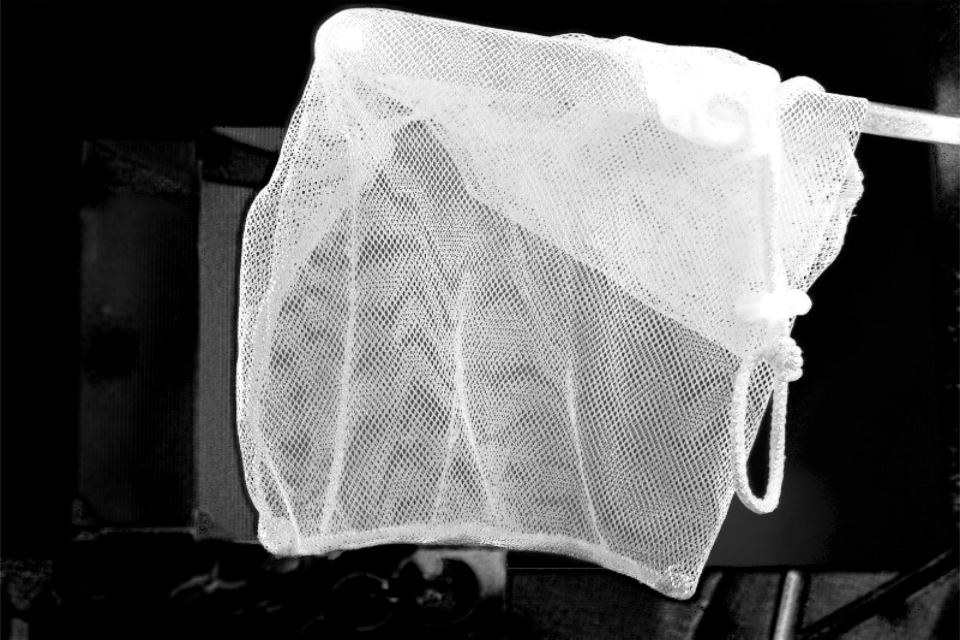} &
			\includegraphics[scale=0.134]{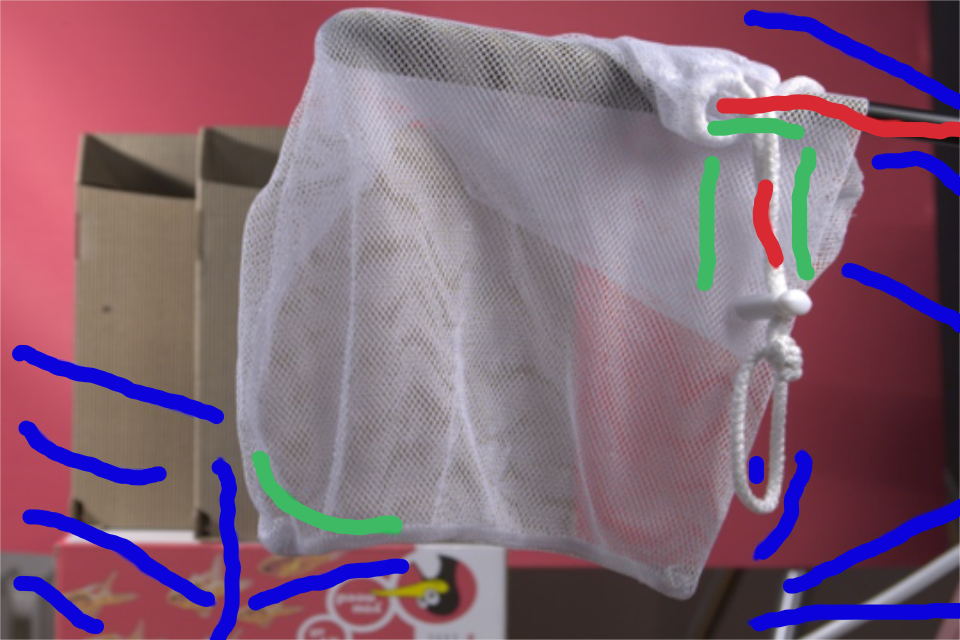} &
			\includegraphics[scale=0.134]{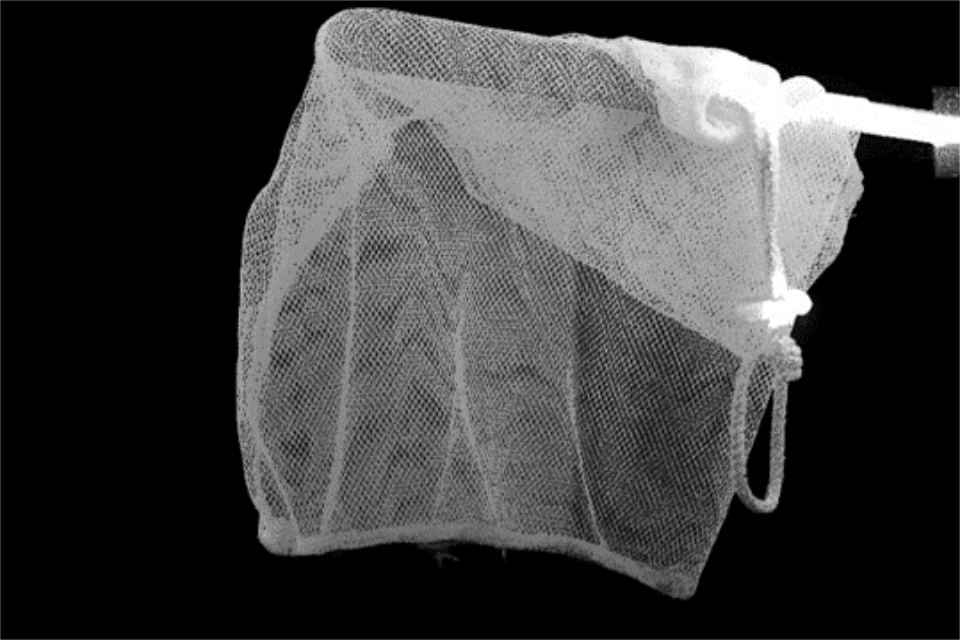} \\
			
			(e) Active Matting~\cite{NIPS2018_7710} & (f) Matte from (e) & (g) Smart Scribbles & (h) Matte from (g) \\
	\end{tabular}}
	\caption{The comparison of \emph{smart scribbles} with the Late Fusion~\cite{Zhang_2019_CVPR}, traditional scribbles and Active Matting~\cite{NIPS2018_7710}. The proposed method is able to produce more accurate alpha mattes with refined texture details and clear foreground contour, requiring relatively little additional input (both scribbles and interactive boxes are appropriately enlarged for better distinction). }
	\label{fig:teaser}
\end{figure*}

There are different object layers in the image, and we always define the layers of interest as the foreground and the rest as the background. Separating the foreground and background from an image precisely, defined as image matting, is a long-standing problem in both academia and industry. Different from common image segmentation~\cite{Wang2016,Shelhamer2017,Kendall2017}, image matting is required to precisely pull out the foreground object, specific to sophisticated internal texture details and clear boundary contour. It has high value in applications such as movie making and image editing, and is modeled by solving the following under-constrained equation:
\begin{equation}
\label{eq:composition}
I_{p}=\alpha_{p}F_{p}+(1-\alpha_{p})B_{p},
\end{equation}
where $I$ represents the observed image, $p$ refers to a pixel location, $F$ and $B$ refer to the foreground and background layers, respectively. $\alpha$ represents the alpha matte, where $\alpha_p$ varies in the range of [0,1] to indicate the foreground proportion. Equation~(\ref{eq:composition}) is a highly ill-posed problem, as the only known variable is the input $I$. Therefore, additional information from users is essential to solve this problem, which can confine the scope of foreground and background. There are mainly two types of additional inputs: trimaps~\cite{Chuang2003A,Wang2007Optimized,Rhemann2011A,Karacan2015Image,Shahrian2013Improving,Feng2016A,Xu2017Deep} and scribbles~\cite{Levin2007A,Levin2008Spectral,Guan2006Easy,Wang2005An}.

A well-defined trimap is a densely-annotated auxiliary input, and each pixel in it has one of the following three category labels: foreground, background, or unknown. The trimap can effectively constrain the input image according to Equation~(\ref{eq:composition}): foreground and background indicate that $\alpha_p=1$ and $\alpha_p=0$ respectively, and unknown represents the critical areas between foreground and background ($\alpha_p \in (0,1)$). As densely-annotated additional input, trimaps can provide rich information for matting problem, hence most state-of-the-art matting methods~\cite{Aksoy2017Designing,Xu2017Deep} typically use input images and trimaps as input, and they can produce comparable alpha mattes. Nevertheless, generating a perfect trimap means dense annotations for each pixel, which is tedious and time-consuming for common users. The complexity of trimap generation limits their versatility in real-world scenarios. Compared to trimaps, scribbles are more user-friendly additional input. In traditional scribbles-based methods, users are nominally free to draw a few scribbles on the input image to suggest the foreground, background and unknown. The certain regions with scribbles can provide reference for the others in matting algorithms. Actually, the quality of resulting alpha mattes depends heavily on the number of scribbles and where they are drawn. In addition, drawing useful scribbles requires matting knowledge and experience to meet some prior assumptions of the matting algorithms, making it little impractical to non-professional participants.

In addition, although recent deep-learning based matting methods~\cite{hou2019context,hao2019indexnet,Tang_2019_CVPR,cai2019disentangled,Zhang_2019_CVPR} can achieve impressive alpha mattes, they sometimes fail when handling real-world images because their prediction models are usually trained on the composite datasets. The composition rules in~\cite{Xu2017Deep} can sometimes make the foreground and background disharmonious, resulting in obvious artifacts on training images. Factors such as illumination and shadow in natural images can amplify the influence of artifacts, and discount the final results. Our motivation is to automatically select relevant regions that meet the prior requirements, and all the user has to do is drawing scribbles on these suggested regions. We refer to these regions that determine the quality of alpha mattes as \emph{informative regions}. The \emph{informative regions} can acquire accurate labels through limited user interaction, which ensures that we have correct labels as a reference when processing different kinds of input images. Besides, we observe that overmuch scribbles are unnecessary for matte generation, because we can infer their category labels from existing scribbles in the \emph{informative regions}.

The main challenge of our motivation is to identify informative regions and effectively reduce user interactions. To strike a balance between the user effort and matting accuracy, we propose a unified framework, referred to as \emph{smart scribbles}, to guide users to draw scribbles on the suggested regions, and then propagate category labels to the remaining regions. \emph{Smart scribbles} can achieve alpha mattes with relatively little user effort: they only need to draw fewer scribbles on \emph{informative regions} to separate the foreground, background and unknown. Specifically, our framework first automatically select \emph{informative regions} based on the similarity, diversity, label entropy, and edge maps~\cite{Zitnick2014Edge} of different regions, and these four terms are summarized as \emph{information content}. Users are then asked to draw scribbles on these \emph{informative regions} to label different categories (\ie, foreground, background and unknown). After that, these labels are propagated to adjacent regions via Markov propagation. The \emph{informative regions} selection, drawing scribbles and Markov propagation are iterated several times to provide essential category labels, then we employ a deep network to capture semantic relevance in the image and adopt high-level semantics to further spread scribbles. In contrast to Markov propagation that is limited to local image features, CNN propagation can refine the alpha mattes in a global manner. Extensive experiments show that our \emph{smart scribbles} can generate high-quality alpha mattes with little user effort. In addition, the proposed two-phase propagation can spread scribbles to the whole image to improve alpha mattes, outperforming the state-of-the-art methods. Figure~\ref{fig:teaser} compares smart scribbles with the Late Fusion~\cite{Zhang_2019_CVPR}, traditional scribbles and Active Matting~\cite{NIPS2018_7710}, and our method illustrates more sophisticated texture details and clear boundary contour.

The main contributions of this paper are:
\begin{itemize}
	\item We propose a novel interactive framework (\emph{smart scribbles}) to generate alpha mattes from limited scribbles, which is more flexible and robust for natural image matting and can dramatically reduce user effort.
	\item We design an efficient method for computing \emph{information content}. It can identify the most \emph{informative regions} that are significant for the quality of alpha mattes.
	\item We present a two-phase propagation approach which can aggregate local and global information to spread limited scribbles to the whole image to generate an alpha matte effectively.
	\item Our method can achieve high-quality alpha mattes independently of large-scale matting datasets, and demonstrate more adaptability in real-world applications.
\end{itemize}

\section{Related Work}
\label{related}

In this section, we briefly review image matting from scribbles-based and trimap-based methods, and then introduce some deep learning based methods that have developed rapidly in recent years.

\textbf{Scribbles-based Matting.} In earlier works \cite{Levin2007A,Levin2008Spectral,Guan2006Easy,Wang2005An}, scribbles are widely used for natural image matting. Users only need to draw several scribbles on the input image with distinguished color to label the foreground, background and unknown. Scribbles-based methods are very convenient for common users and can achieve comparable alpha mattes in a user-friendly way. However, such scribbles only cover a small portion of the image, which means the alpha mattes will discount for slightly more complicated images. Besides, scribbles must fit the initial assumptions or prior distribution of matting algorithms, hence users are required to possess professional knowledge about the matting algorithms and rich experience in where to place their scribbles. For example, some methods \cite{Levin2007A,Levin2008Spectral} assume that all colors within a small window around an unknown pixel lie in a line of the color space. This color-line assumption cannot provide global information, and drawing scribbles that fit these algorithms would be a great challenge for novice users.

\textbf{Trimap-based Matting.} Compared with scribbles, trimaps can supply annotations for each pixel in the image, thus can provide sufficient category information (foreground, background or unknown) for image matting. Due to this advantages, currently, most state-of-the-art image matting algorithms take trimaps as assistant input~\cite{Feng2016A,Cho2016Natural,Aksoy2017Designing,Xu2017Deep}. Their core ideas are spreading certain labels (the foreground and background) to unknown areas. According to the different way of utilizing foreground-background information, these algorithms can be divided into two categories: sampling-based and propagation-based. Sampling-based methods~\cite{Chuang2003A,Wang2007Optimized,Rhemann2011A,Karacan2015Image,Shahrian2013Improving,Feng2016A} assume that each unknown pixel can be represented by a pair of certain foreground/background pixels. Propagation-based methods~\cite{Levin2007A,Levin2008Spectral,Sun2004Poisson,Grady2005Random,Lee2011Nonlocal,Chen2013KNN,Aksoy2017Designing} use affinity of neighboring pixels to propagate the alpha values from certain areas to unknown ones.

Compared with scribbles, trimap-based methods do not need experience or professional skills but consume a lot of user labors. A well-defined trimap requires pixel-level interpret provided by users, which is very time-consuming and fussy. Therefore, traditional matting methods usually utilize the online benchmark~\cite{Rhemann2009A} dataset to evaluate their algorithms, which only contains 27 training examples and 8 test images. Although trimap-based methods can achieve competent alpha mattes, they have poor performance in practical application because delicate trimaps are inconvenient to label for common users.


\textbf{Deep Learning Based Matting.} Deep learning has contribute a lot to computer vision~\cite{xu2018efficient,zhang2020multi,yang2019drfn}, also been applied in alpha matting problem. Cho \textit{et al.}~\cite{Cho2016Natural} fused the results of KNN~\cite{Chen2013KNN} and ClosedForm~\cite{Levin2007A} with input images, and then fed them to a well-designed CNNs to generate alpha mattes. Xu \textit{et al.}~\cite{Xu2017Deep} concatenated the input image and trimap as 4-channels input, and applied an encoder-decoder network to reconstruct the alpha mattes from high-level semantic representation. The subsequent matting methods~\cite{lutz2018alphagan,Tang_2019_CVPR,hao2019indexnet,hou2019context,cai2019disentangled} mostly follow this design philosophy: extracting high-level features from the input image and the corresponding trimap, and then predict the alpha matte through complicated network architecture. However, they all require trimaps as auxiliary input, which limits their promotion in practical applications.

Some matting networks employ semantic segmentation network~\cite{Chen2018SHM,Zhang_2019_CVPR} or attention mechanism~\cite{Qiao_2020_CVPR} to accomplish alpha mattes without trimaps, and failure case will probably occur when semantic segmentation is not applicable~\cite{Zhang_2019_CVPR}. Xin \textit{et al.}~\cite{NIPS2018_7710} proposed an active framework to guide matte generation. Although the above methods can produce alpha mattes without trimaps, they all are trained on the composite images, which can sometimes result in obvious artifacts or poor performance on real-world images. Our \emph{smart scribbles} accept user scribbles, then execute Markov and CNN propagation in superpixels to spread category labels. The proposed model is independent of large-scale composite data, thus exhibits greater robustness and generalization on real-world images.

\begin{figure*}[t]
	\centering
	\includegraphics[scale=.36]{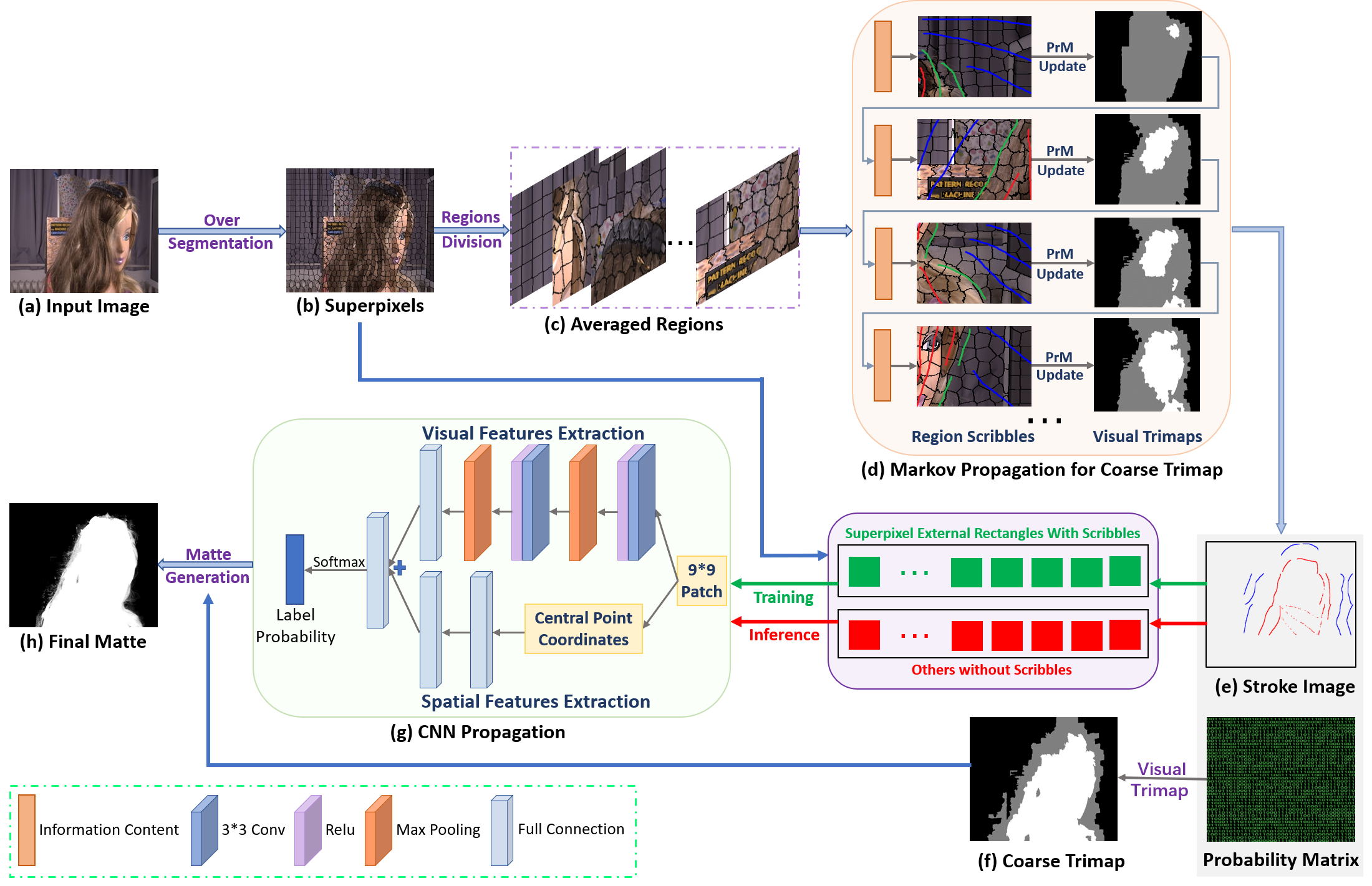}
	\caption{The pipeline of the proposed method. The input image is first over-segmented into superpixels, then is divided into regular rectangle regions of the same size. We calculate the information content of each region and the most informative region is automatically selected for users to draw scribbles, specifying the foreground (in red), background (in blue) and unknown areas (in green). These labels are then propagated to unlabeled regions to update the probability matrix (PrM) via two-phase propagation. During CNN propagation, we gather all superpixel external rectangles as input, and the superpixels with scribbles are used for training, while for the others we predict category labels for them using the trained model. After CNN propagation,  we update PrM to generate a refined trimap and the final matte can be produced by an embedded existing matting algorithm. }
	\label{fig:pipeline}
\end{figure*}

\section{Methodology}
\label{sec:methodology}

According to our previous analysis, a limited number of scribbles must meet some prior conditions to achieve comparable alpha mattes, while trimaps can help produce accurate matting results at the expense of much time and user labor in practical applications. Our goal is to achieve a balance between user effort and the quality of alpha mattes. For this purpose, we first use automatic region selection to replace the professional knowledge requirements for drawing scribbles, then we employ well-designed two-phase propagation to maximize the diffusion of limited category scribbles across the whole image, which can effectively reduce the number of scribbles drawn by users. Based on the observation that different image regions contribute unequally for matting algorithms and scribbles on critical ones can determine a desired alpha matte, we define \emph{informative regions} to represent the whole image and users only need to draw a few random scribbles on them. Since the informative regions receive accurate labels (foreground, background or unknown) from users, they can actively contribute to the rest part of the input image (\ie, effectively propagating user labels to uncertain regions).

\subsection{Overview}
\label{ssec:overview}
To interpret our \emph{smart scribbles}, we divide our framework shown in Figure~\ref{fig:pipeline} into four main components:
\\
\textbf{Over Segmentation:} Our framework is performed on superpixels level for a faster and more efficient generation. We intend to spread limited scribbles across all image areas, while pixel-level nodes do not reflect color and texture correlations between regions. In addition, compared to pixels, propagation on superpixels level can effectively reduce the number of nodes to decrease the total amount of computation. The input image is segmented into superpixels via simple linear iterative clustering (SLIC) algorithm~\cite{achanta2012slic} (as shown in Figure~\ref{fig:pipeline}). The number of superpixels can automatically adapt to the input size, thus we can obtain decent segmentation results for arbitrary-size images. The following steps are all based on superpixels, irrelevant to the attributes of image itself (size or classes etc.).
\\
\textbf{Regions Division and Selection:} After over-segmentation, as shown in Figure~\ref{fig:pipeline}, informative regions are chosen for expressing the whole image. We evenly divide the input image into $M \times M$ regular regions, then select the regions crucial for alpha mattes automatically, which can replace the professional knowledge requirements of users. After the regions division, we define the \emph{information content} of each region according to the superpixels inside it, which is explained in~\ref{ssec:information_definition}. For each iteration, the most critical region (the one with largest \emph{information content}) is selected and users only need to simply draw few scribbles on it. In practice, M can be adjusted according to the size of the input image and is empirically set as 4 considering the convenience of users in most cases.
\\
\textbf{Drawing Scribbles:} Users are required to draw scribbles on the selected region with distinguished color to label different areas (Foreground--red, Background--blue and Unknown--green), as shown in Figure~\ref{fig:pipeline}.
\\
\textbf{Information Propagation:} \emph{Smart scribbles} will record category labels and propagate these information to unlabeled regions. To achieve this, we construct the probability matrix (PrM) to represent the transition possibility between different superpixels, and the label probabilities of different categories ($pb$-background, $pu$-unknown and $pf$-foreground) of each superpixel can be calculated according to the probability matrix. The proposed two-phase propagation is essentially a continuous update of the probability matrix, including two stages: (1) The informative labels propagate to their adjacent regions using Markov chain~\cite{Mauro1999Markov}, by taking spatial and appearance similarity into account (as illustrated in Figure~\ref{fig:pipeline}), dubbed Markov propagation. (2) The output is further refined using a convolutional neural network by exploiting high-level features of superpixels in a global manner, dubbed CNN propagation. The detailed propagation process is developed in section~\ref{ssec:information_propagation}.


Overall, we iterate three operations, regions selection, drawing scribbles, and Markov propagation for $N$ times, to accomplish the full spread of scribbles in the neighborhood, and CNN propagation is then applied to generate the fine trimaps. Final alpha mattes are produced with embedded existing matting algorithms. In our experiments, parameter $N$ is set as 6 to balance the trade-off between efficiency and quality. More iterations can improve mattes slightly with user effort increasing apparently, and the detailed analysis about iteration number can refer to baseline $2$ in section~\ref{ssec:baselines}.


\subsection{Information Content Formulation}
\label{ssec:information_definition}

Scribbles often need to meet some prior assumptions, hence we propose \emph{informative regions} to replace such professional requirements of users. We hold that \emph{informative regions} should contain representative image semantics or decisive local features, and placing limited scribbles on these regions can achieve a competent alpha matte.
In contrast, image areas with single color or smooth texture are insignificant for drawing scribbles (we can assign their labels through later propagation). It is impractical for a novice user to distinguish where are \emph{informative regions}, and the definition of \emph{information content} can select the most informative region iteratively guiding users to draw scribbles.

Specifically, we conclude that the informative ones should: (1) have a high similarity with the neighborhood, which means representative; (2) have a high color and texture diversity inside, which makes the region contain as much scenarios as possible; (3) involve both foreground and background for having diverse and balanced labels; (4) locate on the boundary of some object, which may fully exploit context correlation. According to these prerequisites, we present a formulation of \emph{information content}, termed as $Info$, of a region based on the superpixels inside it. During each iteration, the region with maximum $Info$ is selected as the most informative one for drawing scribbles. $Info$ is expressed as Equation~(\ref{eq:info}).
\begin{equation}
\label{eq:info}
Info=\Upsilon+\Gamma+\Lambda+\Delta,
\end{equation}
where $\Upsilon$ denotes the similarity with its neighbors, $\Gamma$ is the diversity inside the region, $\Lambda$ represents the region entropy (foreground, background and unknown distribution) and $\Delta$ is the edge score of the region, corresponding the above prerequisites respectively. In practice, all these four entries will be balanced according to the number of superpixels in the current region. Among those, similarity ($\Upsilon$) is calculated as:
\begin{equation}
\label{eq:sim}
\begin{split}
&\Upsilon=\lambda_{1}\sum_{i}^{\Omega_{in}}\sum_{j}^{\Omega_{out}}exp(-\frac{||cm_{i}-cm_{j}||^{2}}{2\sigma^{2}})+\\
&\lambda_{2}\sum_{i}^{\Omega_{in}}\sum_{j}^{\Omega_{out}}\frac{2(ch_{i}-ch_{j})^{2}}{ch_{i}+ch_{j}+\theta}+\lambda_{3}\sum_{i}^{\Omega_{in}}\sum_{j}^{\Omega_{out}}\frac{2(th_{i}-th_{j})^{2}}{th_{i}+th_{j}+\theta},
\end{split}
\end{equation}

meanwhile, diversity ($\Gamma$) is expressed as :
\begin{equation}
\label{eq:diversity}
\begin{split}
&\Gamma=-(\lambda_{1}\sum_{i}^{\Omega_{in}}\sum_{j}^{\Omega_{in}}exp(-\frac{||cm_{i}-cm_{j}||^{2}}{2\sigma^{2}})+\\
&\lambda_{2}\sum_{i}^{\Omega_{in}}\sum_{j}^{\Omega_{in}}\frac{2(ch_{i}-ch_{j})^{2}}{ch_{i}+ch_{j}+\theta}+\lambda_{3}\sum_{i}^{\Omega_{in}}\sum_{j}^{\Omega_{in}}\frac{2(th_{i}-th_{j})^{2}}{th_{i}+th_{j}+\theta}),
\end{split}
\end{equation}
where $\Omega_{in}$ is the set of superpixels inside the region, while $\Omega_{out}$ is the set of superpixels outside the region. $cm_{i}$, $ch_{i}$, and $th_{i}$ are middle-level features (color mean, color histogram, and texture histogram) at superpixel $i$. $\theta$ is a bias to prevent the denominator from being $0$. The coefficients $\lambda_{1}$, $\lambda_{2}$ and $\lambda_{3}$ are set to $0.4$, $0.35$ and $0.25$ in practice. The diversity is obtained by taking the overall similarity negative, because the higher the similarity, the smaller the difference, and vice versa.

We define region entropy ($\Lambda$) as:
\begin{equation}
\label{eq:entro}
\begin{split}
\Lambda=-\sum_{i}^{\Omega_{in}}[&(pb_{i})log(pb_{i})+(pu_{i})log(pu_{i})\\
&+(pf_{i})log(pf_{i})],
\end{split}
\end{equation}
where $pb$, $pu$ and $pf$ refer to the Foreground, Background and Unknown probabilities of superpixel $i$ respectively.

\begin{figure*}[htp]
	\setlength{\tabcolsep}{1pt}\small{
		\begin{tabular}{ccccc}
			\includegraphics[scale=0.625]{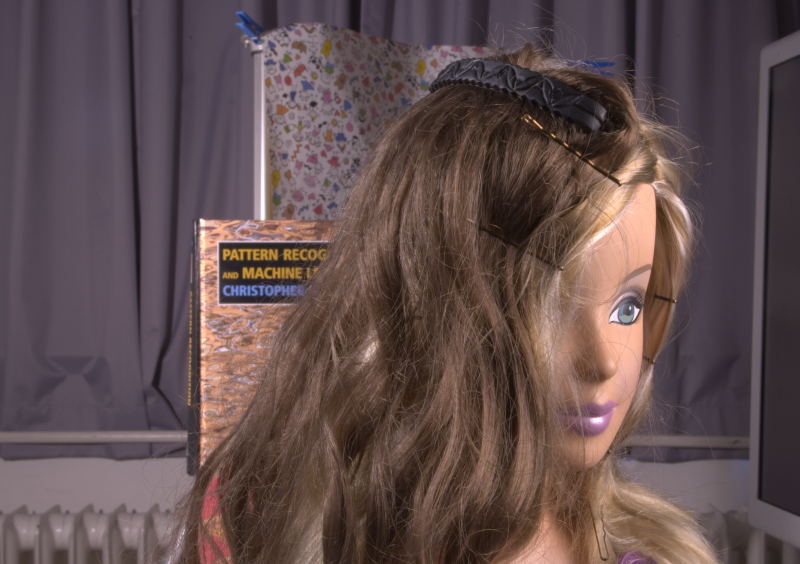} &
			\includegraphics[scale=0.15]{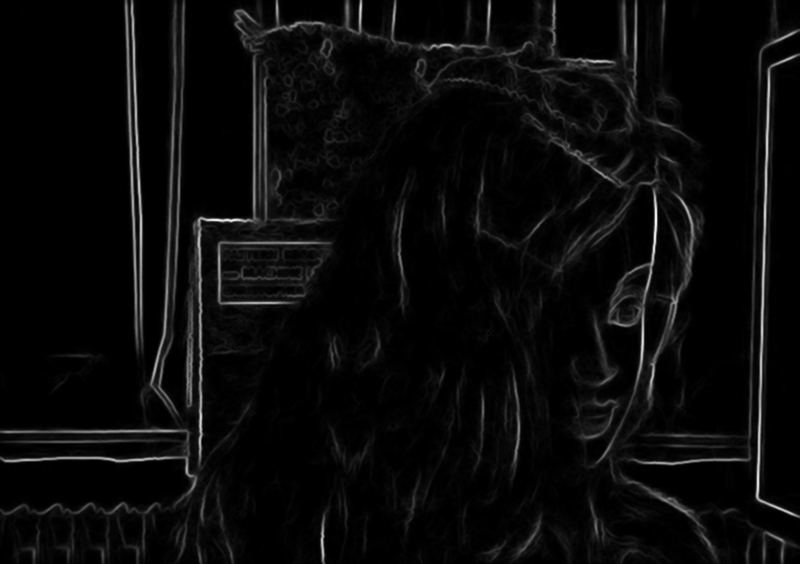} \\
			
			Input Image & Edgemap Image \\
	\end{tabular}}
	\caption{Edge score (S) is calculated according to the corresponding edgemap image. }
	\label{fig:rectangles}
\end{figure*}

According to the object edges in~\cite{Zitnick2014Edge}, we define edge score ($\Delta$) as:
\begin{equation}
\label{eq:edge}
\Delta=\sum_{i}^{\Omega_{in}}e_{i}=\sum_{i}^{\Omega_{in}}\frac{\sum_{k}^{\psi_{i}}exp(em_{k}\delta)}{\varepsilon},
\end{equation}
where $e_{i}$ is the edge value of superpixel $i$, calculated via edge value of pixels inside it. $\psi_{i}$ is the set of pixels inside the superpixel $i$, $em_{k}$ is the edge value of pixel k, and $\delta$, $\varepsilon$ are coefficients. An example edgemap image is shown in Figure~\ref{fig:rectangles}, where white lines denotes potential object edges. Obviously, a high edge score means the region has a high probability of being at the edge of the object, corresponding to potential unknown in trimaps. The edge score can suggest transition content between different types, providing context correlation of separate regions.

\subsection{Two-phase Information Propagation}
\label{ssec:information_propagation}
The region with maximum \emph{information content} is selected for drawing scribbles with distinguished colors, and some superpixels inside it are assigned with explicit category labels (foreground-red, background-blue, unknown-green) correspondingly. In order to minimize user effort and exploit the limited feedback from users (\ie, the labeled superpixels), we present a two-phase information propagation strategy to spread scribbles to the whole image, which not only takes advantage of local/low-level features with a re-modeled Markov process, but also the global/high-level semantics with a CNN model. We construct a probability matrix (PrM) to solve the label possibility of foreground, background and unknown areas. The element $PrM_{ij}$ in PrM indicates the transfer probability from superpixel i to j, and the two-phase propagation can update PrM continuously. Finally, a refined trimap can be inferred from PrM according to Algorithm~\ref{al:prop}.

\textbf{Markov Propagation:} We expect that user scribbles can propagate to the adjacent unlabeled superpixels massively, so employ the excellent spatial transfer capacity of Markov Chain~\cite{Mauro1999Markov,Sun2015Saliency} to achieve this purpose (see Figure~\ref{fig:pipeline}). In our framework, each superpixel in the image is modeled as a `node' in the Markov Chain (\ie, node i denotes superpixel i). We consider the superpixels with certain labels as absorbing nodes, while the unlabeled ones are transition nodes. Therefore the labels propagation can be treated as the state transfer in Markov Chain, and solving unknown superpixel category labels is a process of probability transfer from transition nodes to absorbing ones. For the concrete implementation, suppose we divide n superpixels from over-segmentation, firstly, we can construct a connection matrix (CoM) to describe the position relevance of superpixels in the whole image. Then we can establish an affinity matrix (AfM) to measure the color and texture similarity between the superpixels. Both CoM and AfM are obviously $n \times n$ symmetric matrix. The probability matrix (PrM) is calculated by the CoM, AfM and labels information, which is a $n \times m$ matrix (suppose there are $m$ superpixels with user scribbles until current iteration). The element $PrM_{ij}$ in PrM indicates the probability of transition from node $j$ (denotes labeled superpixel $j$) to node $i$ (denotes unlabeled superpixel $i$), which is named as transition probability.

\begin{algorithm*}[t]
	\caption{Markov Propagation}
	\leftline{\hspace*{0.02in} {\bf Input:} Connection Matrix (CoM), Affinity Matrix (AfM), Labeled Superpixels Set (${L}$), Unlabeled}
	\leftline{\hspace*{0.48in}Superpixels Set(${U}$).}
	\leftline{\hspace*{0.02in} {\bf Output:}
		Probability of Unlabeled Superpixels {$\{pf_{i}\},\{pb_{i}\},\{pu_{i}\}$}}
	\begin{algorithmic}[1]
		\FOR{iterations}
		\STATE receives labels information from user drawing scribbles;
		\STATE   update ${L}$ and ${U}$;
		\STATE $(CoM, AfM, ${L}$, ${U}$)$ $\rightarrow$ Probability Matrix $(PrM)$;
		\FOR{Superpixel $i$ $\in$ ${U}$}
		\STATE $SUM_{f}=0$, $SUM_{b}=0$, $SUM_{u}=0$;
		\FOR{Superpixel $j$ $\in$ ${L}$}
		\STATE \textbf{Case} {$j \in Foreground$} : $SUM_{f}=SUM_{f}+PrM_{ij}$
		\STATE \textbf{Case} {$j \in Background$} : $SUM_{b}=SUM_{b}+PrM_{ij}$
		\STATE \textbf{Case} {$j \in Unknown$} : $SUM_{u}=SUM_{u}+PrM_{ij}$
		\ENDFOR
		\STATE $pf_{i}=\frac{SUM_{f}}{SUM_{f}+SUM_{b}+SUM_{u}}$
		\STATE $pb_{i}=\frac{SUM_{b}}{SUM_{f}+SUM_{b}+SUM_{u}}$
		\STATE $pu_{i}=\frac{SUM_{u}}{SUM_{f}+SUM_{b}+SUM_{u}}$
		\ENDFOR
		\ENDFOR
		\STATE \RETURN $\{pf_{i}\},\{{pb_{i}}\},\{{pu_{i}}\}$
	\end{algorithmic}
	\label{al:prop}
\end{algorithm*}

Transition probability is defined by the low-level appearance (\eg, color, texture) and spatial similarity between superpixels, and a higher transition probability $PrM_{ij}$ indicates that $i$ and $j$ more likely share the same labels (foreground, background or unknown). With the labeled superpixels increasing by drawing scribbles, the PrM is extended through the Markov Propagation in each iteration. For unlabeled superpixel $i$, its labels probabilities $pf_{i}$, $pb_{i}$ and $pu_{i}$ in Equation~(\ref{eq:entro}) will be revised according to algorithm~\ref{al:prop}. The updated labels probabilities will in turn renovate region entropy (E) in Equation~(\ref{eq:info}) for next selection. Markov Propagation iterates along with informative regions selection, and labels probabilities are revised continuously.

\textbf{CNN Propagation:} After $N$ times of Markov Propagation, a coarse trimap is generated as shown in Figure~\ref{fig:pipeline}. Although most superpixels can acquire proper labels, the propagation flow is localized and lack of consideration for high-level semantics of the superpixels.  Convolutional Neural Network (CNN) has gained success in extracting high-level features of the input images~\cite{Long2015Fully,Chen2014Semantic}, and is also proved to be capable of spreading information and labels regardless of spatial limit~\cite{Endo2016DeepProp}. Thereby, we develop an efficient CNN to extract features from the superpixels for global information propagation. We gather all the superpixels and extract their external rectangles to feed in the CNN Propagation, then acquire corresponding labels information from the preserved stroke image (Figure~\ref{fig:pipeline}), the course trimap and the input image. The labeled superpixels (with users scribbles or sky-high label probabilities) are considered as training set. The unknown typically lie in the transition between foreground and background, and their characteristics vary widely. Therefore in context irrelevant superpixels-level CNN propagation, we only consider foreground and background (Figure~\ref{fig:pipeline}) labels. After training, we can predict the label probabilities ${pf_{i}},{pb_{i}}$ for unlabeled superpixels via the trained model. 

Figure~\ref{fig:pipeline} illustrates the CNN Propagation. The architecture has two branches: one is to extract high-level semantics from the centering patch of each superpixel via $3$ convolutional layers. The visual features extraction takes a mini-batch for input, and output a vector with $256$ elements suggesting semantic relevance. The other branch is to encode the spatial information to a vector with $256$ elements via a fully connected (FC) layer. We feed the center coordinates of the mini-batch into FC layer to provide some context correlation considering the global propagation. The semantic and spatial vectors are integrated by adding element-wisely. The integrated output is converted to a $3*1$ label probability map using a softmax function at last. The whole network is trained 20 epoches with descending learning rate and cross-entropy loss.

The proposed two-phase propagation can accomplish the overall update of PrM, and we can predict a desired trimap via algorithm~\ref{al:prop}. Finally, some superpixels in the whole image obtain their labels (foreground, background, or unknown) from scribbles, while others from the two-phase information propagation by thresholding label probabilities. Here we regard the superpixels whose ${pf_{i}}$ is bigger than 0.65 as foreground empirically. Similarly, our framework can infer more background superpixels from the unlabeled ones. Remainder unlabeled superpixels are considered as Unknown. Afterwards, \emph{smart scribbles} transform the foreground superpixels to white; background ones to black; unknown ones to gray. Thus, we can obtain a refined trimap and the corresponding alpha matte is generated by embedding existing matting algorithms.

\section{Experiments and Results}
\label{sec:experiments}

In this section, we present our experiments and results analysis. The visual results on the real-world images are shown in~\ref{ssec:comWithAM}, which can demonstrate the robustness and generalization of \emph{smart scribbles}. In addition, we compare with traditional scribble-based methods on the matting benchmark~\cite{Rhemann2009A}, the portrait dataset~\cite{Shen2016Deep} and the deep image matting~\cite{Xu2017Deep} dataset (denoted as DIM) to exhibit the superiority of \emph{smart scribbles}. We compare the alpha mattes generated by \emph{smart scribbles} with several kinds of artificial trimaps based on the matting benchmark~\cite{Rhemann2009A} to demonstrate the applicability with different matting algorithms. Then we evaluate our approach with $3$ baselines and perform ablation study for \emph{smart scribbles}, both on the matting benchmark~\cite{Rhemann2009A}. The following experiments are all implemented using MATLAB, on a PC with an NVIDIA GTX 1080Ti GPU. We invite 20 users to participate in our experiments, and all of them are inexperienced to matting as well as our framework. All these users draw scribbles through 'mspaint' program under windows system. In all following illustrations, red scribbles, blue scribbles and green scribbles represent the foreground, background and unknown respectively (the scribbles are appropriately bolded to make them more visible).
\begin{figure*}[t]
	\setlength{\tabcolsep}{1pt}\small{
		\begin{tabular}{cccccc}
			\includegraphics[scale=0.056]{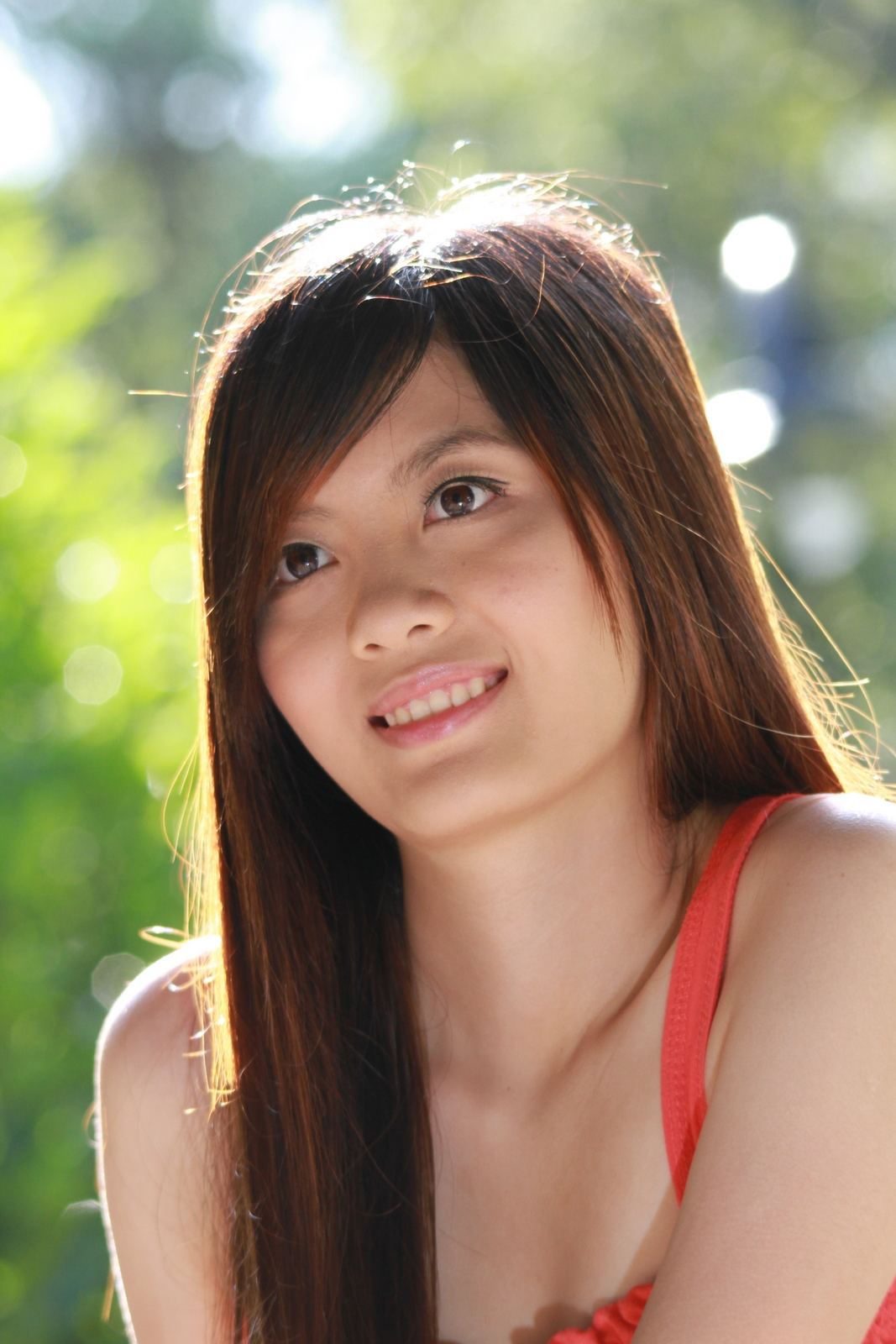} &
			\includegraphics[scale=0.0745]{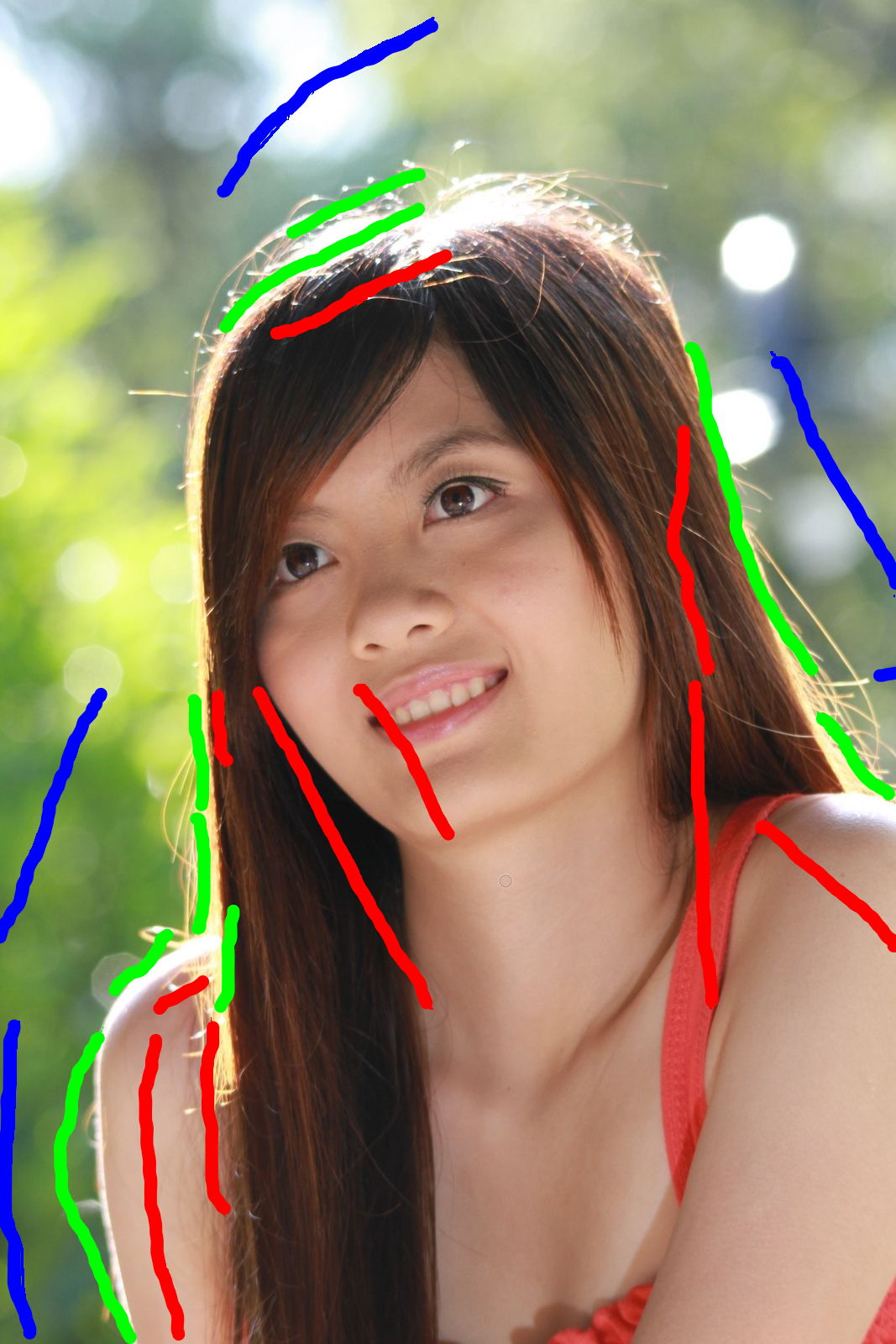} &
			\includegraphics[scale=0.056]{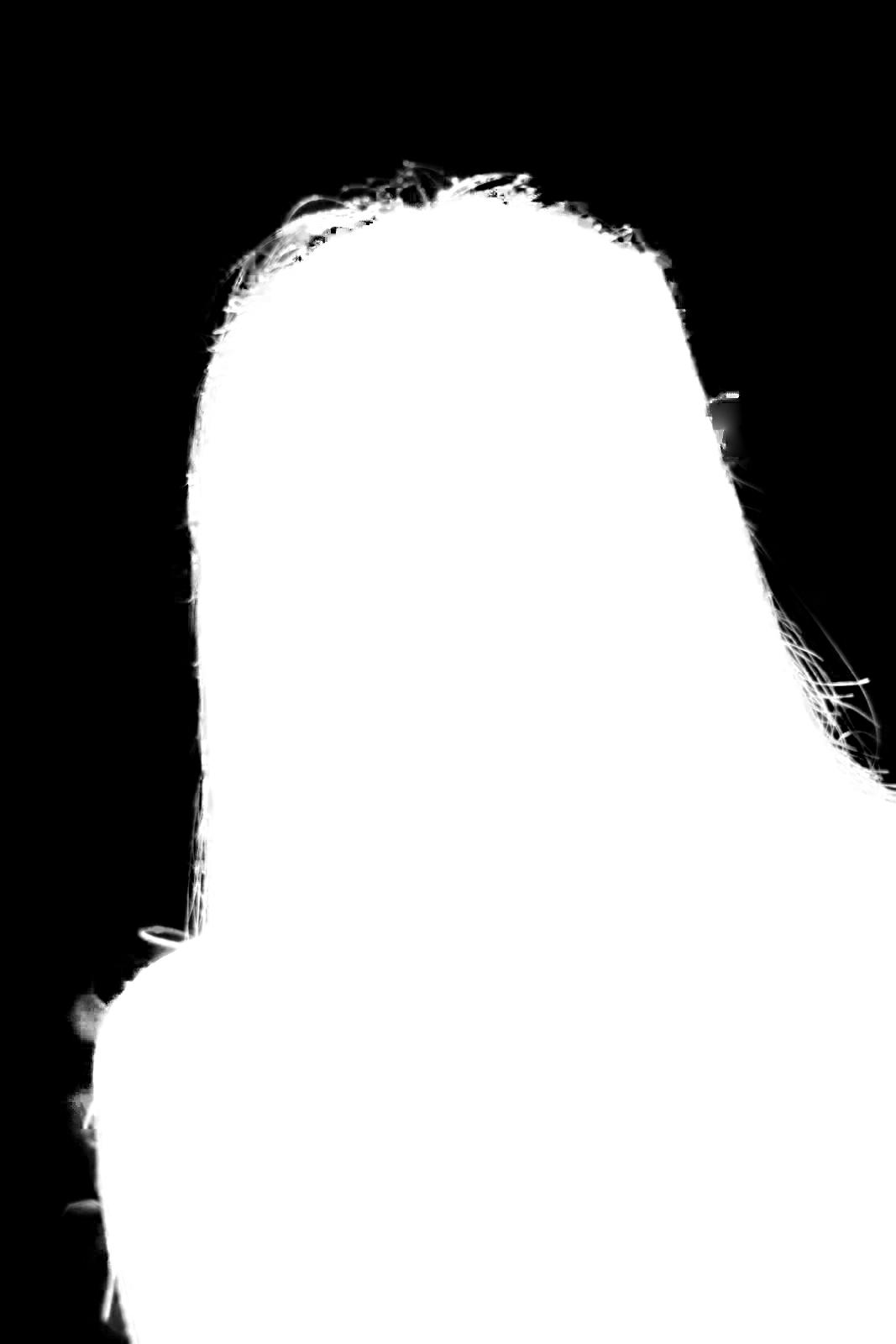} &
			\includegraphics[scale=0.056]{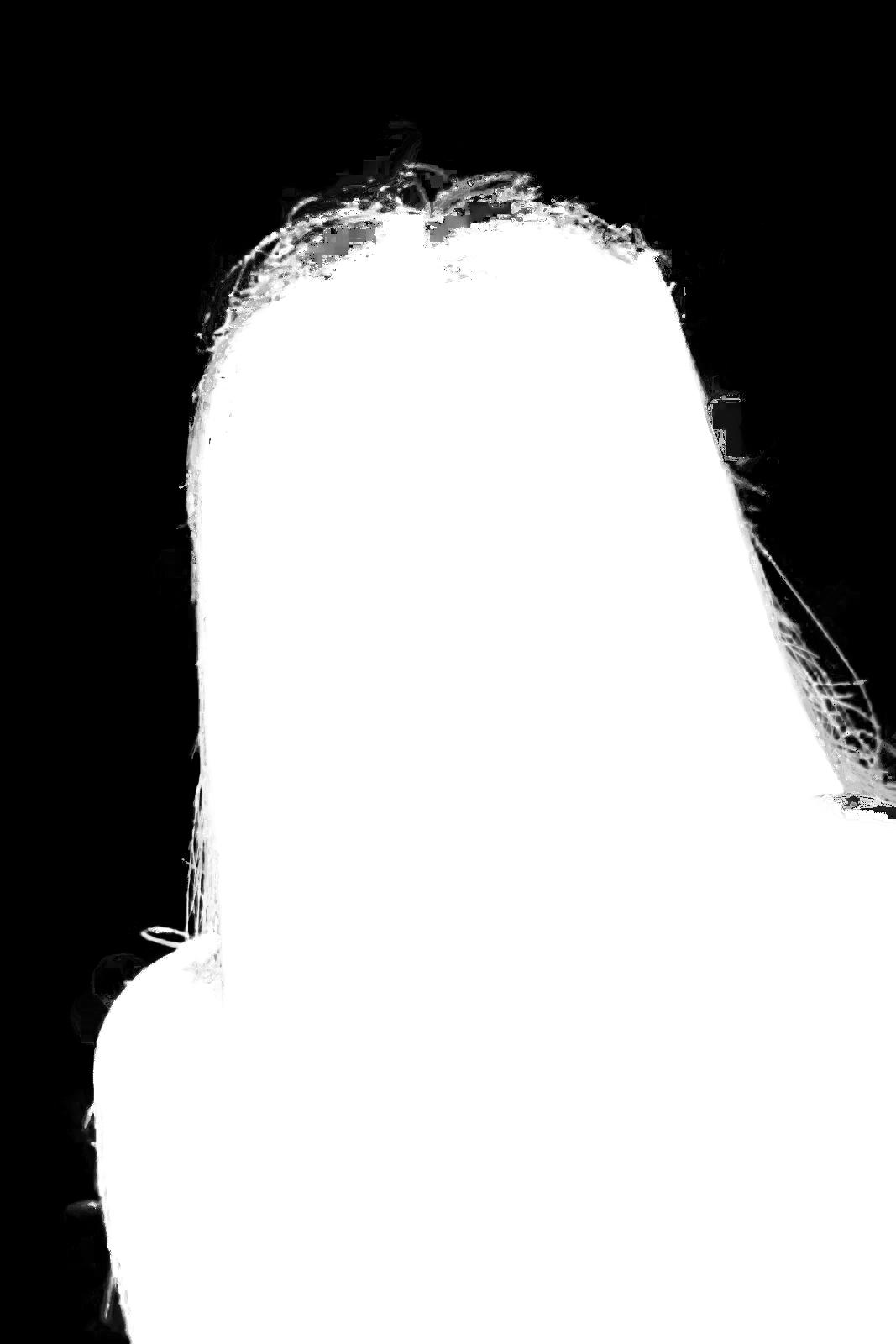} &
			\includegraphics[scale=0.056]{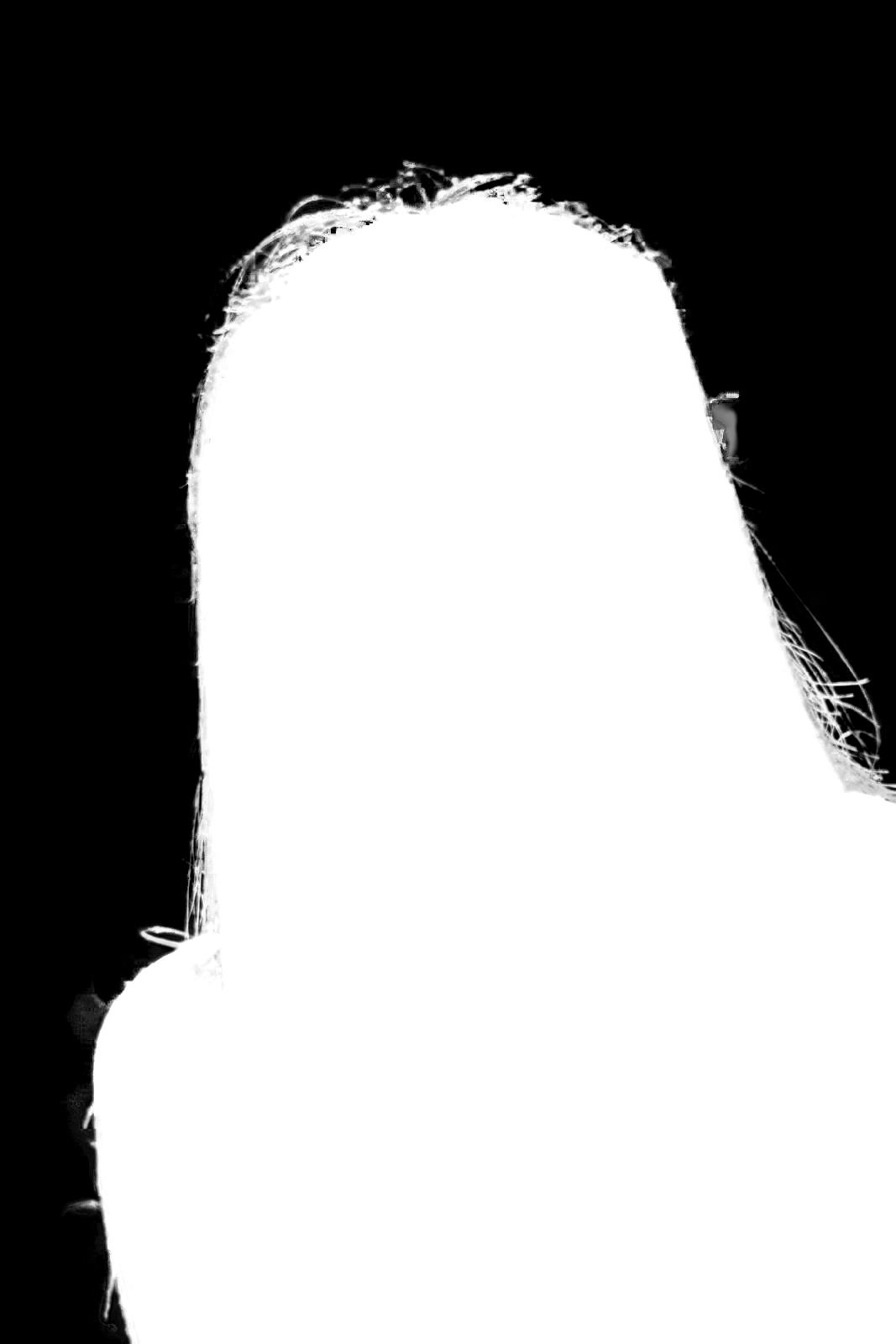} &
			\includegraphics[scale=0.056]{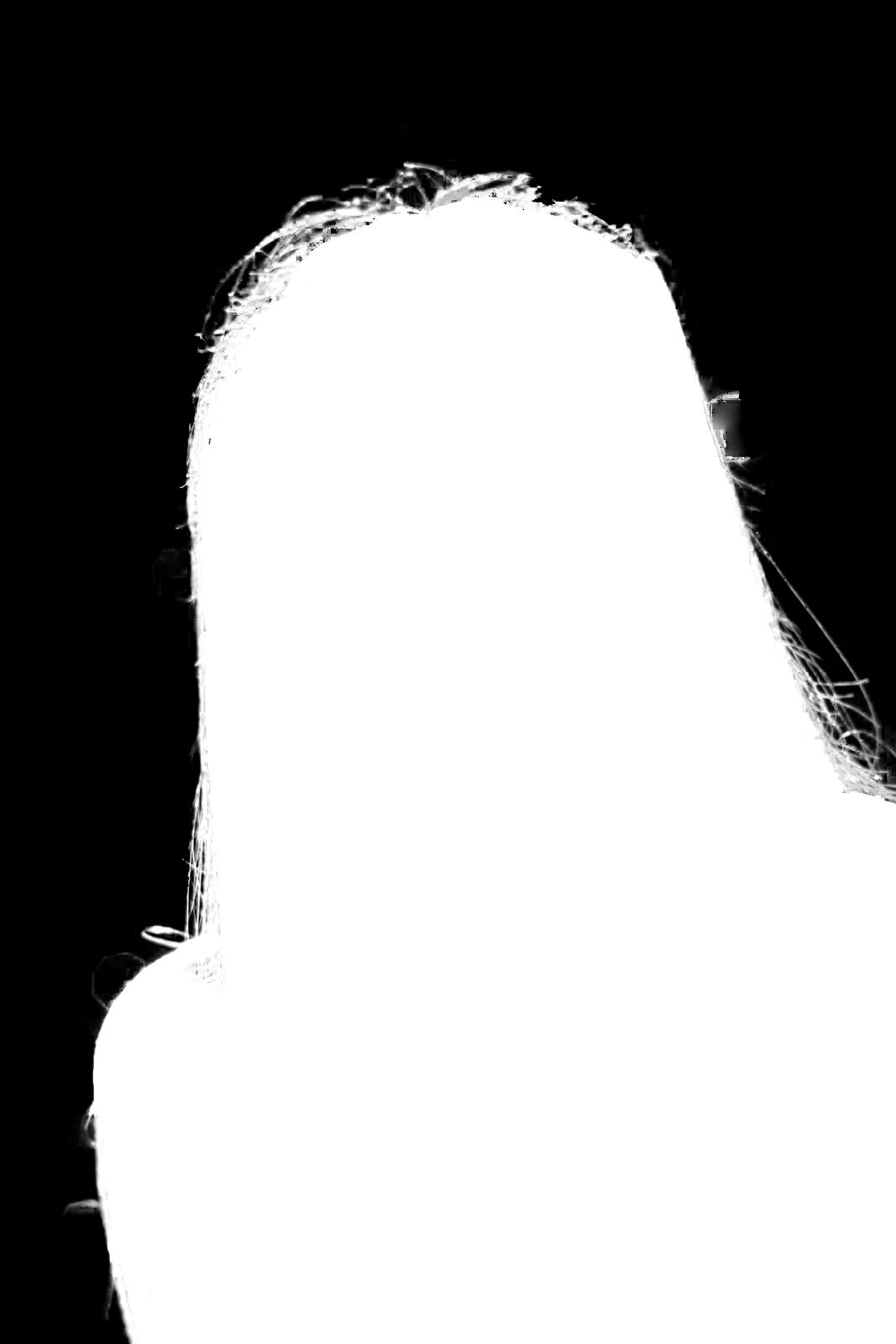} \\
			
			Input Image & Smart Scribbles & SS+Closed~\cite{Levin2007A} & SS+KNN~\cite{Chen2013KNN} & SS+DCNN~\cite{Cho2016Natural} & SS+IFM~\cite{Aksoy2017Designing} \\
			
			\includegraphics[scale=0.075]{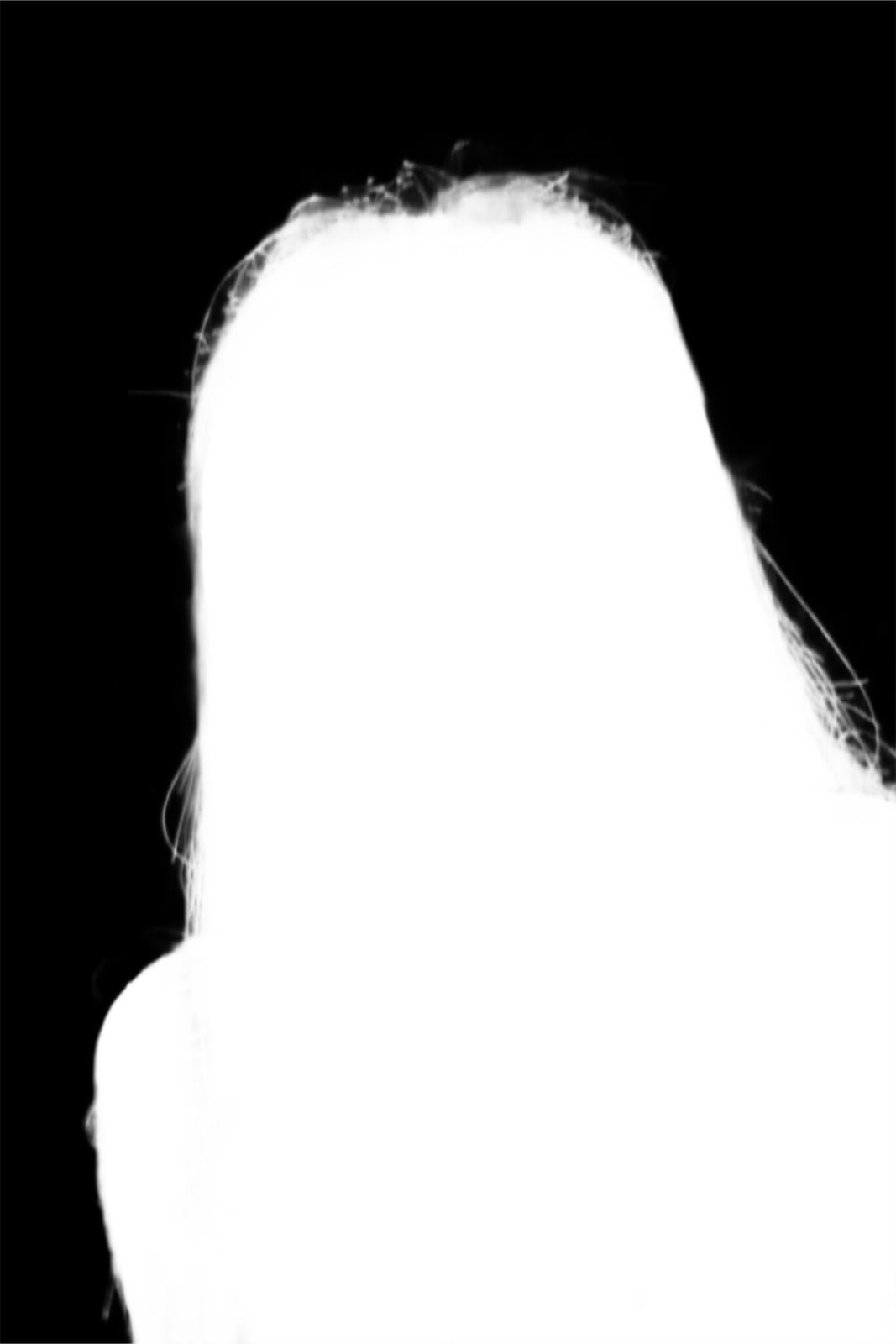} &
			\includegraphics[scale=0.056]{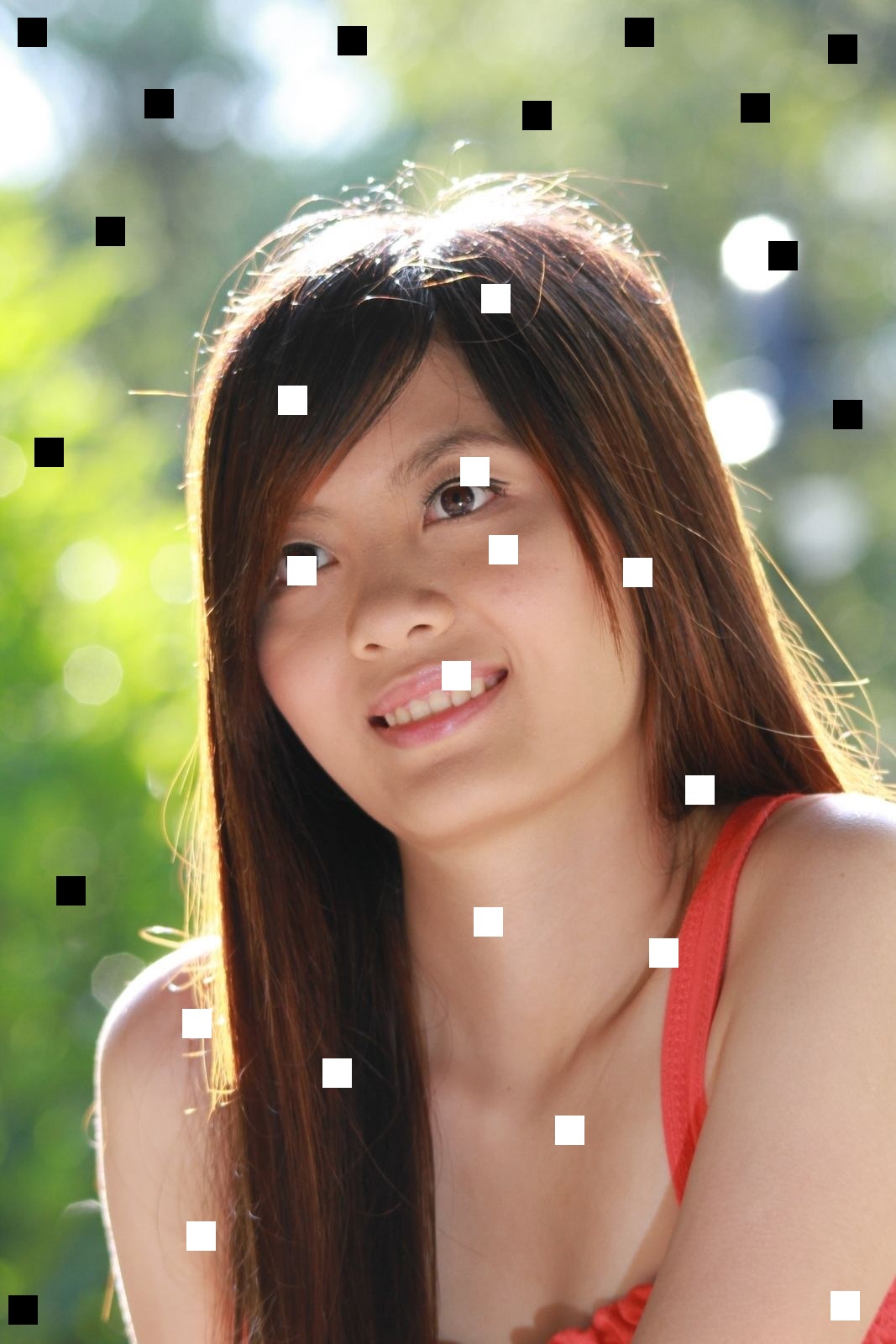} &
			\includegraphics[scale=0.056]{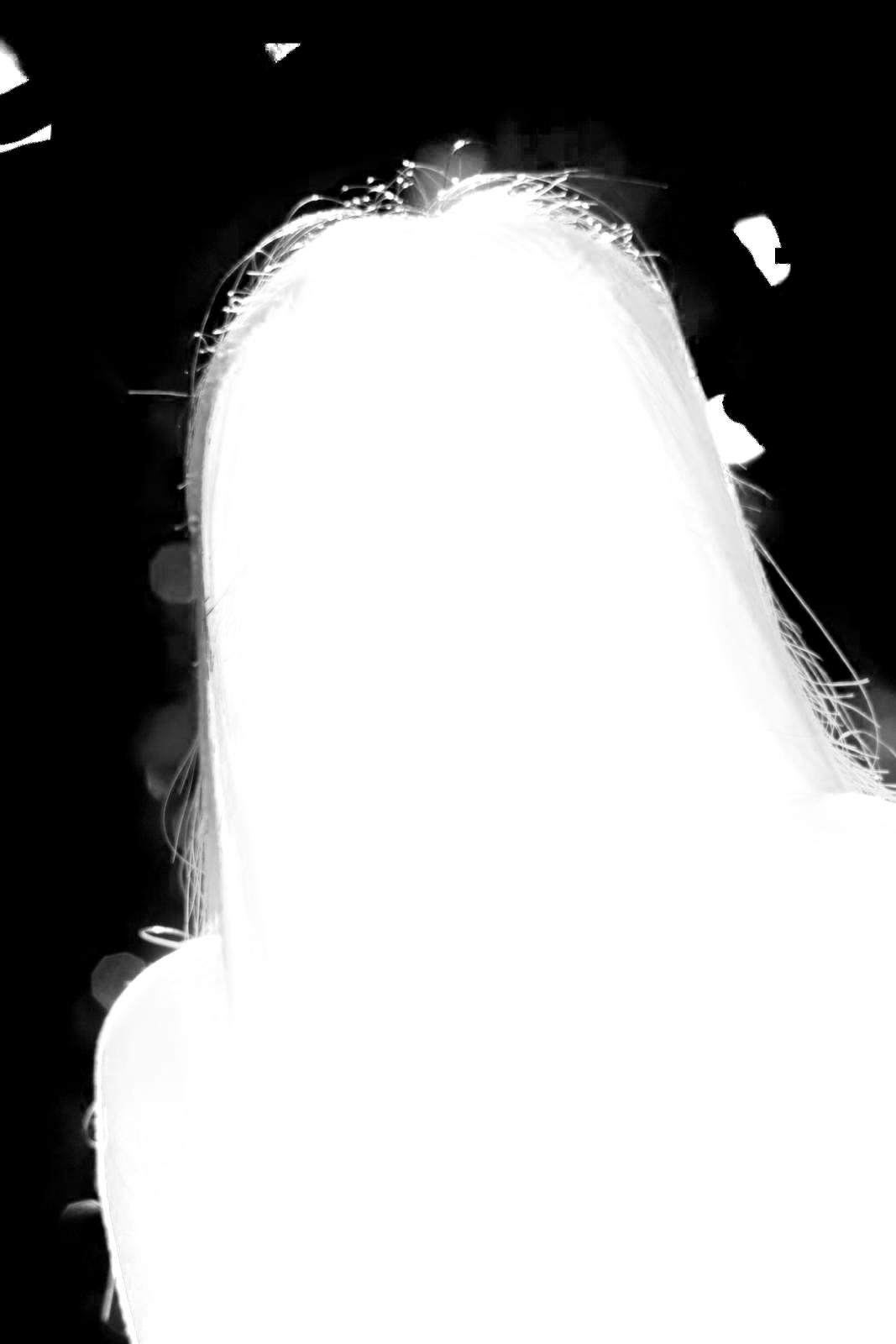} &
			\includegraphics[scale=0.056]{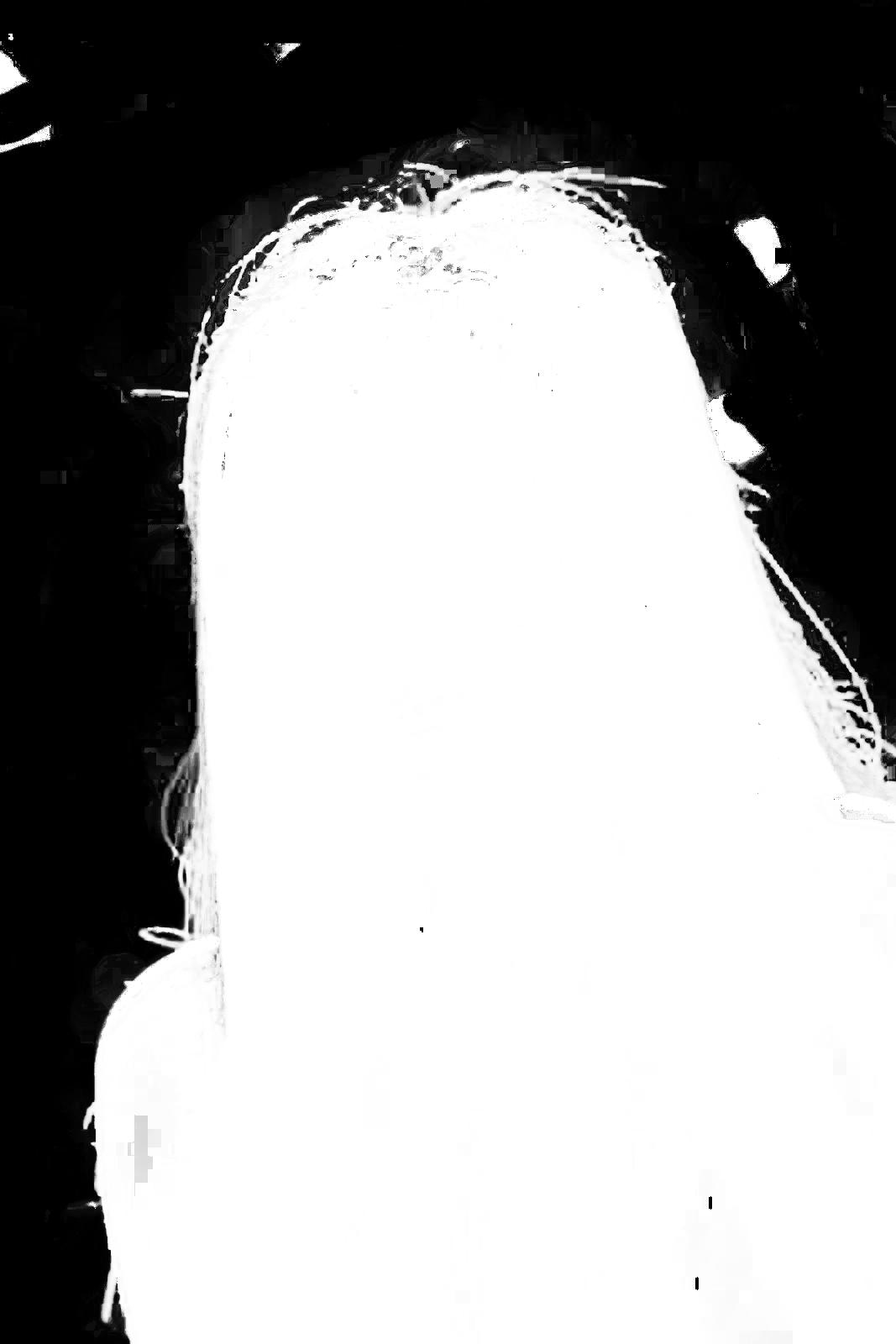} &
			\includegraphics[scale=0.056]{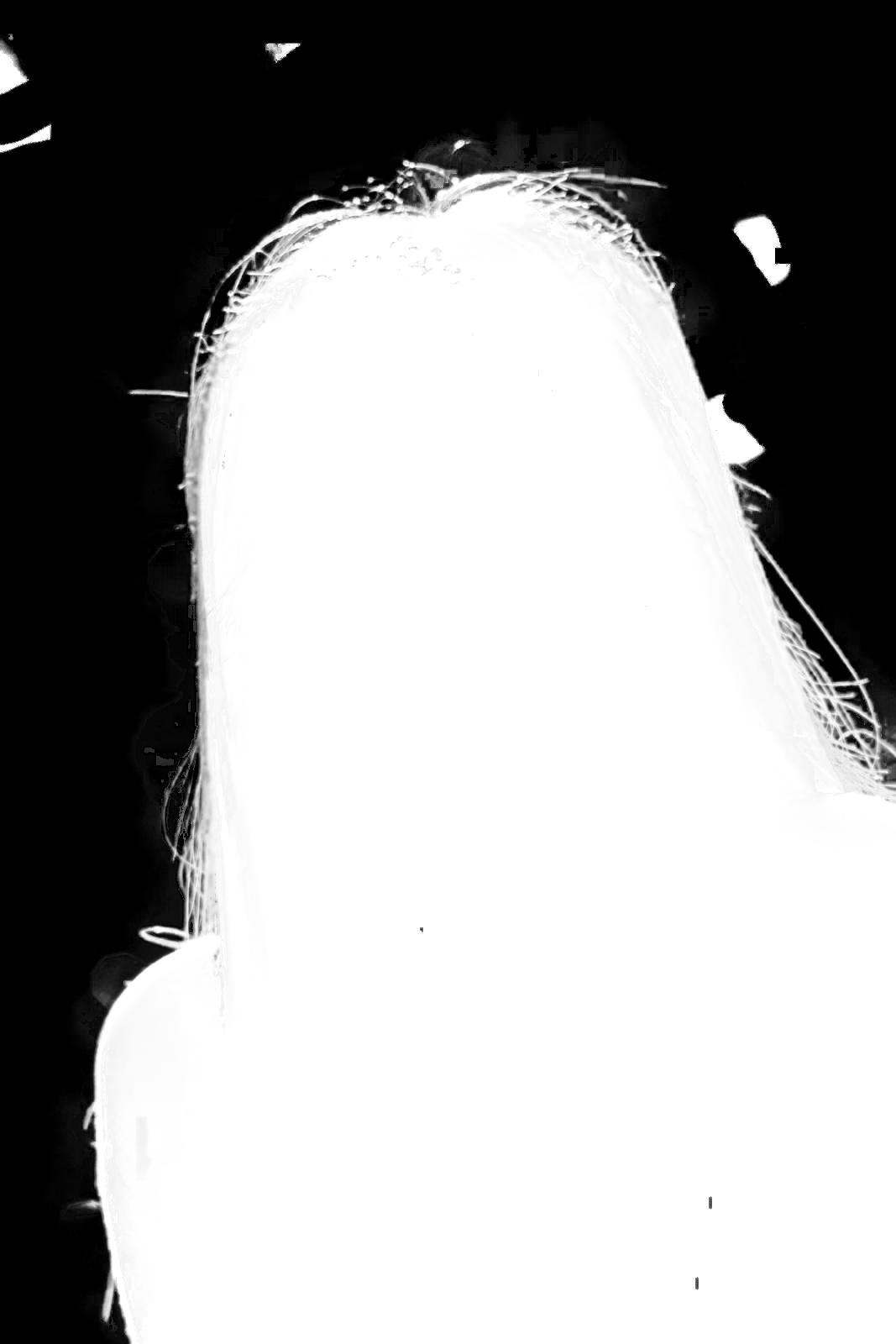} &
			\includegraphics[scale=0.056]{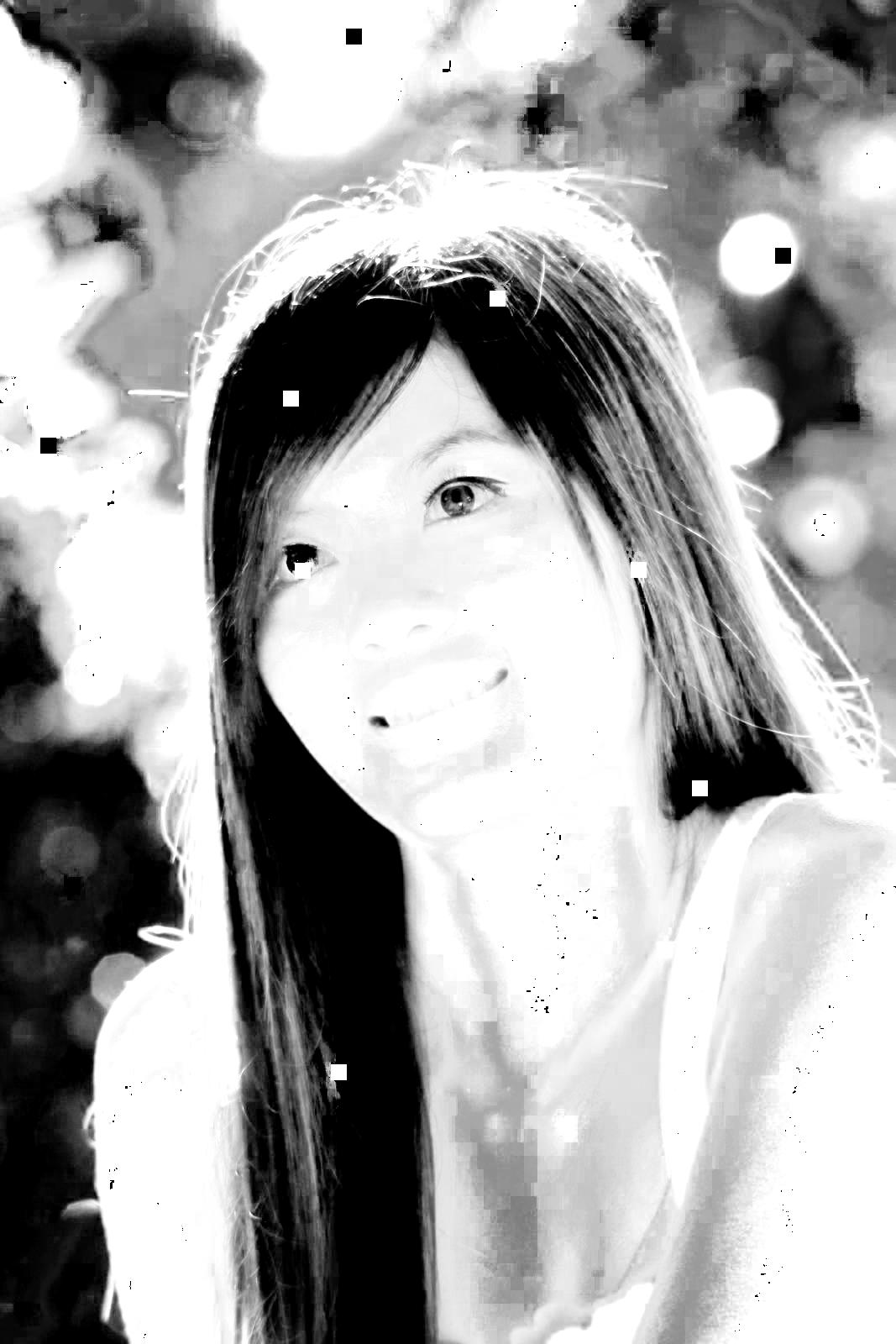} \\
			
			Late Fusion~\cite{Zhang_2019_CVPR} & AM~\cite{NIPS2018_7710} & AM+Closed~\cite{Levin2007A} & AM+KNN~\cite{Chen2013KNN} & AM+DCNN~\cite{Cho2016Natural} & AM+IFM~\cite{Aksoy2017Designing} \\
			
			\includegraphics[scale=0.04]{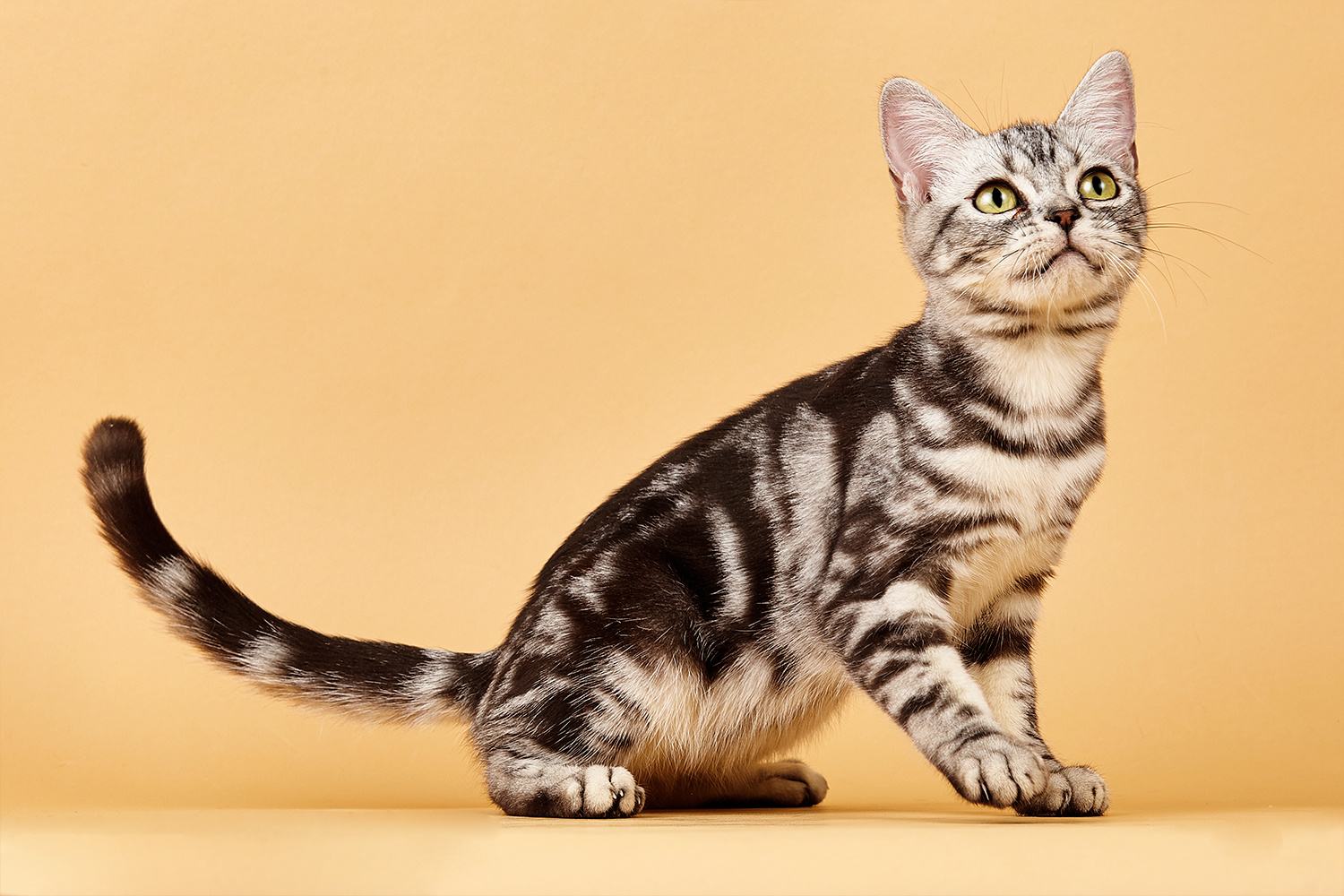} &
			\includegraphics[scale=0.0535]{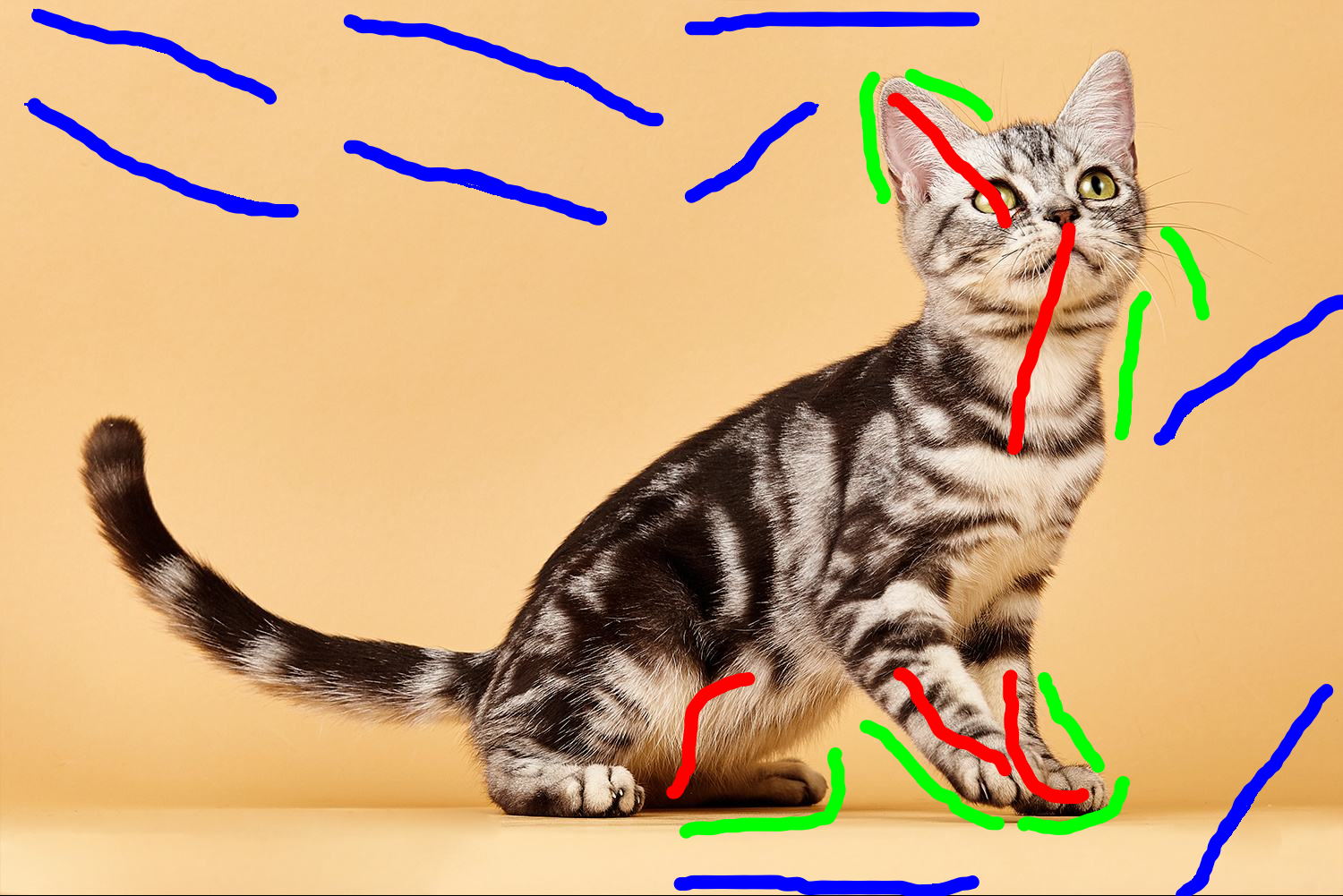} &
			\includegraphics[scale=0.04]{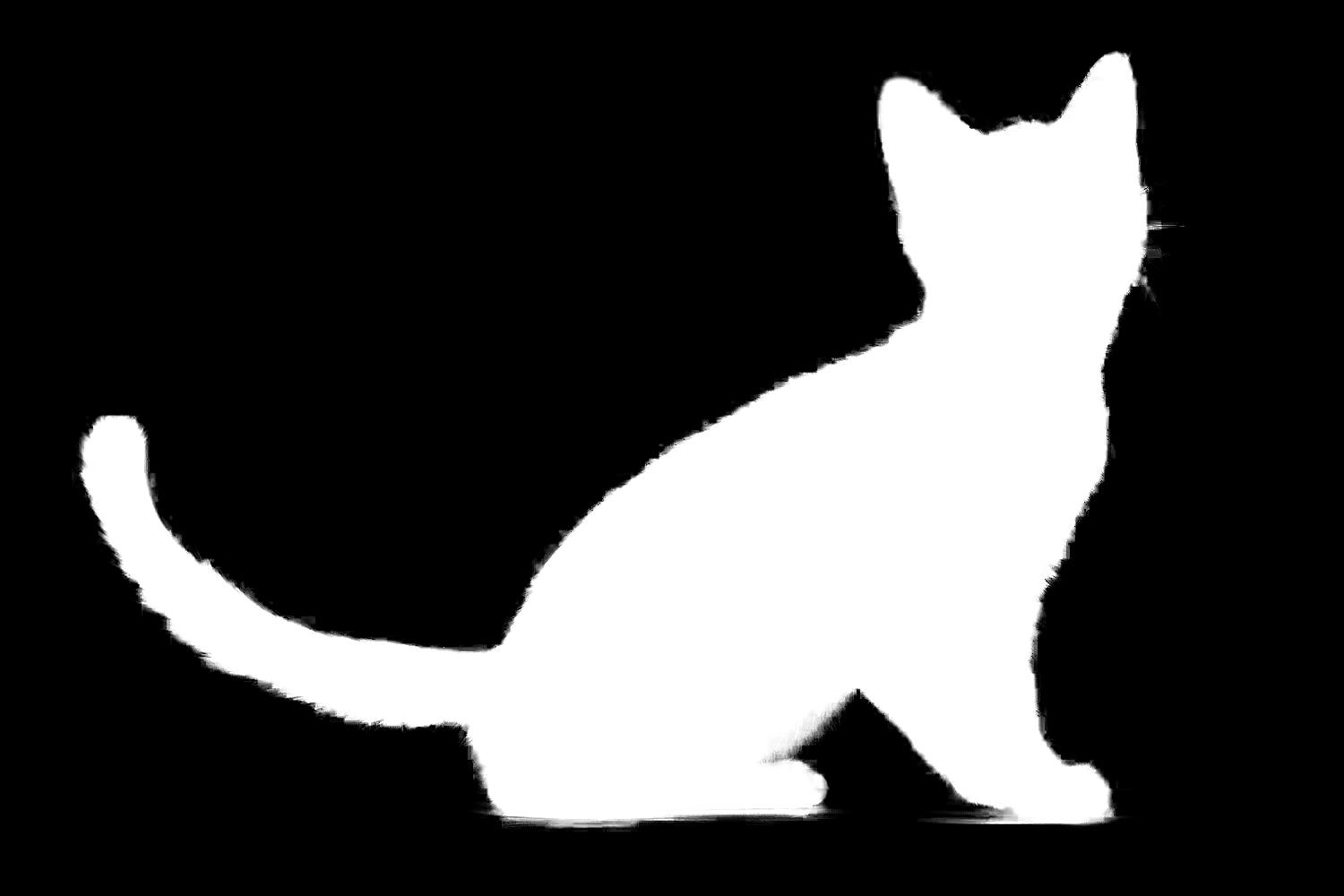} &
			\includegraphics[scale=0.04]{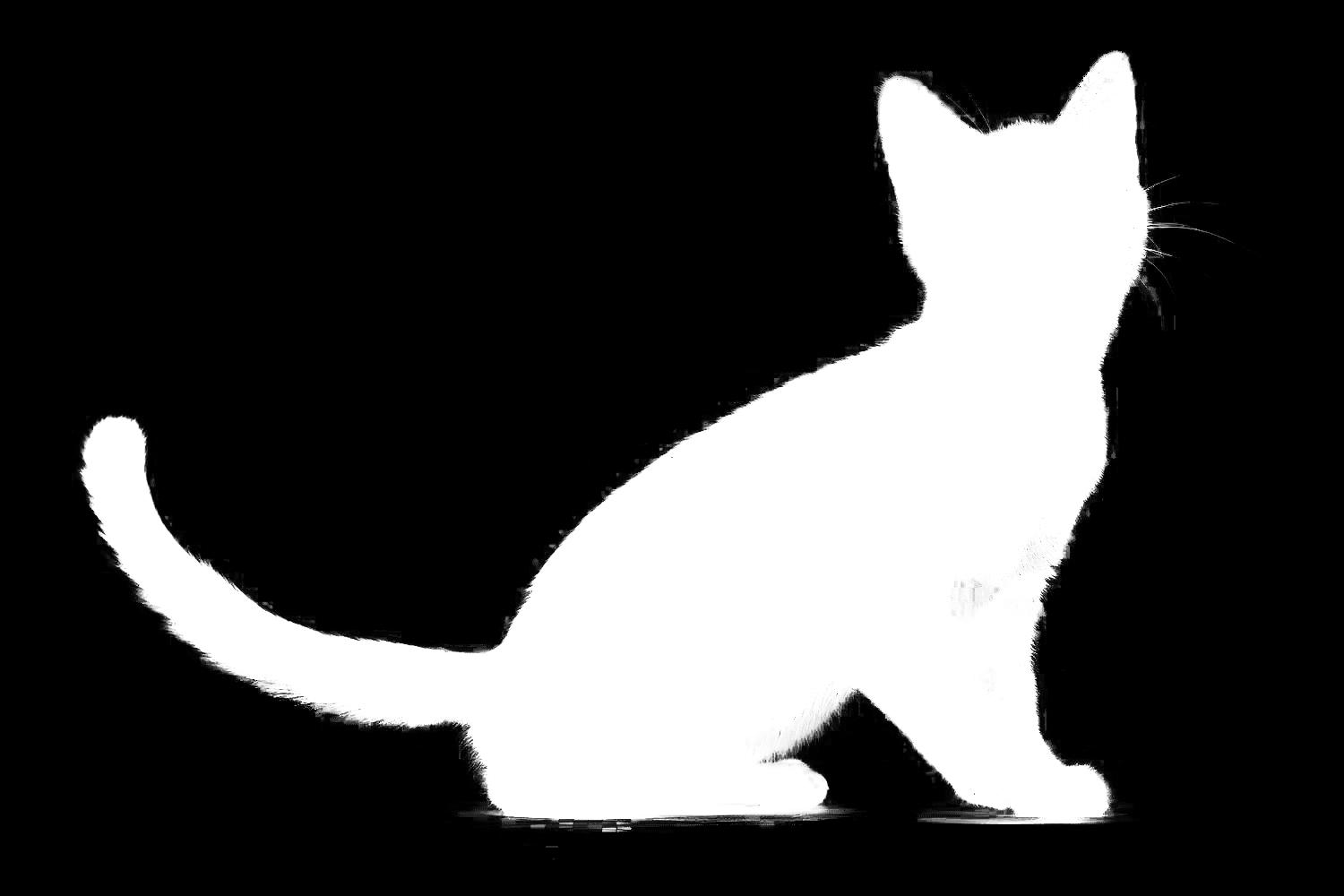} &
			\includegraphics[scale=0.04]{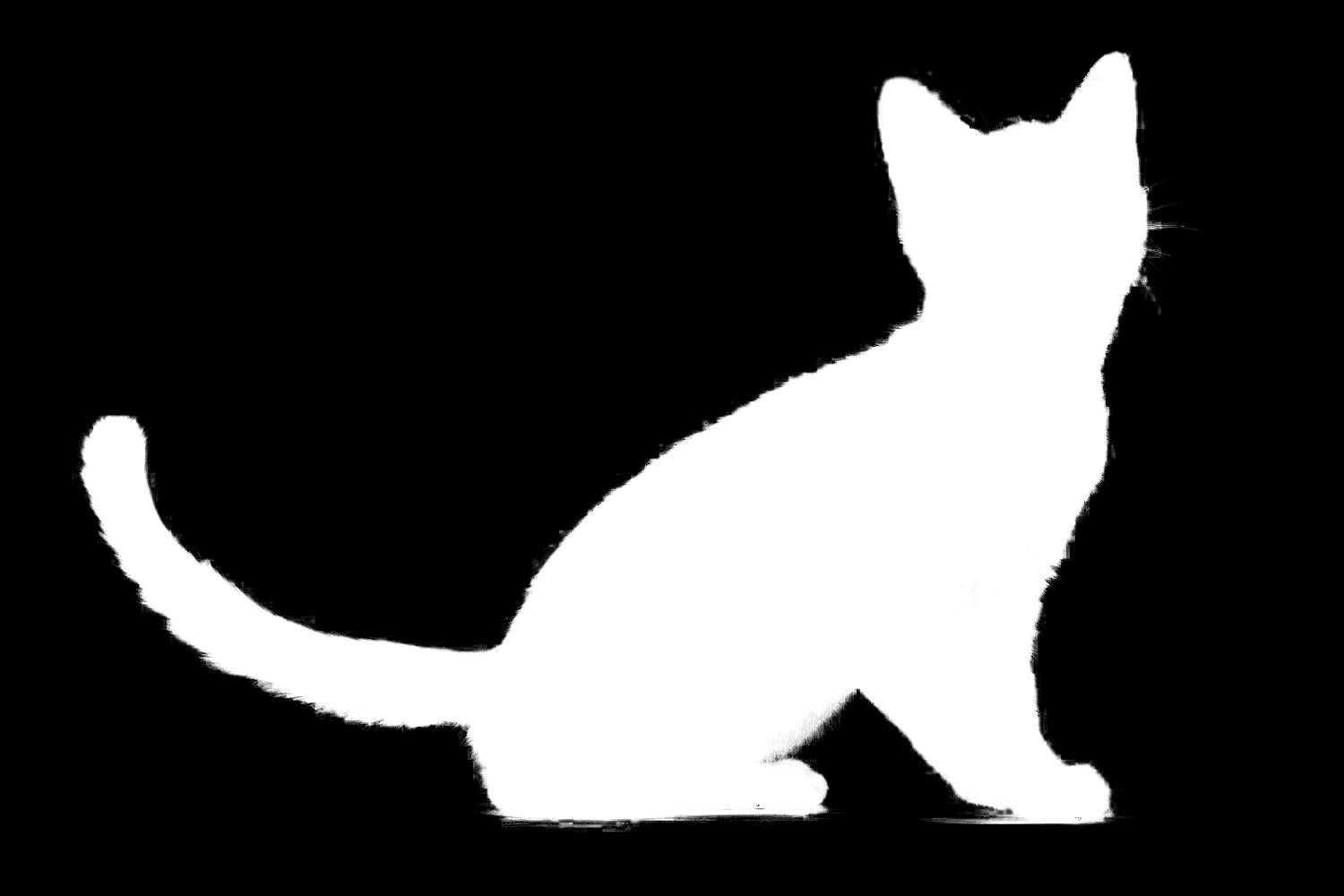} &
			\includegraphics[scale=0.04]{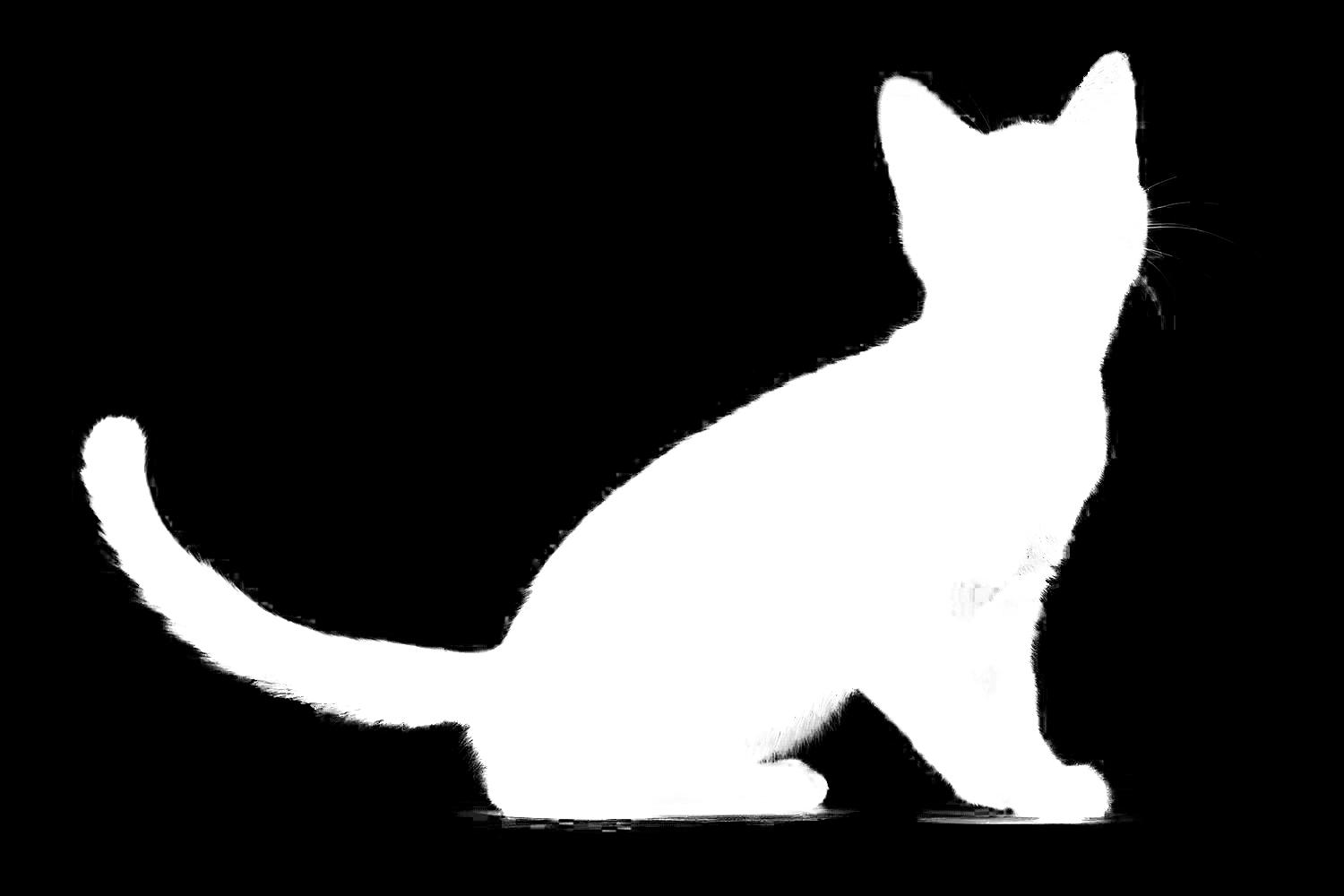} \\
			
			Input Image & Smart Scribbles & SS+Closed~\cite{Levin2007A} & SS+KNN~\cite{Chen2013KNN} & SS+DCNN~\cite{Cho2016Natural} & SS+IFM~\cite{Aksoy2017Designing} \\
			
			\includegraphics[scale=0.0533]{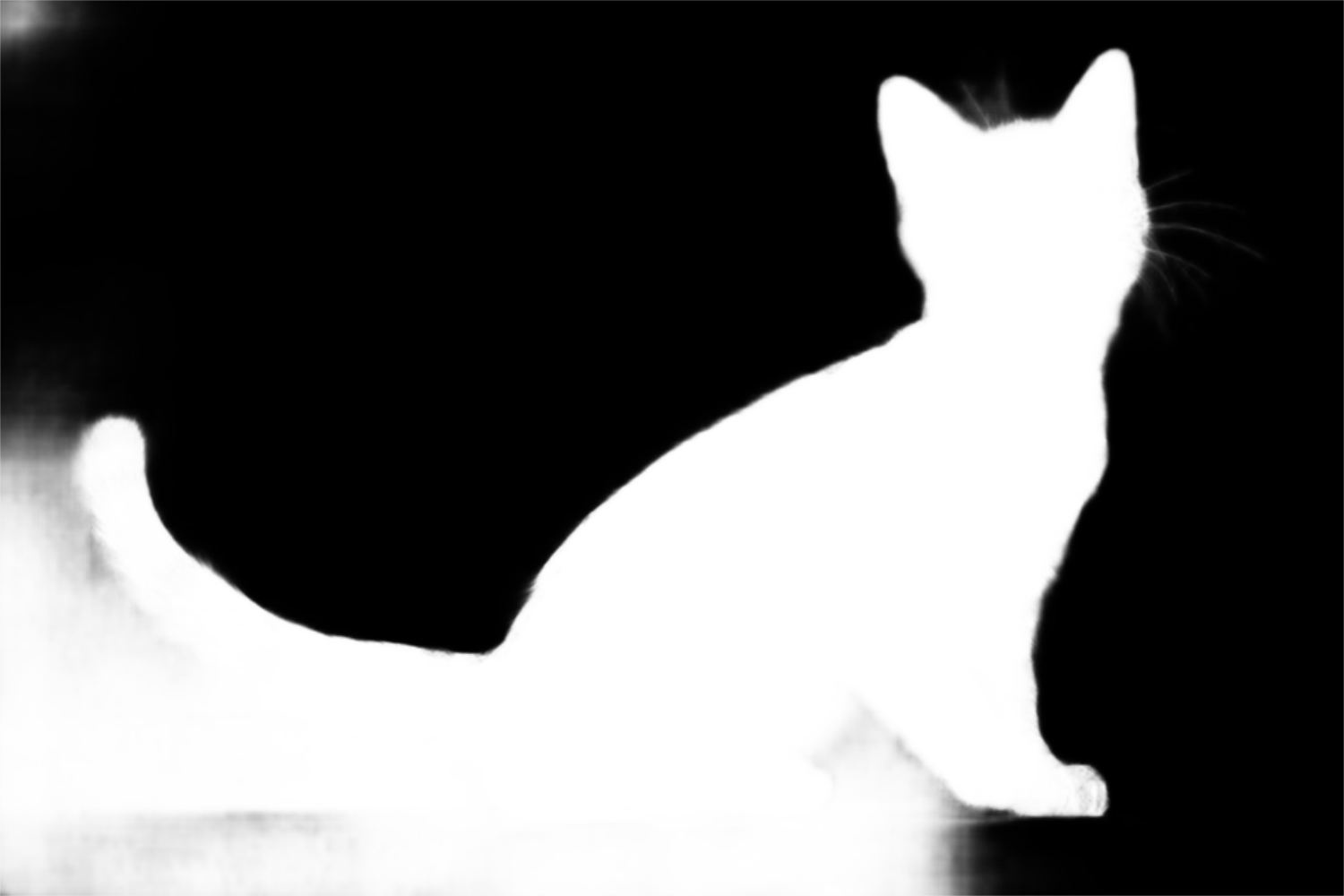} &
			\includegraphics[scale=0.054]{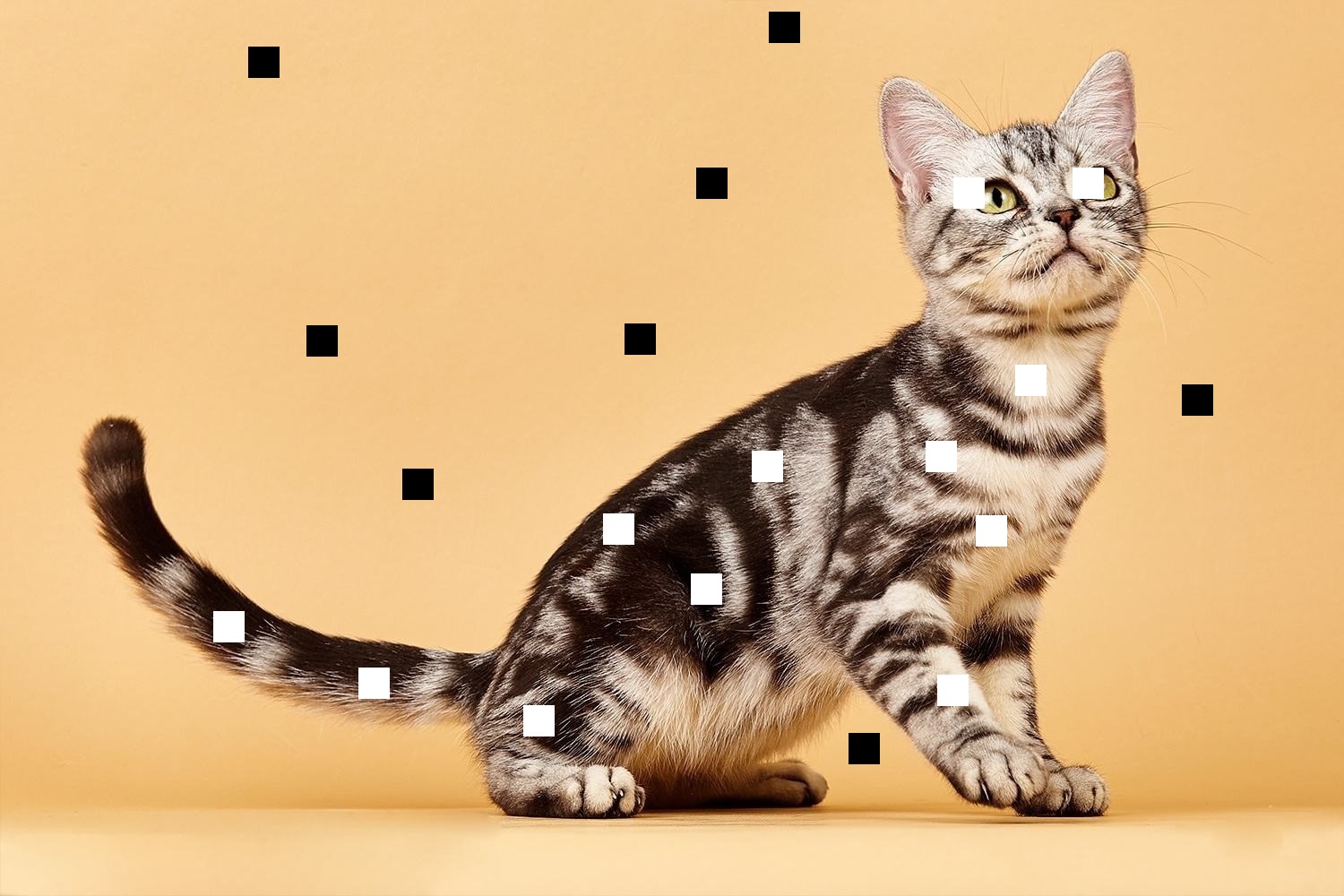} &
			\includegraphics[scale=0.04]{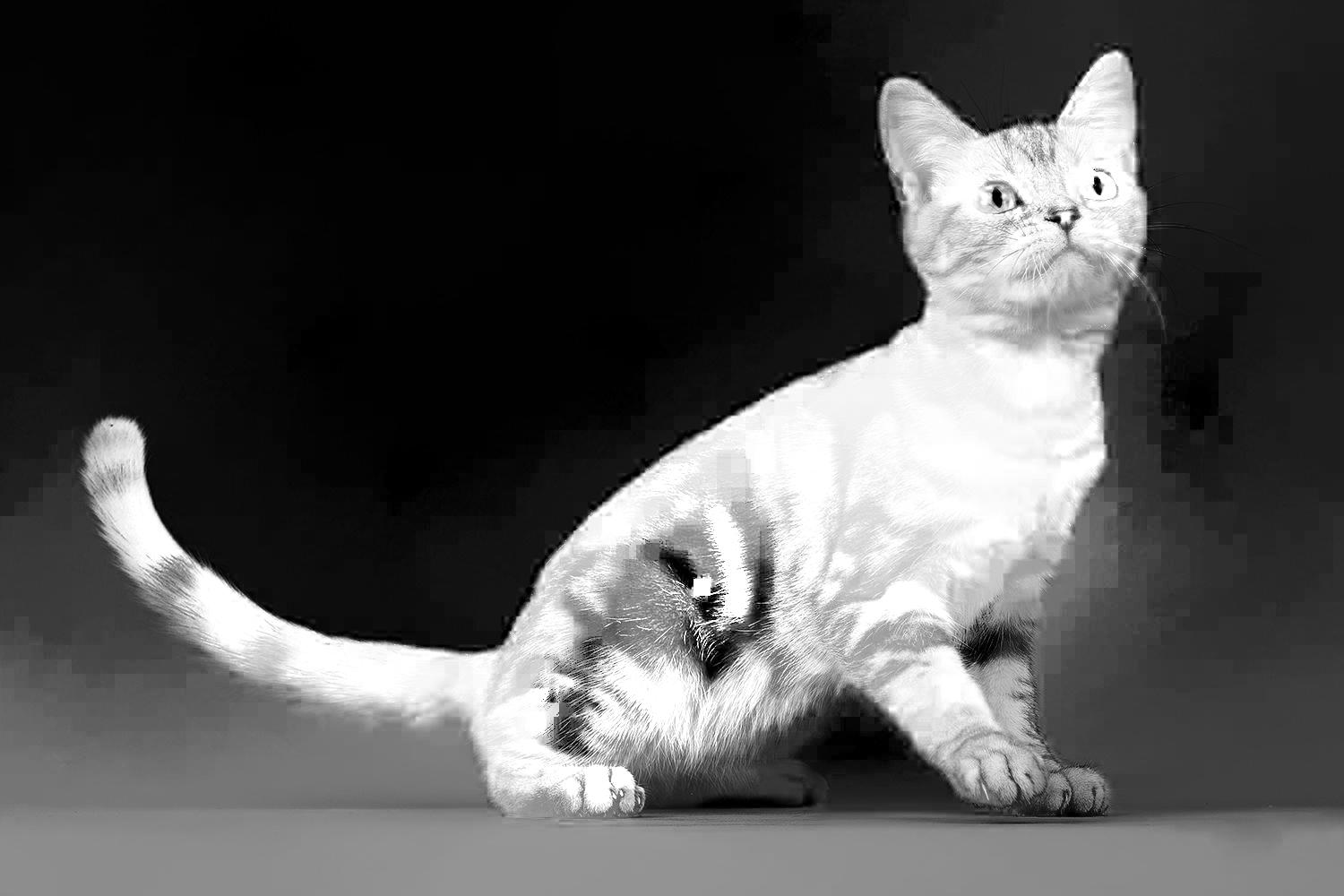} &
			\includegraphics[scale=0.04]{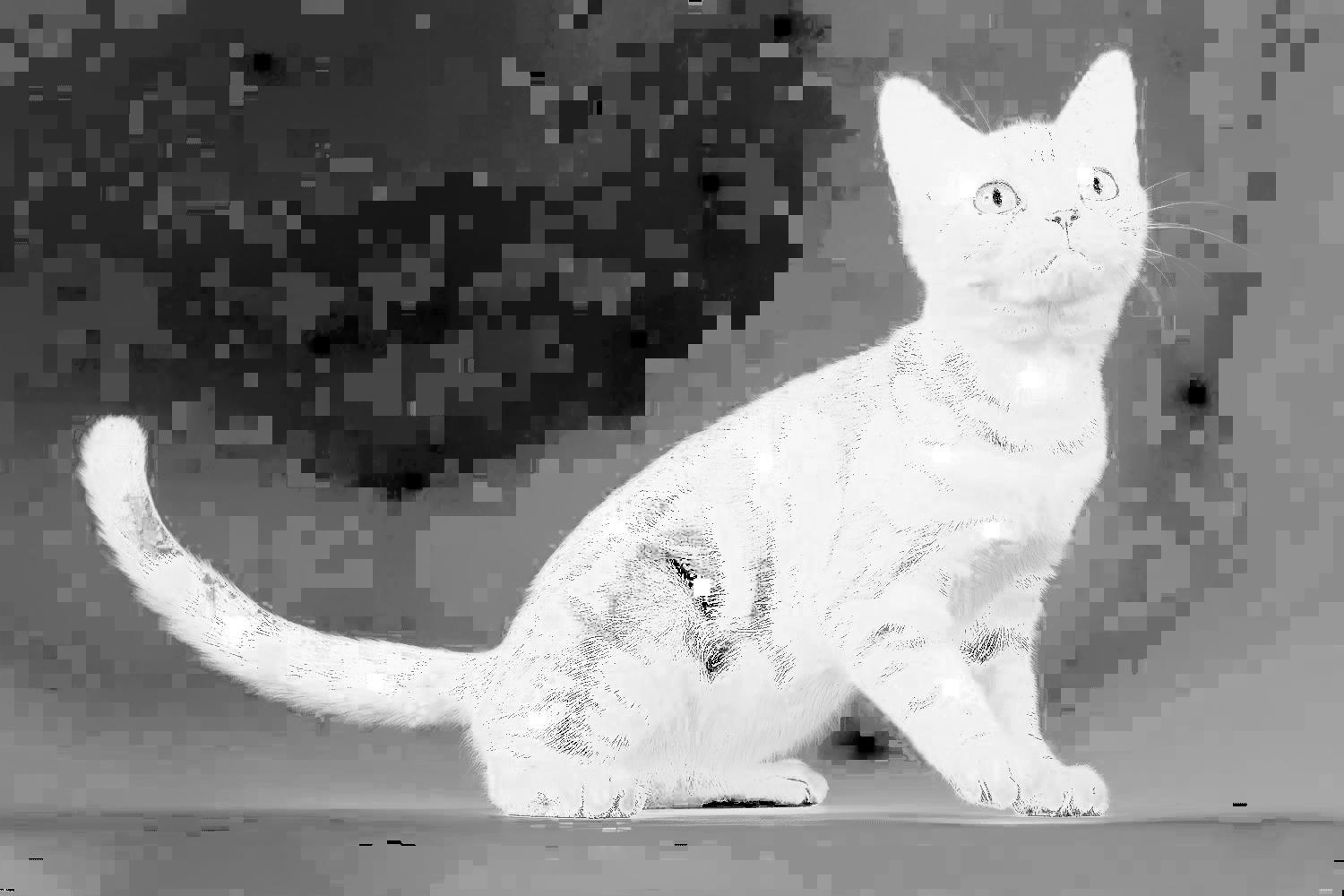} &
			\includegraphics[scale=0.04]{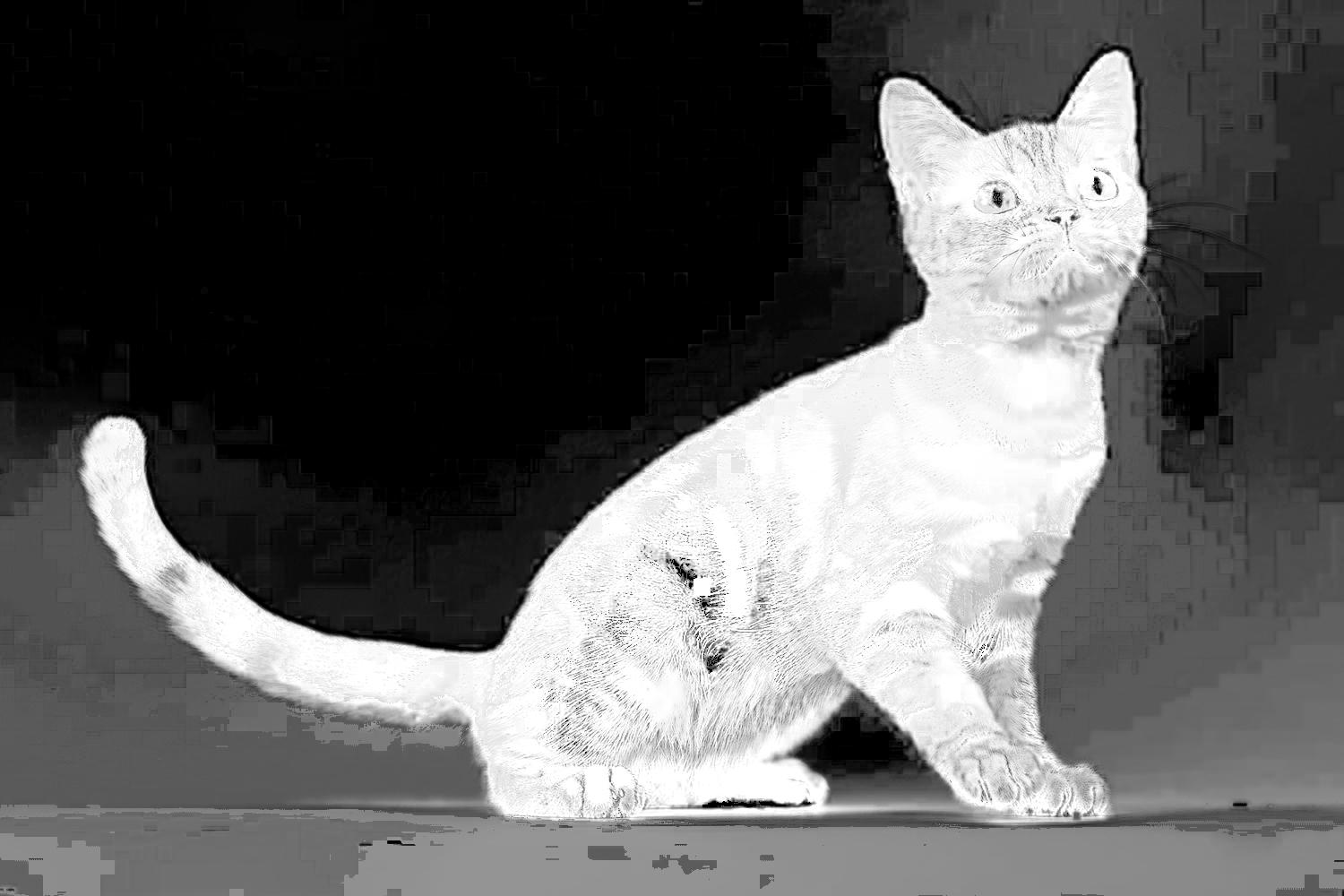} &
			\includegraphics[scale=0.04]{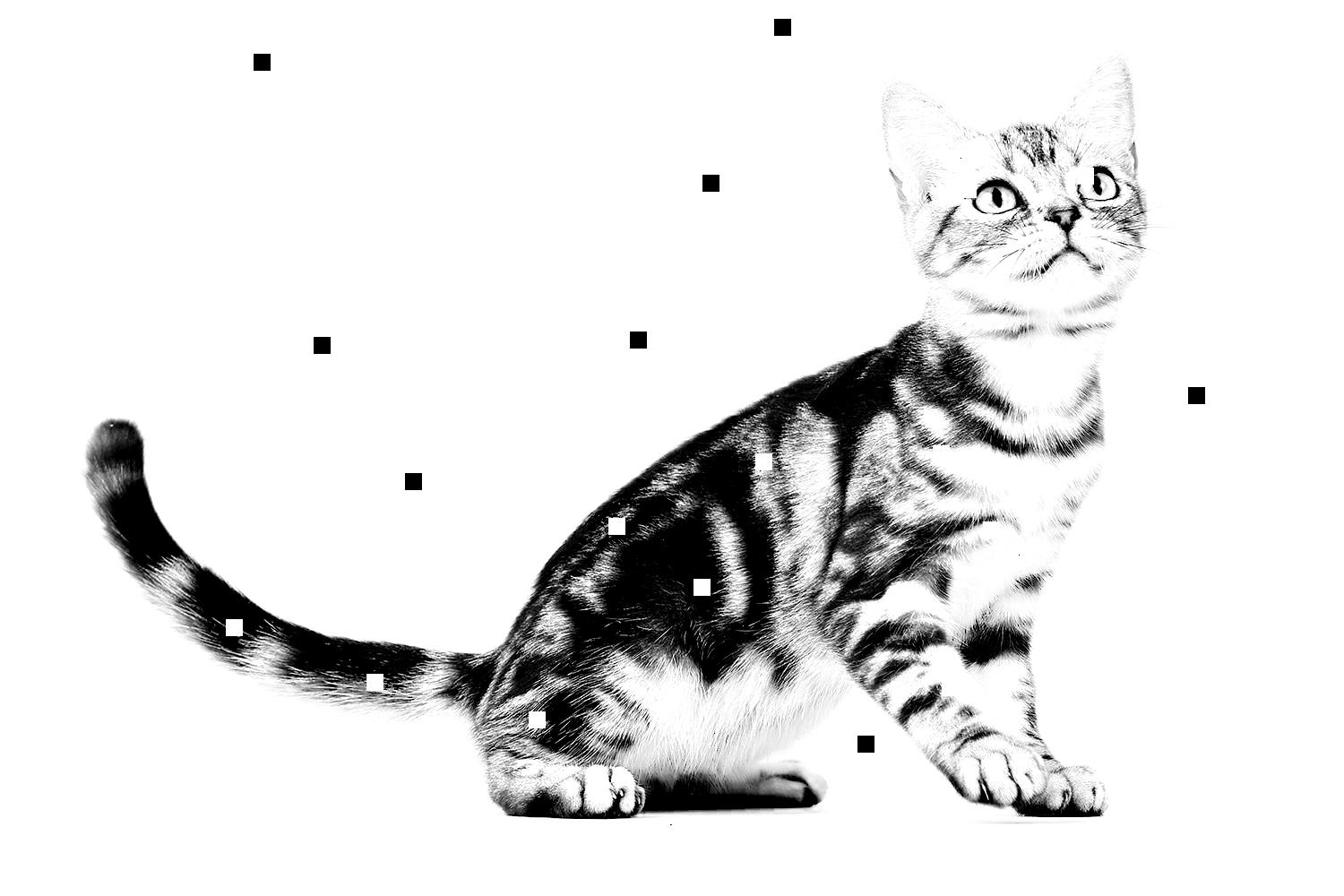} \\
			
			Late Fusion~\cite{Zhang_2019_CVPR} & AM~\cite{NIPS2018_7710} & AM+Closed~\cite{Levin2007A} & AM+KNN~\cite{Chen2013KNN} & AM+DCNN~\cite{Cho2016Natural} & AM+IFM~\cite{Aksoy2017Designing} \\
			
			\includegraphics[scale=0.036]{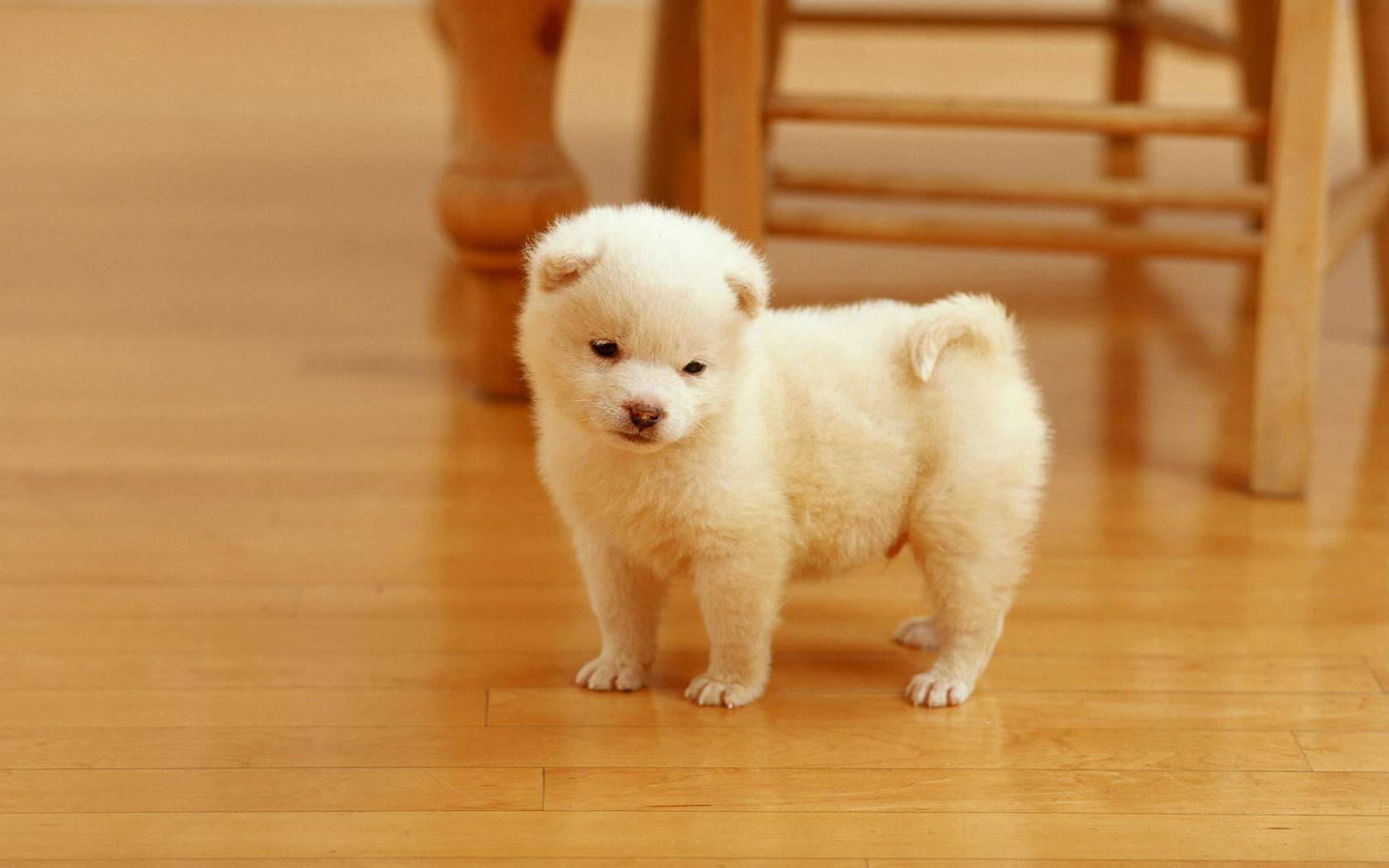} &
			\includegraphics[scale=0.048]{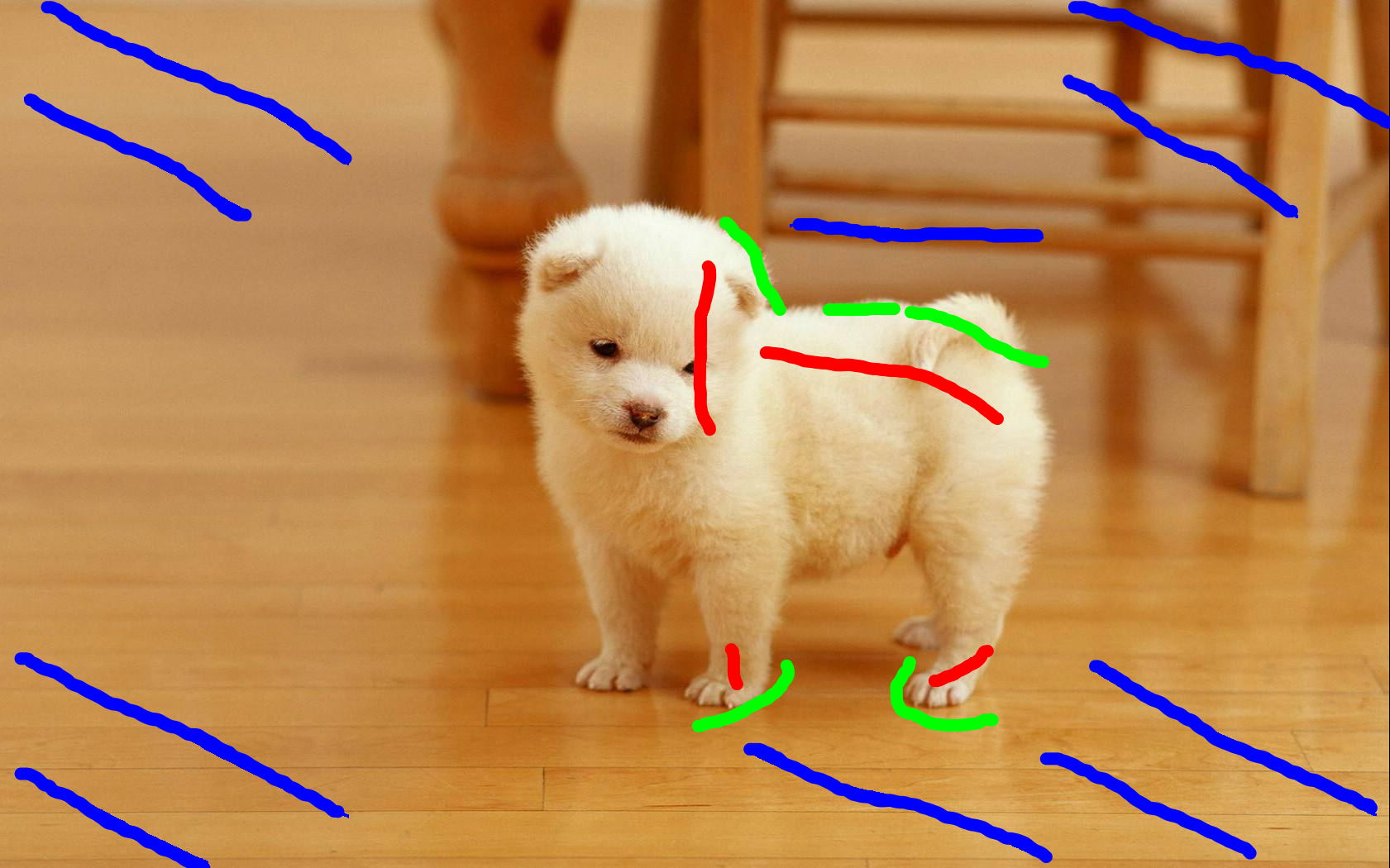} &
			\includegraphics[scale=0.036]{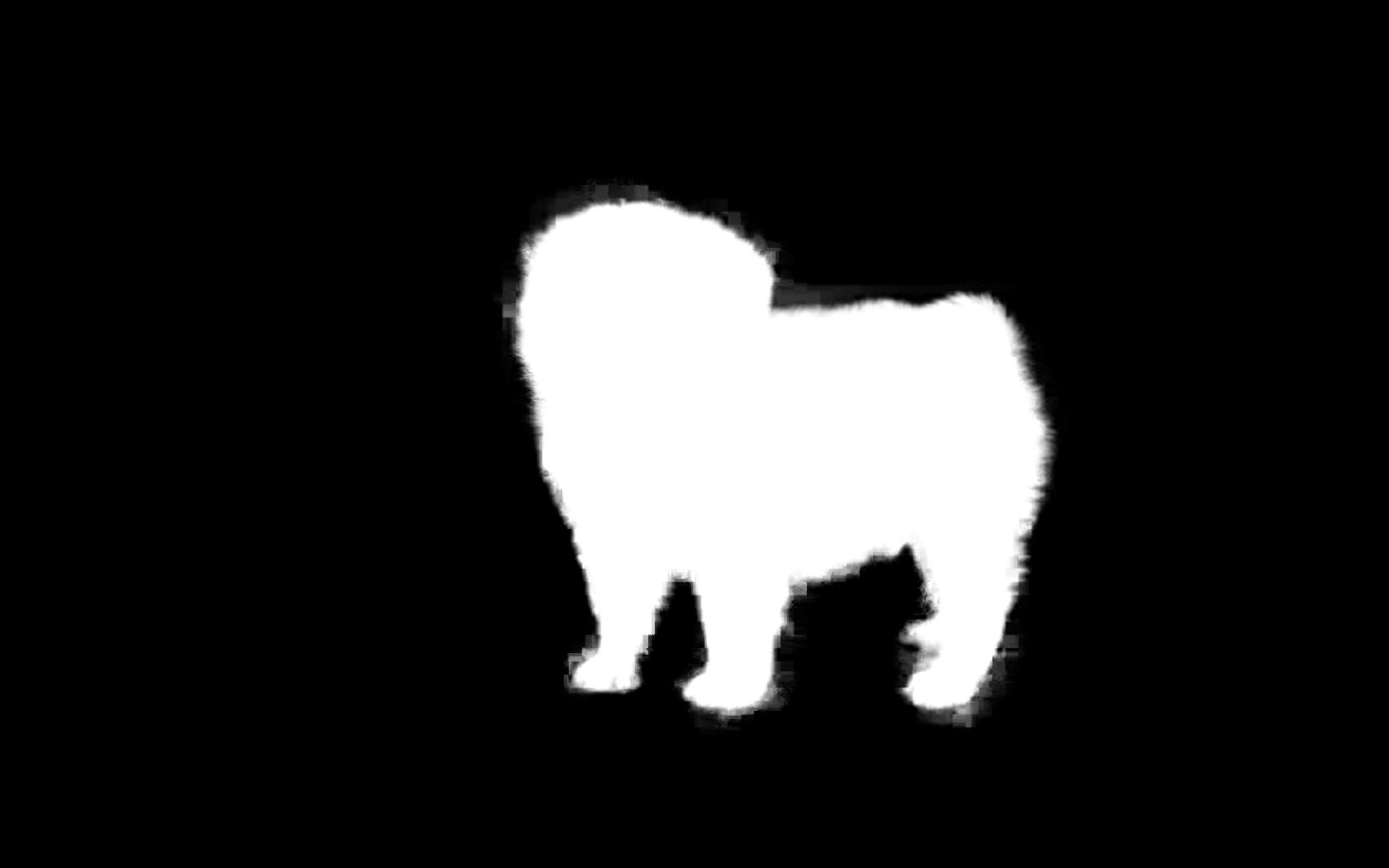} &
			\includegraphics[scale=0.036]{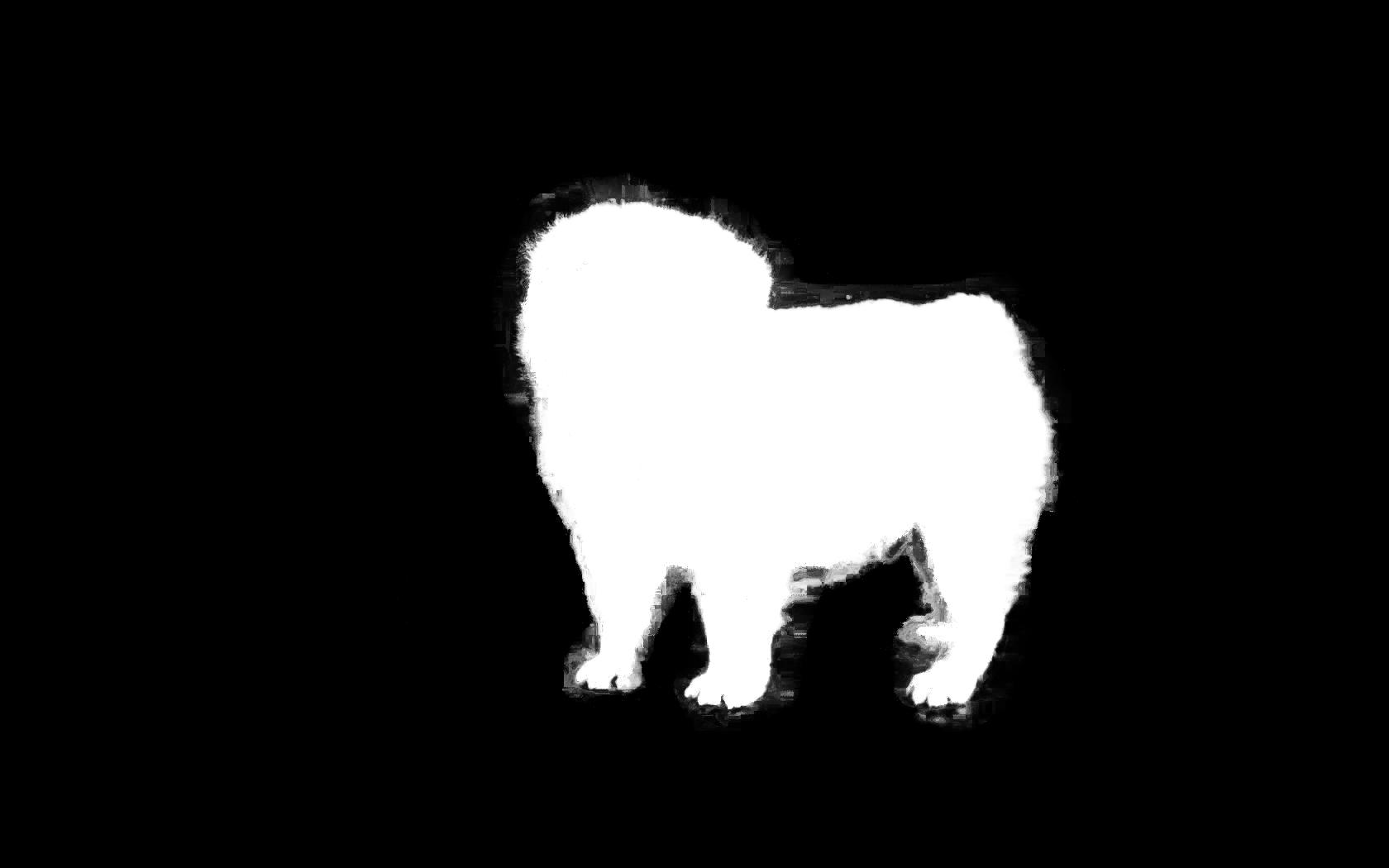} &
			\includegraphics[scale=0.036]{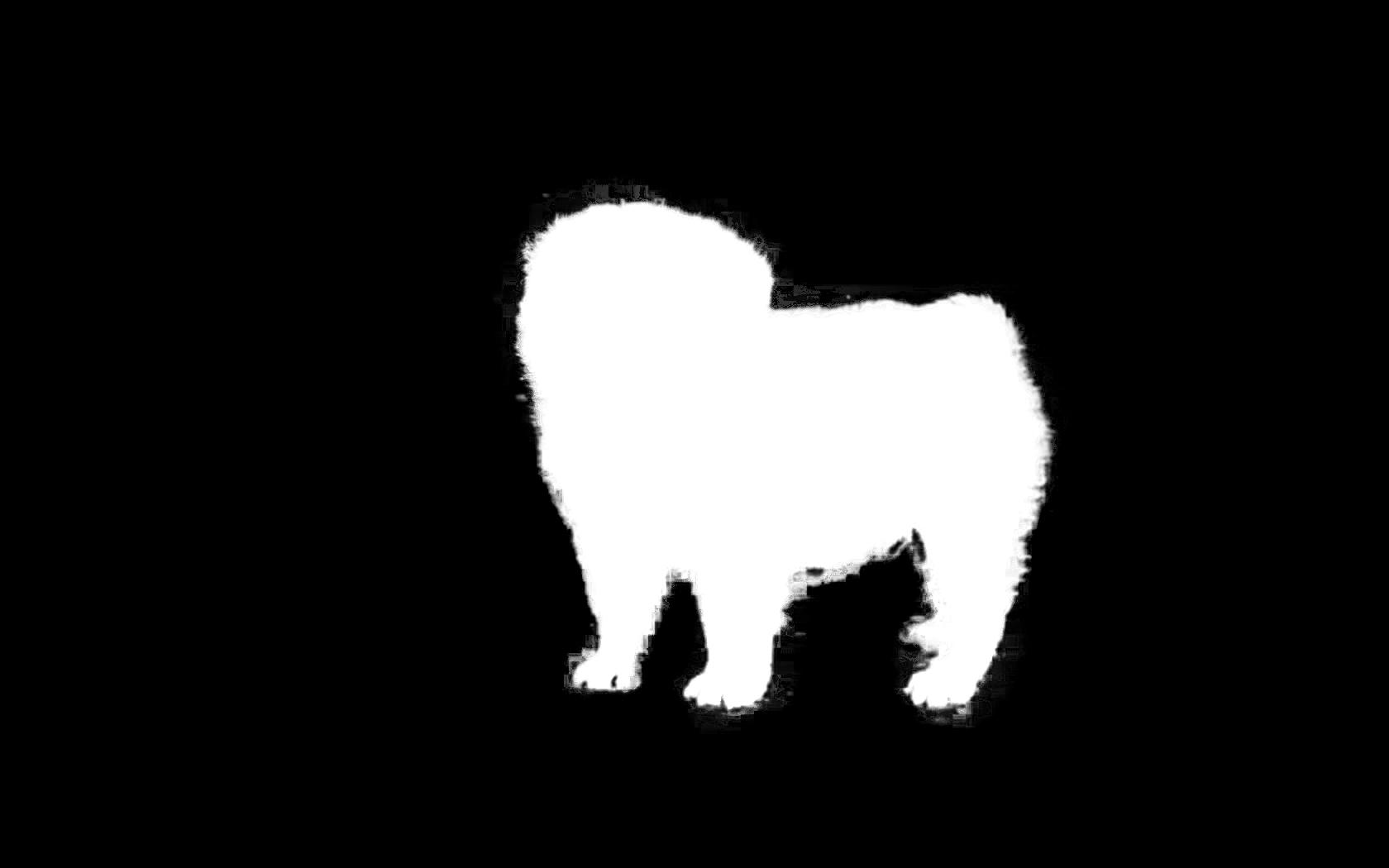} &
			\includegraphics[scale=0.048]{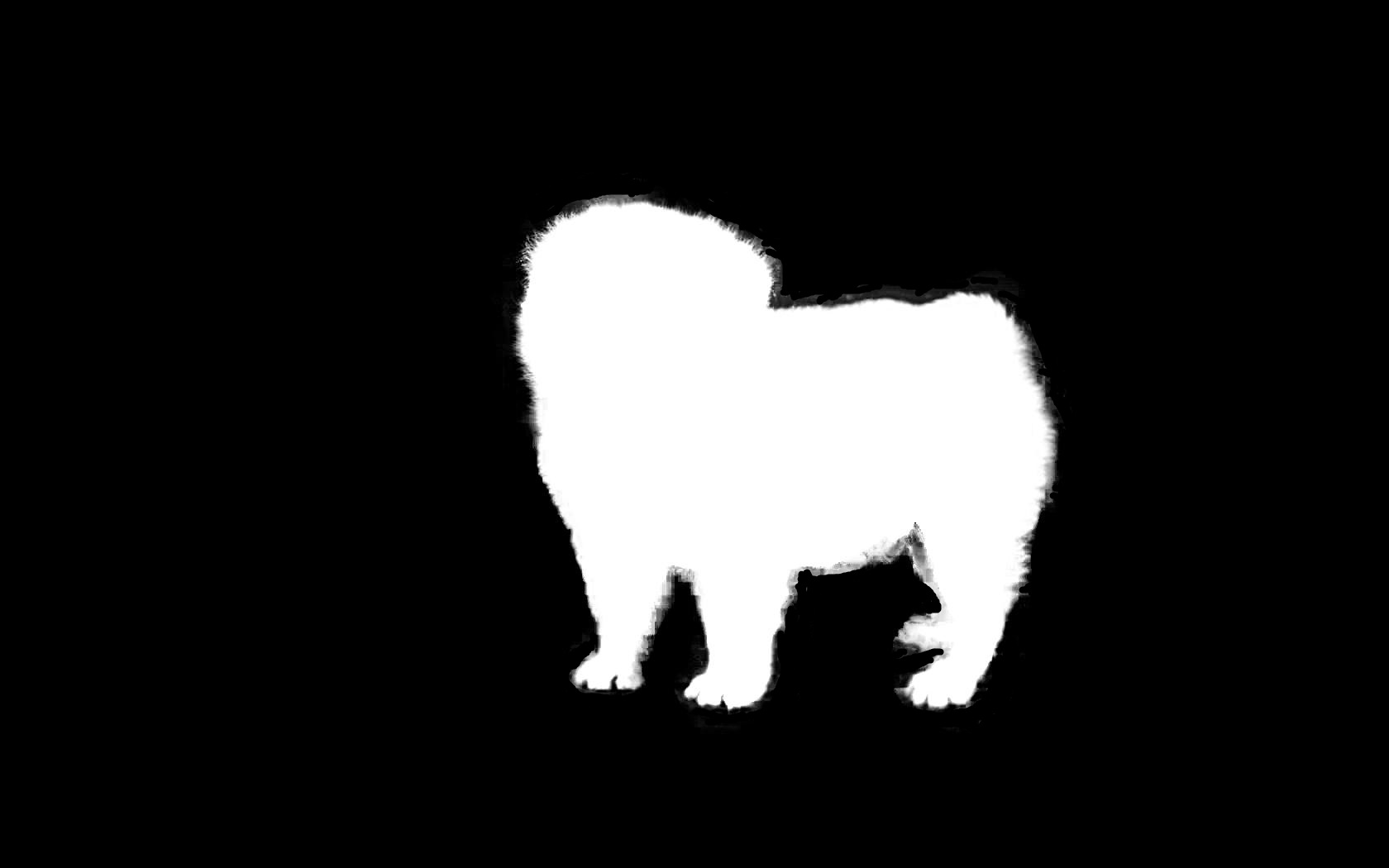} \\
			
			Input Image & Smart Scribbles & SS+Closed~\cite{Levin2007A} & SS+KNN~\cite{Chen2013KNN} & SS+DCNN~\cite{Cho2016Natural} & SS+IFM~\cite{Aksoy2017Designing} \\
			
			\includegraphics[scale=0.048]{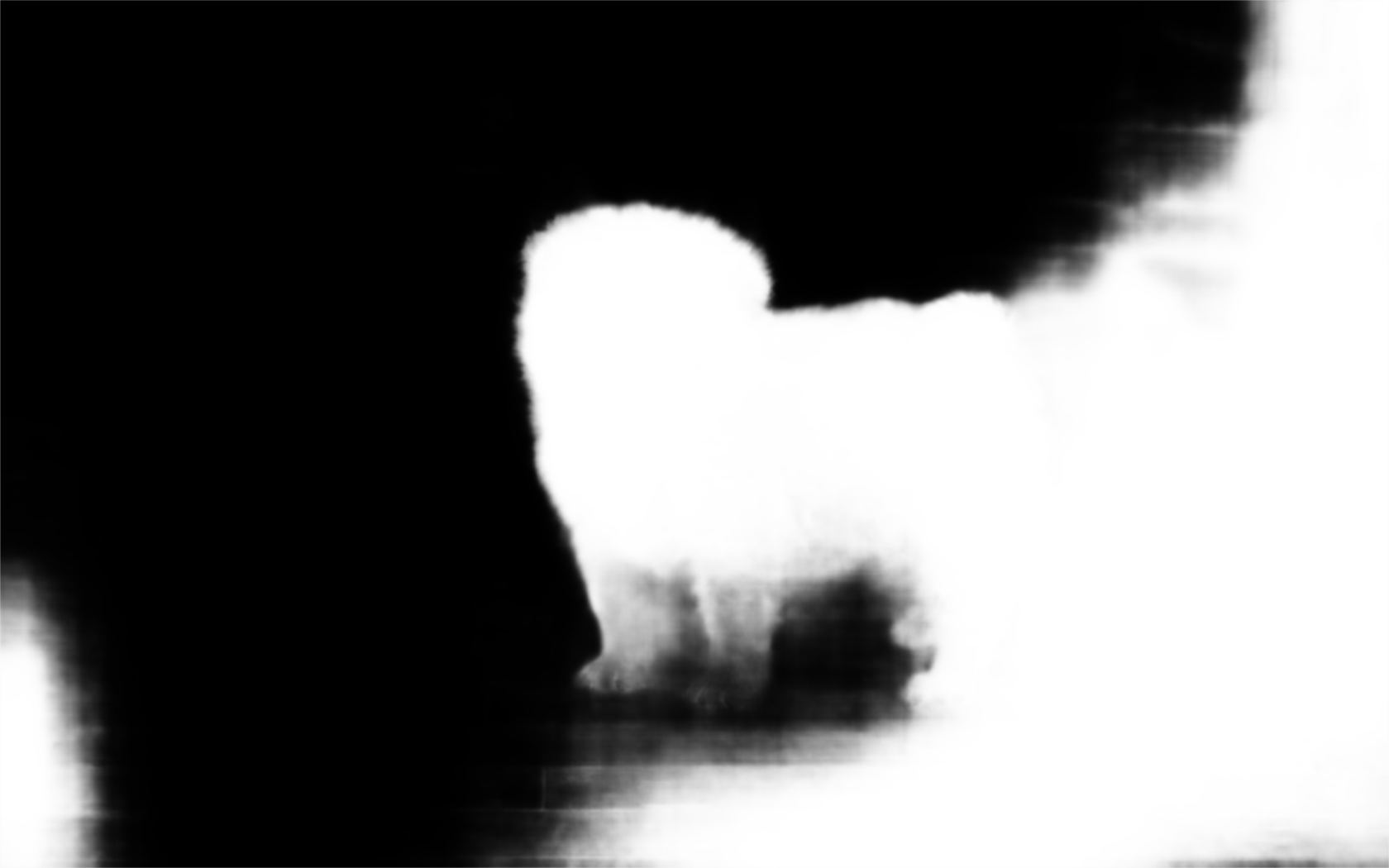} &
			\includegraphics[scale=0.036]{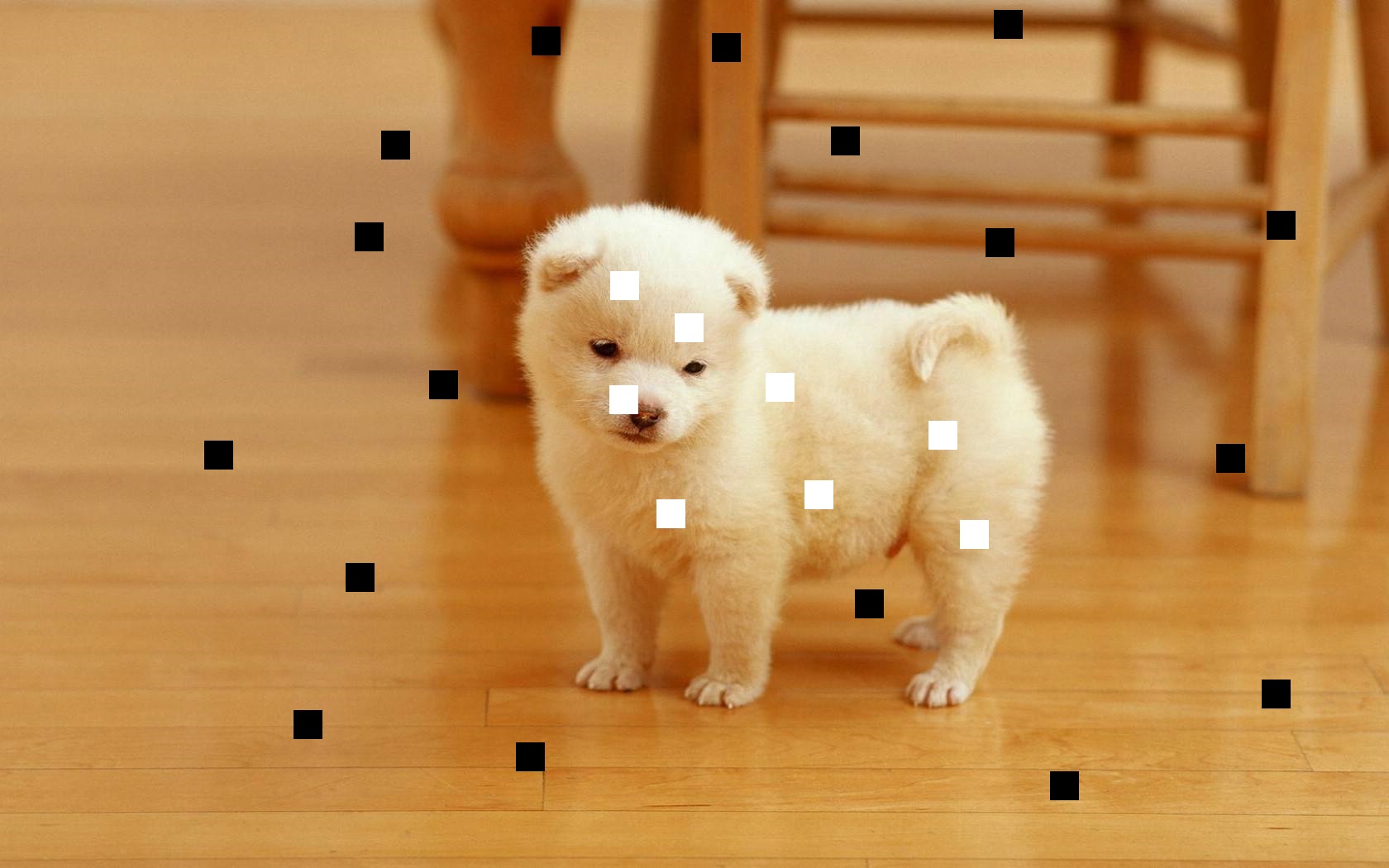} &
			\includegraphics[scale=0.036]{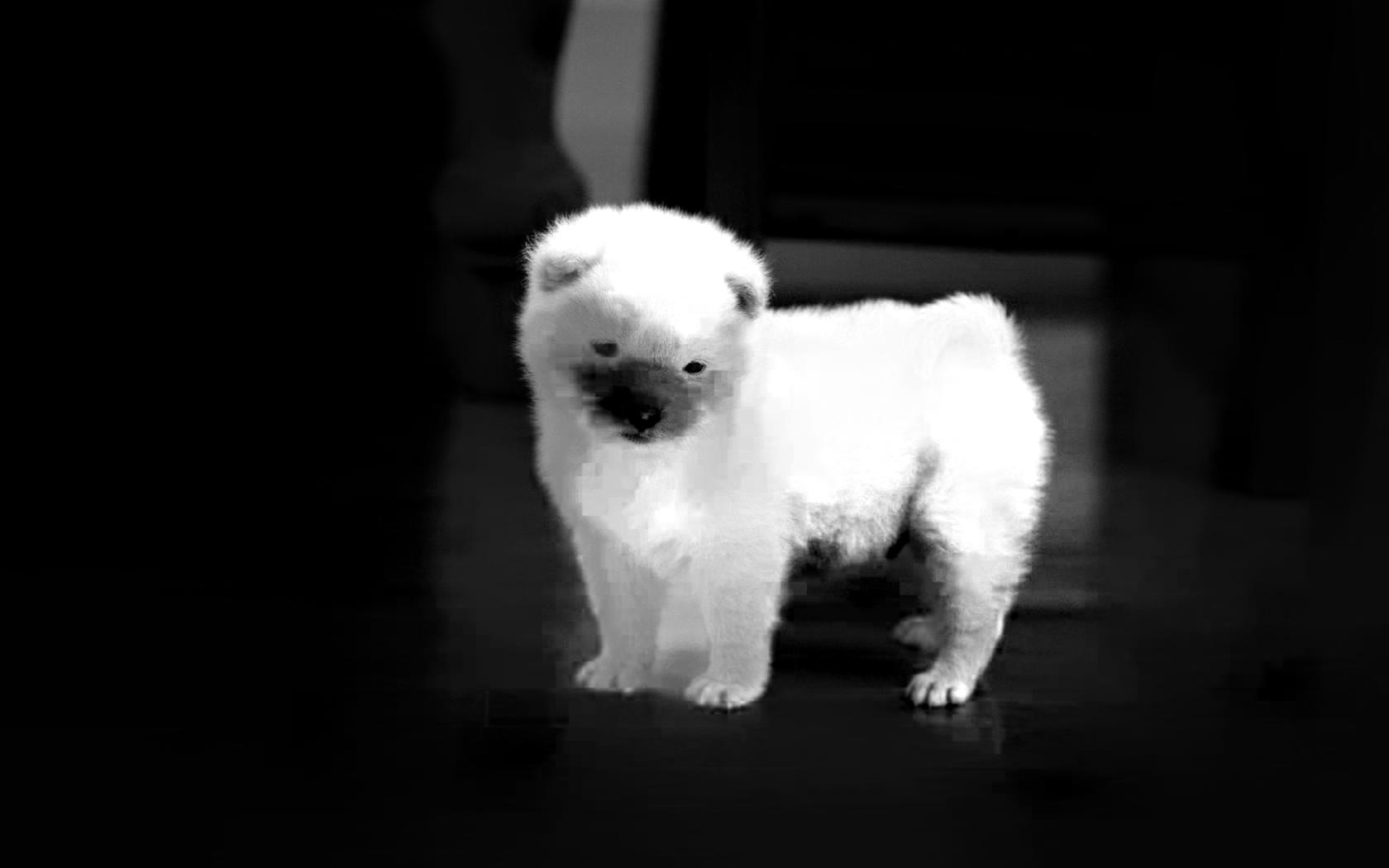} &
			\includegraphics[scale=0.036]{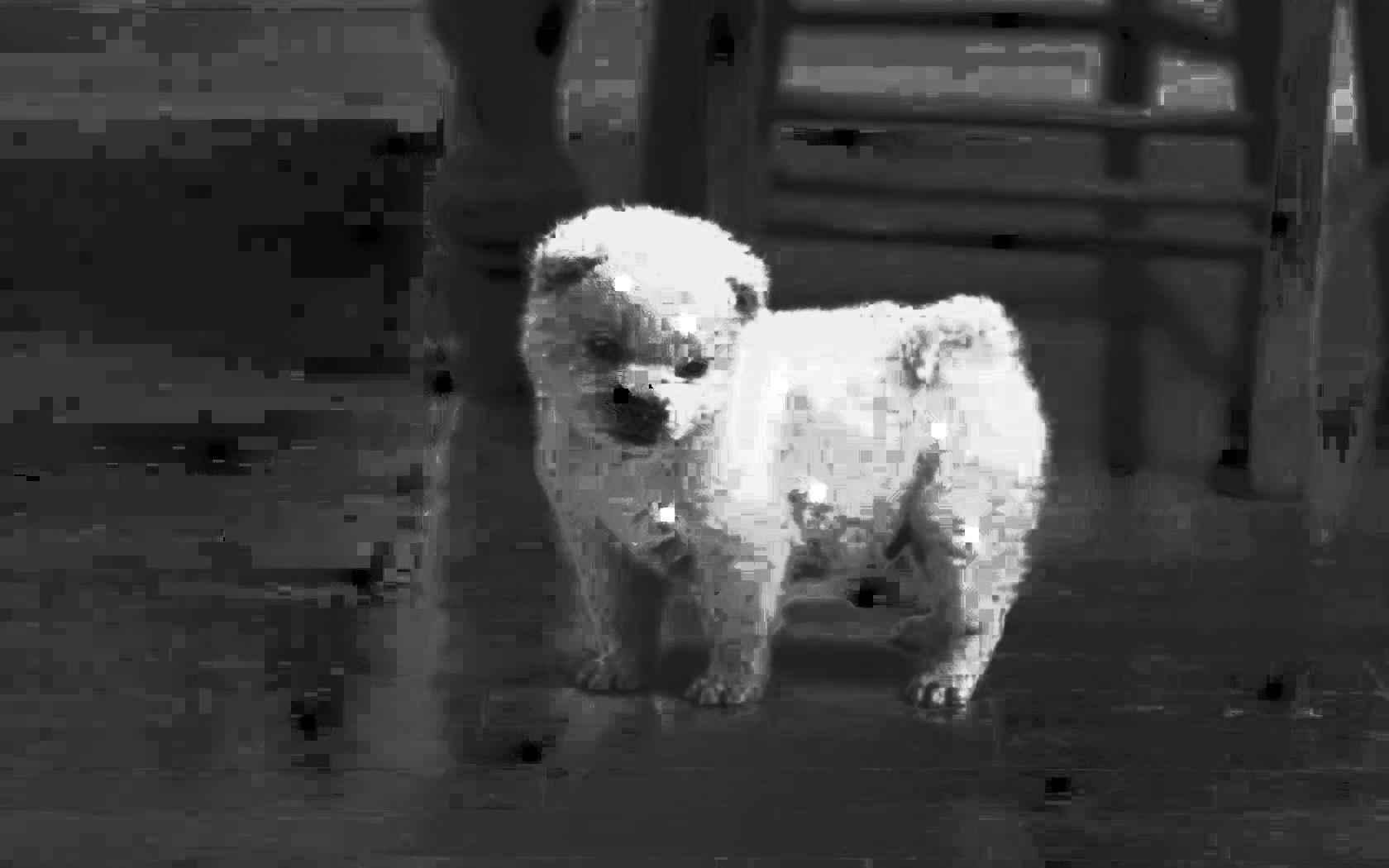} &
			\includegraphics[scale=0.036]{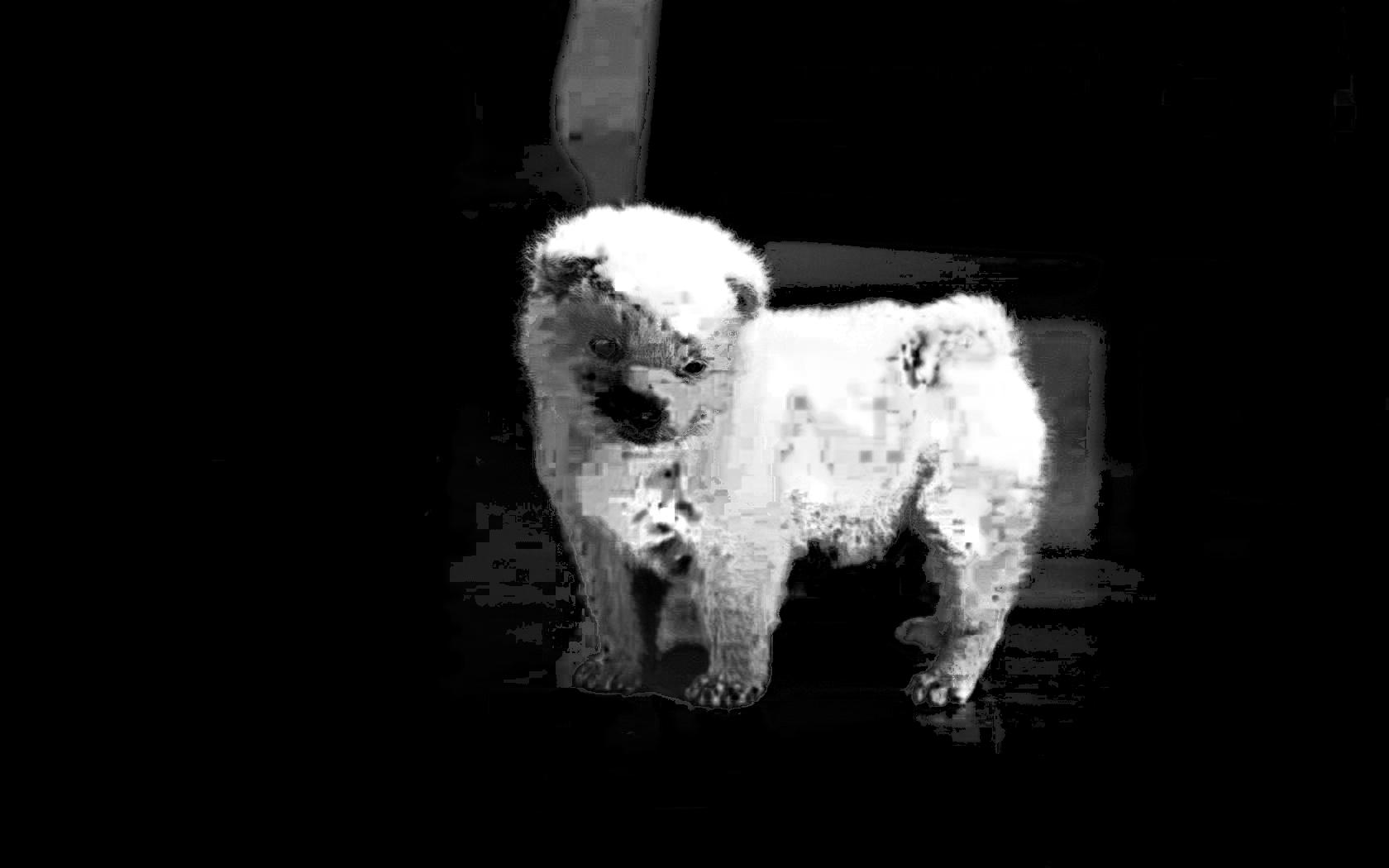} &
			\includegraphics[scale=0.036]{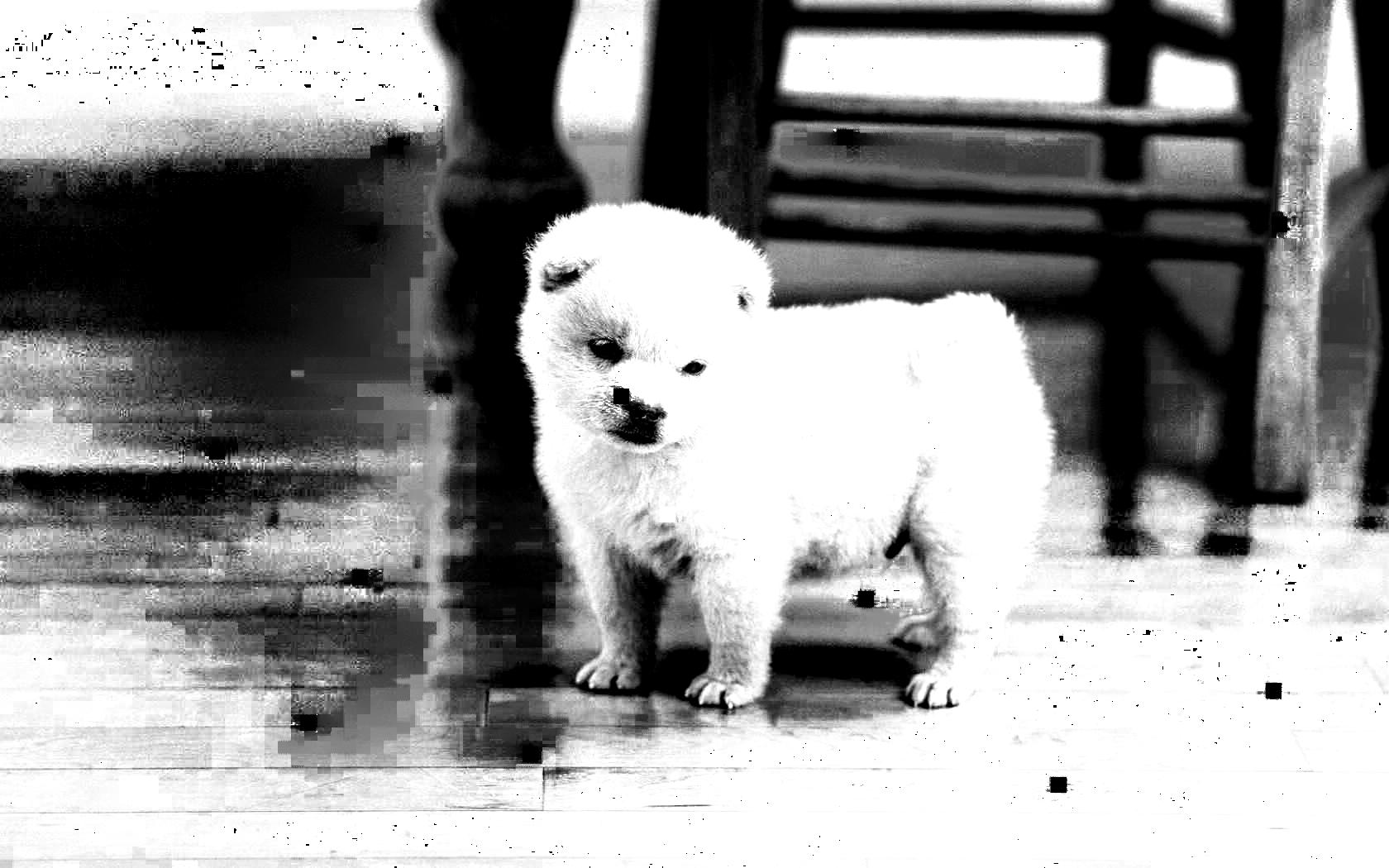} \\
			
			Late Fusion~\cite{Zhang_2019_CVPR} & AM~\cite{NIPS2018_7710} & AM+Closed~\cite{Levin2007A} & AM+KNN~\cite{Chen2013KNN} & AM+DCNN~\cite{Cho2016Natural} & AM+IFM~\cite{Aksoy2017Designing} \\
	\end{tabular}}
	\caption{The visual comparisons with Active Matting~\cite{NIPS2018_7710} and the Late Fusion~\cite{Zhang_2019_CVPR} on the real-world images. }
	\vspace{-3mm}
	\label{fig:visual_nips_1}
\end{figure*}

\begin{figure*}[t]
	\setlength{\tabcolsep}{1pt}\small{
		\begin{tabular}{cccccc}			
			\includegraphics[scale=0.101]{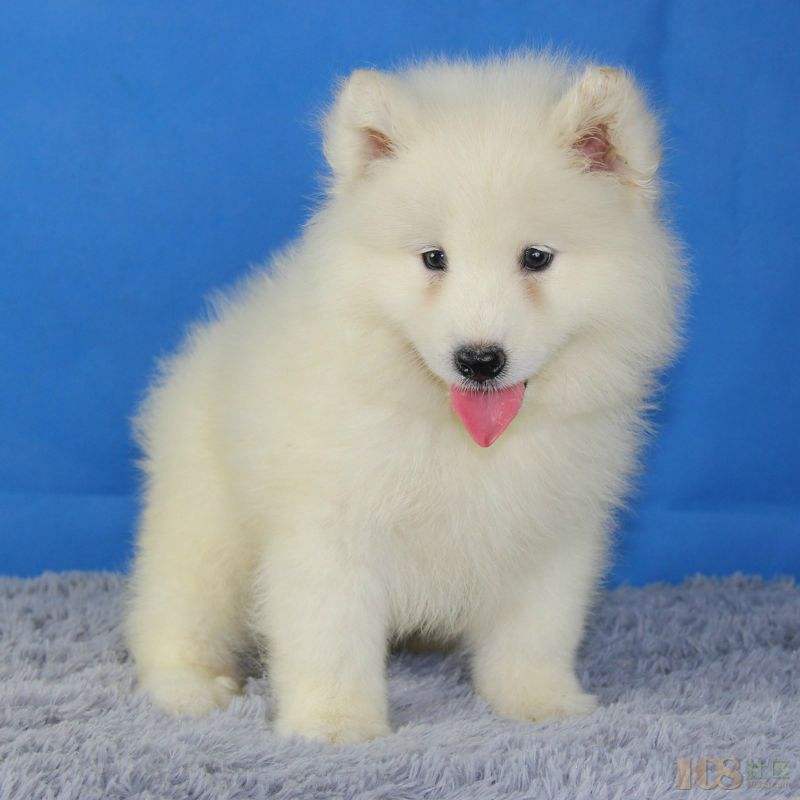} &
			\includegraphics[scale=0.101]{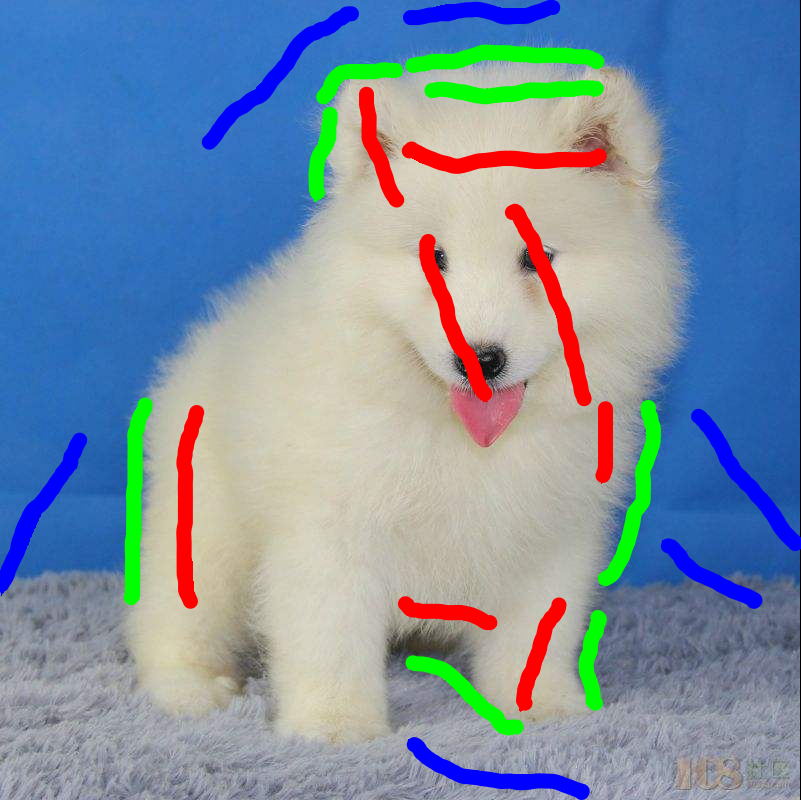} &
			\includegraphics[scale=0.076]{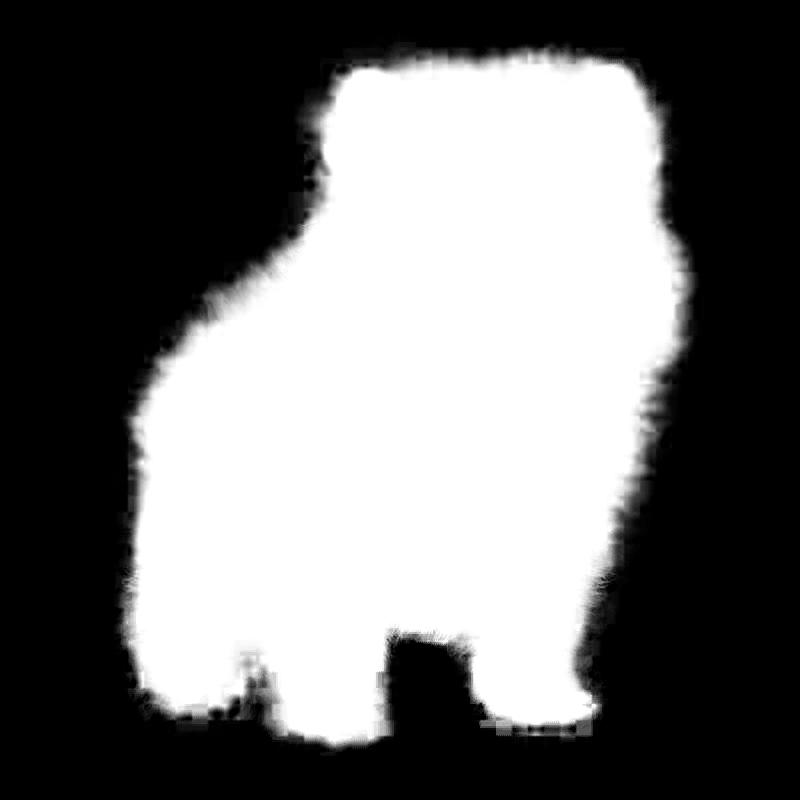} &
			\includegraphics[scale=0.076]{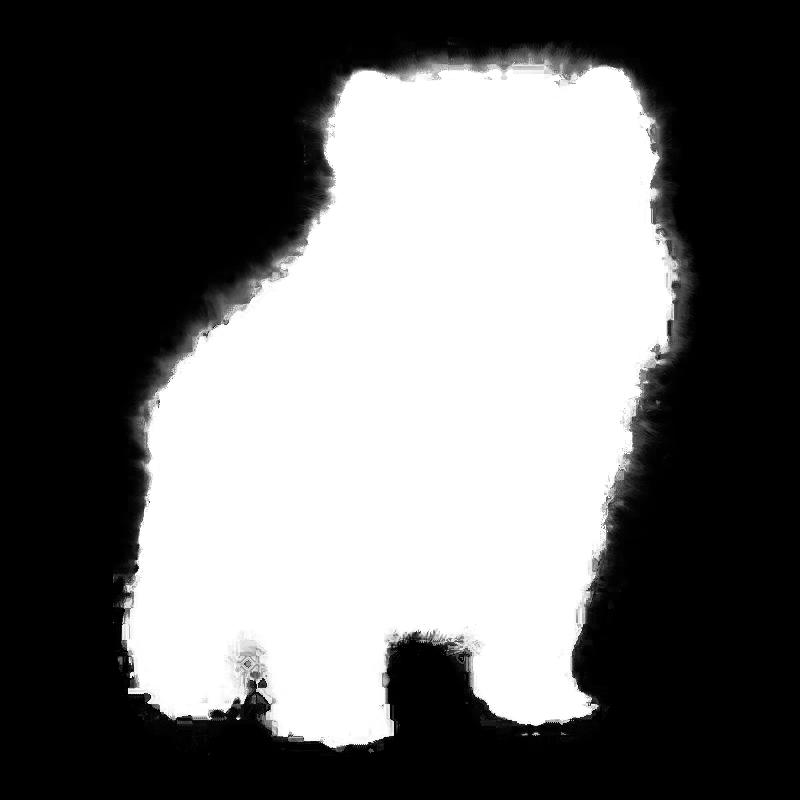} &
			\includegraphics[scale=0.076]{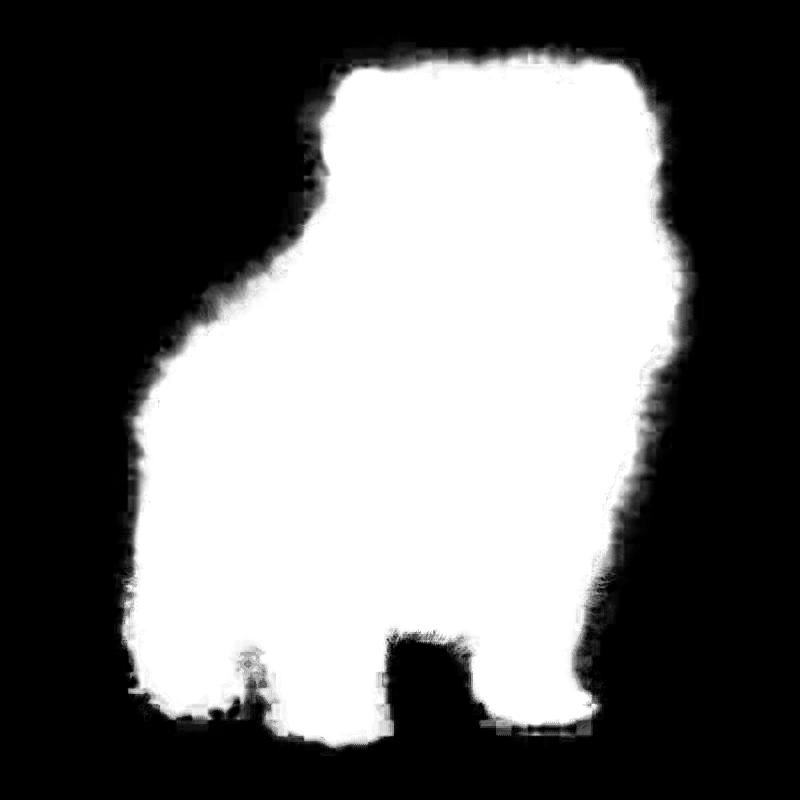} &
			\includegraphics[scale=0.076]{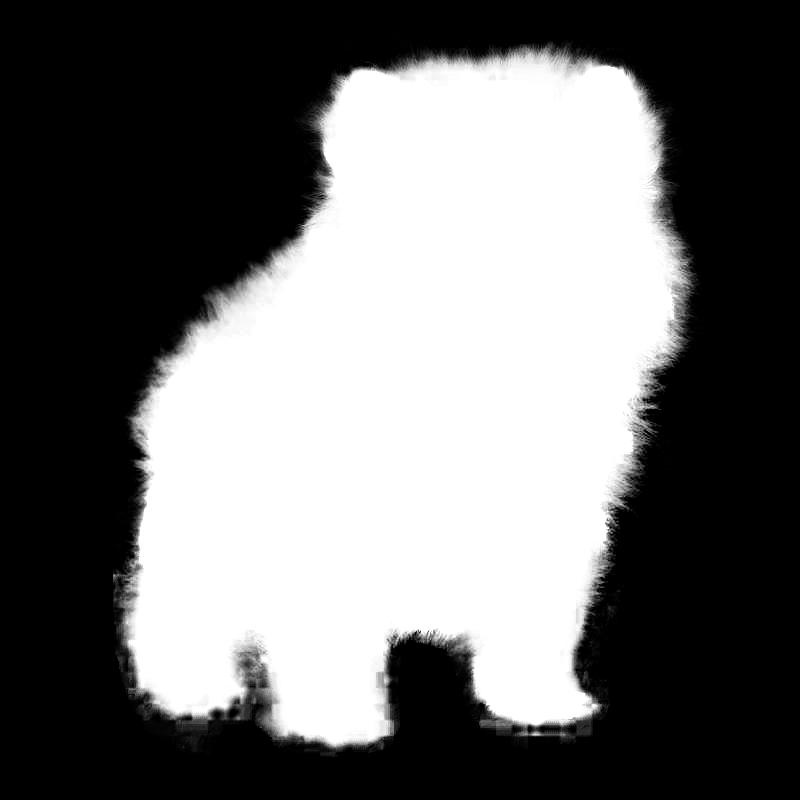} \\
			
			Input Image & Smart Scribbles & SS+Closed~\cite{Levin2007A} & SS+KNN~\cite{Chen2013KNN} & SS+DCNN~\cite{Cho2016Natural} & SS+IFM~\cite{Aksoy2017Designing} \\
			
			\includegraphics[scale=0.101]{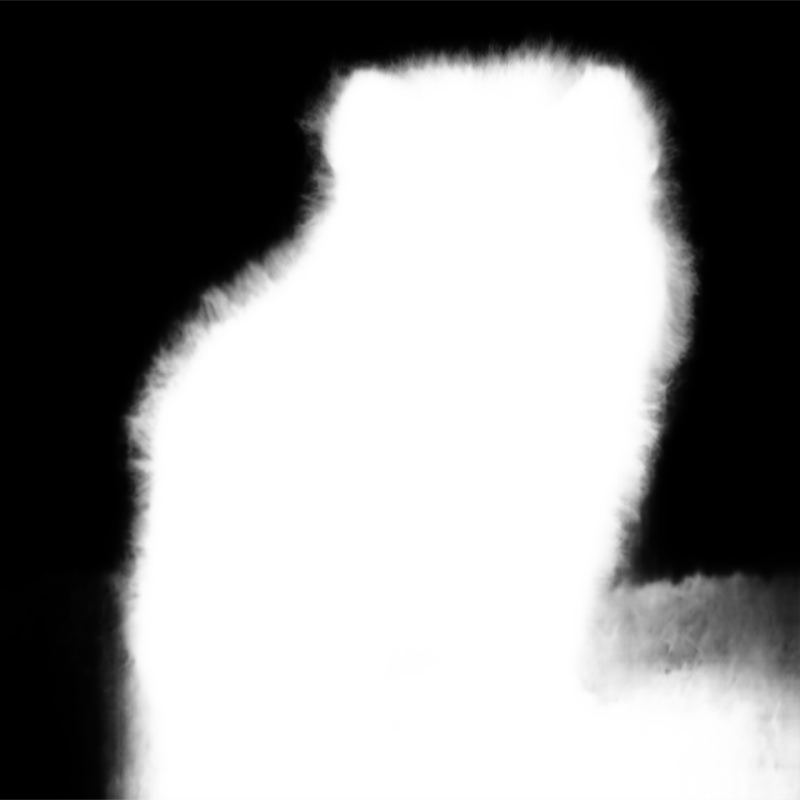} &
			\includegraphics[scale=0.101]{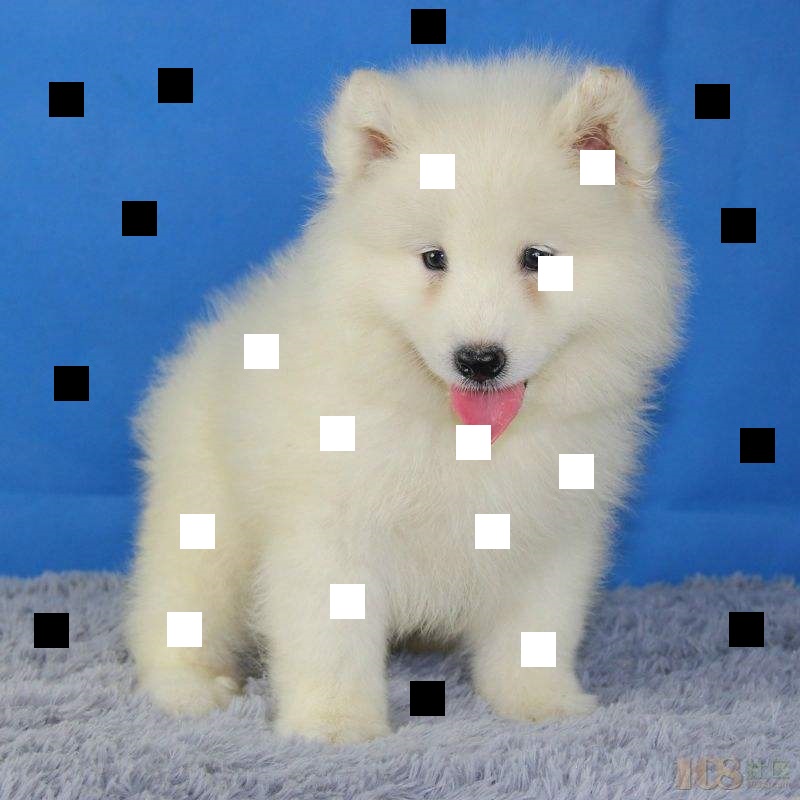} &
			\includegraphics[scale=0.076]{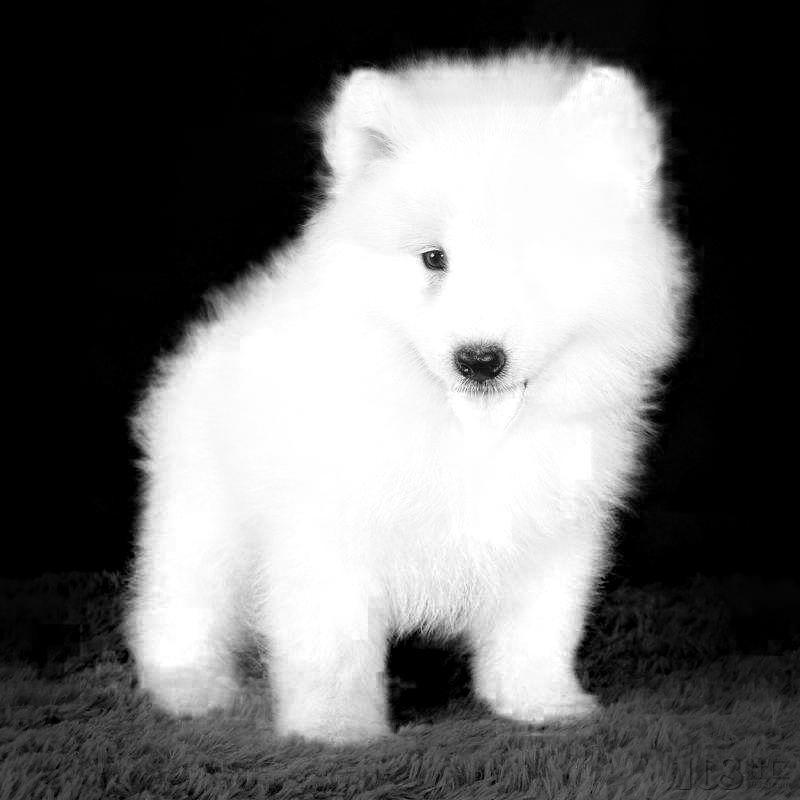} &
			\includegraphics[scale=0.076]{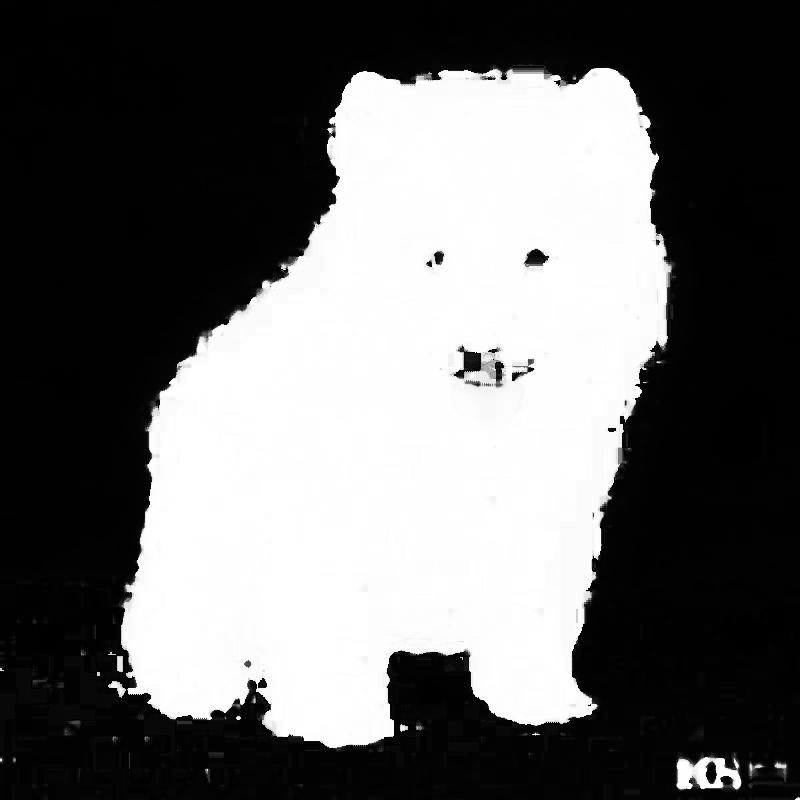} &
			\includegraphics[scale=0.076]{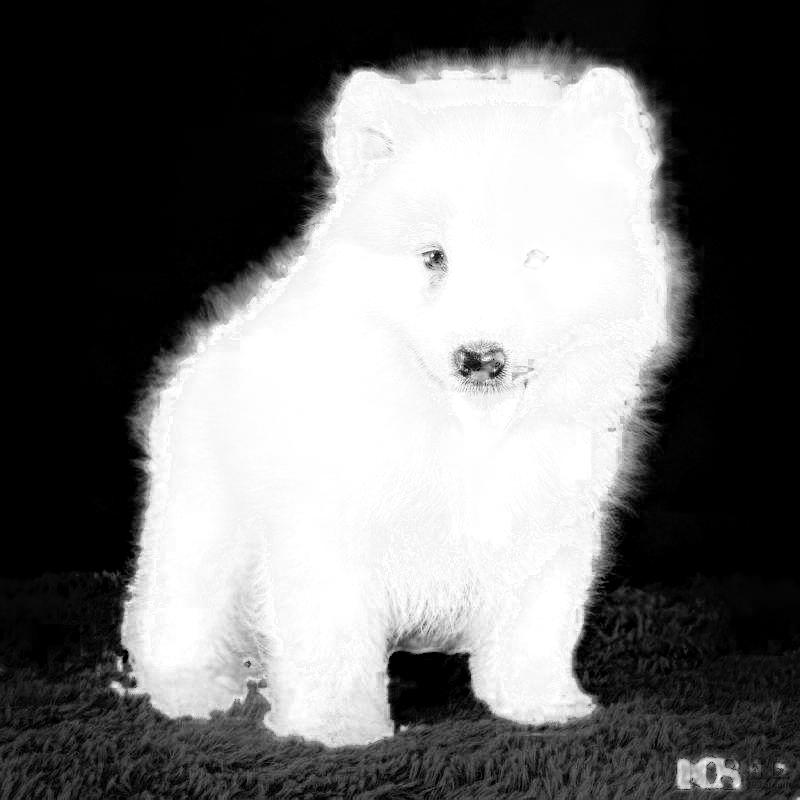} &
			\includegraphics[scale=0.076]{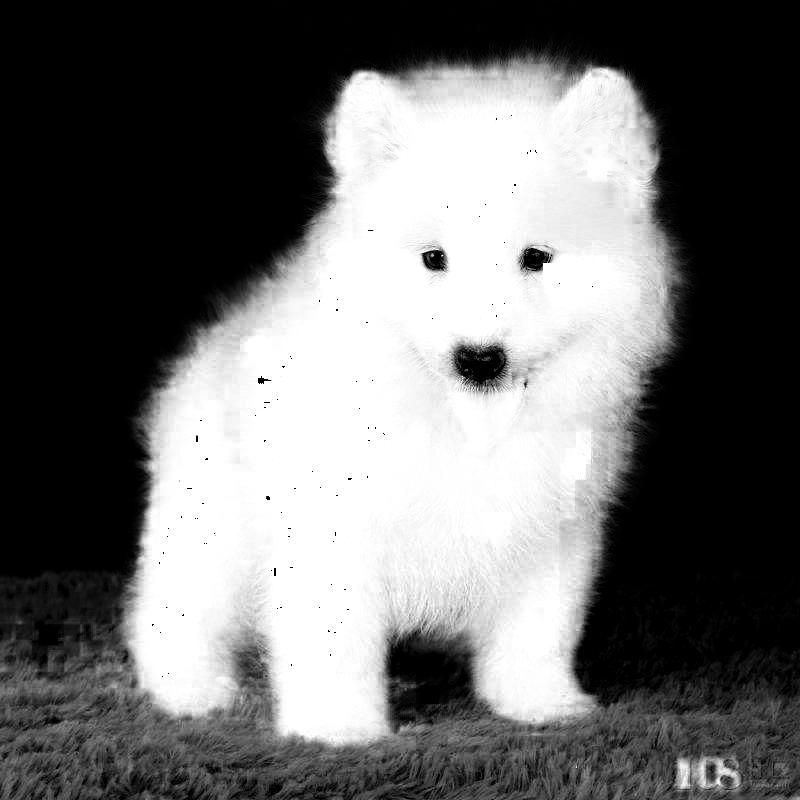} \\
			
			Late Fusion~\cite{Zhang_2019_CVPR} & AM~\cite{NIPS2018_7710} & AM+Closed~\cite{Levin2007A} & AM+KNN~\cite{Chen2013KNN} & AM+DCNN~\cite{Cho2016Natural} & AM+IFM~\cite{Aksoy2017Designing} \\
			
			\includegraphics[scale=0.244]{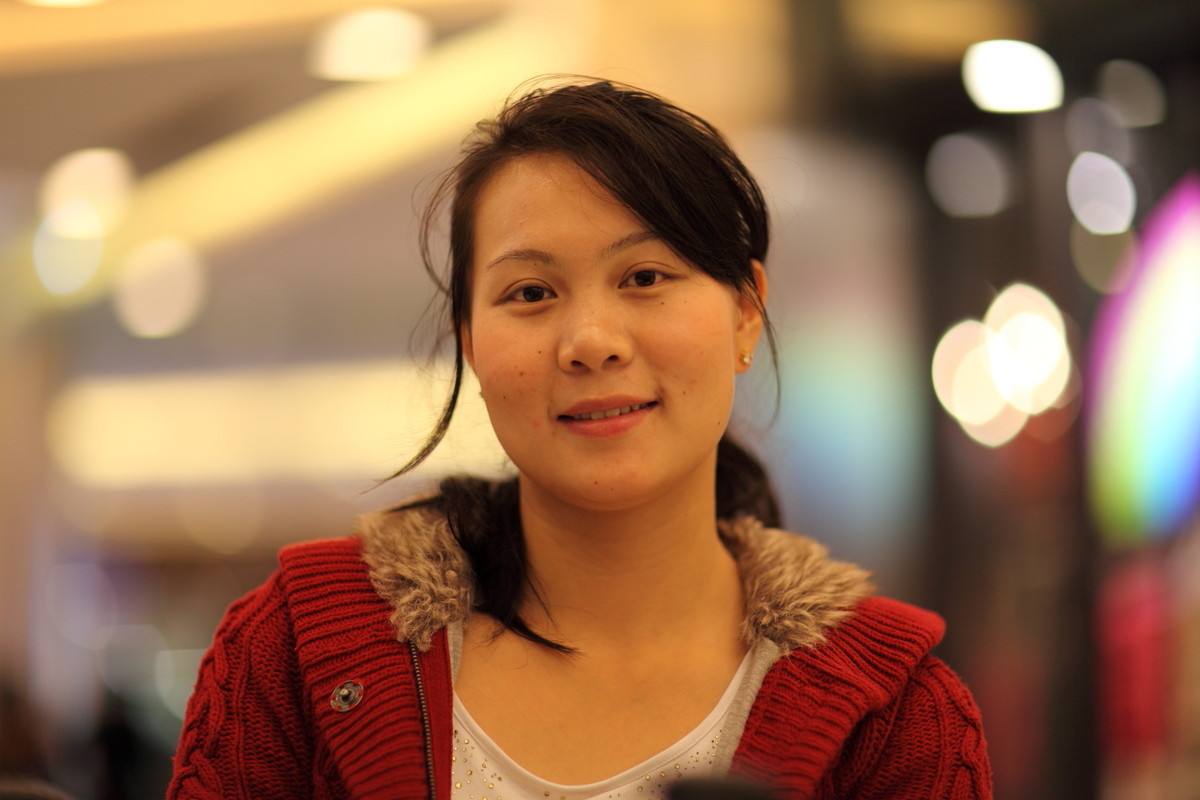} &
			\includegraphics[scale=0.0665]{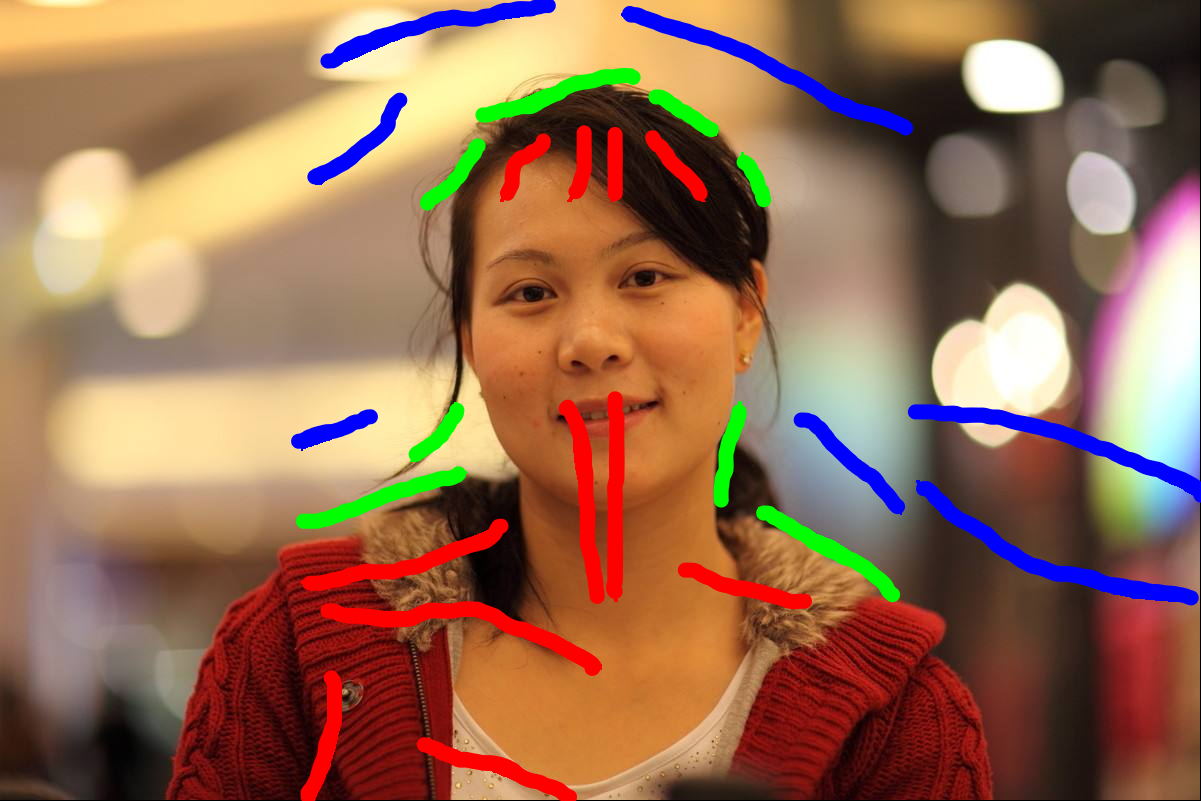} &
			\includegraphics[scale=0.0501]{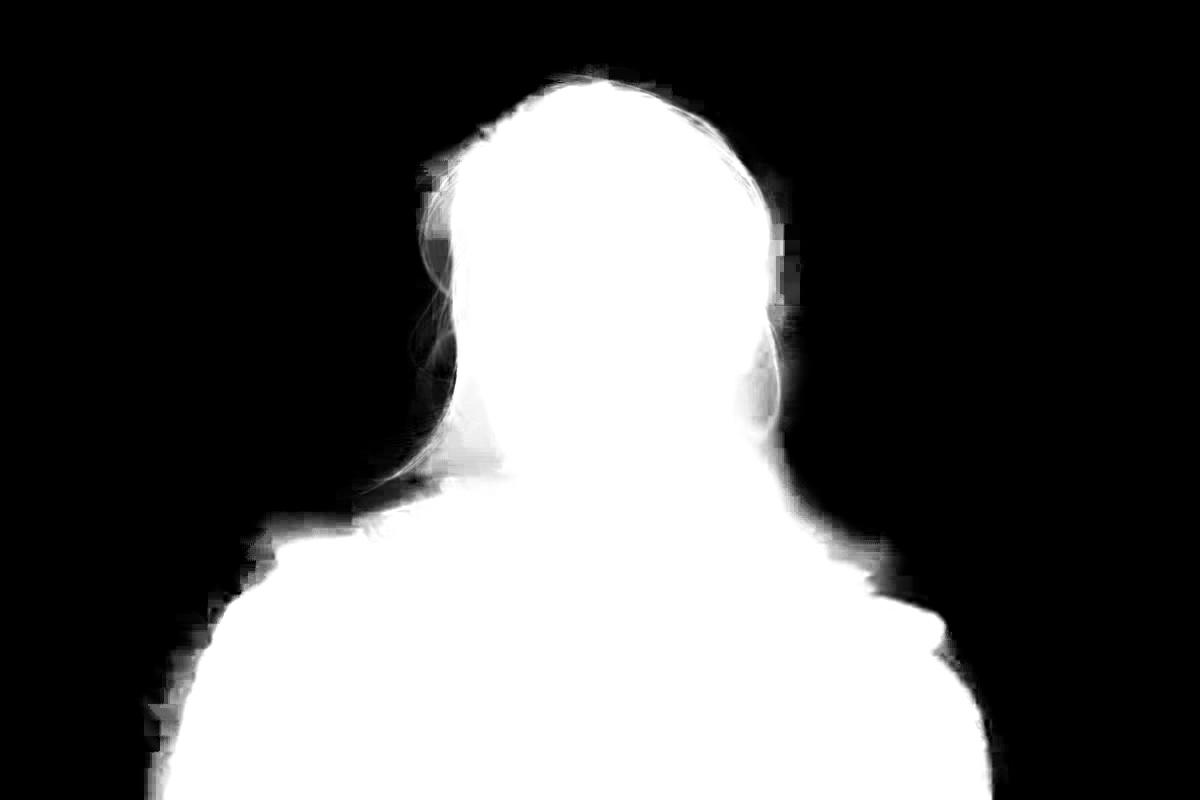} &
			\includegraphics[scale=0.0501]{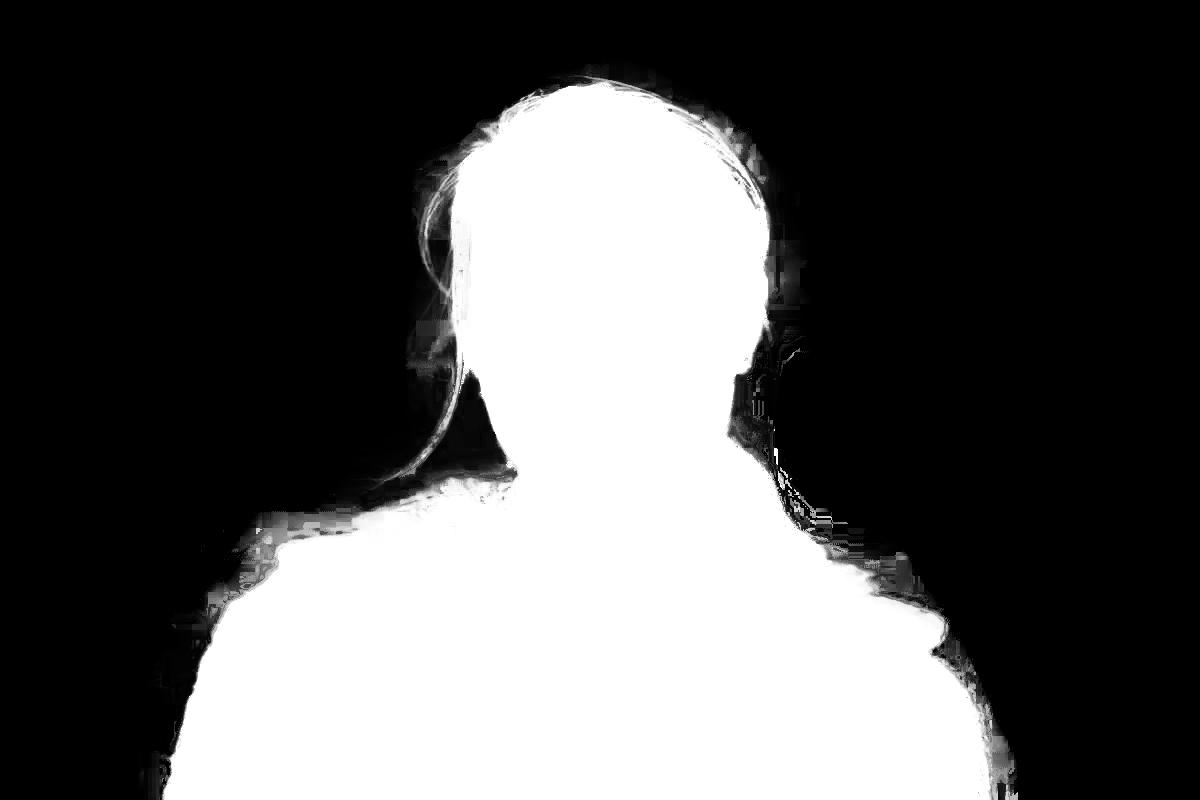} &
			\includegraphics[scale=0.0501]{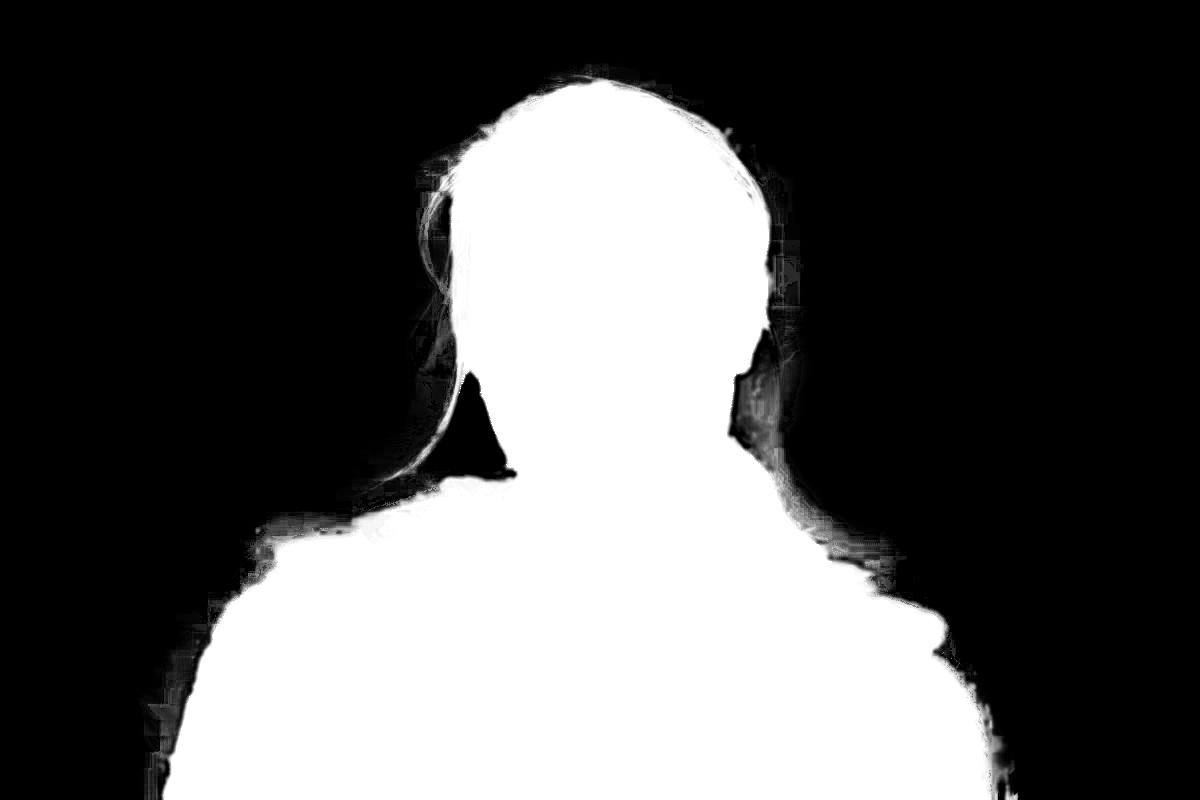} &
			\includegraphics[scale=0.0668]{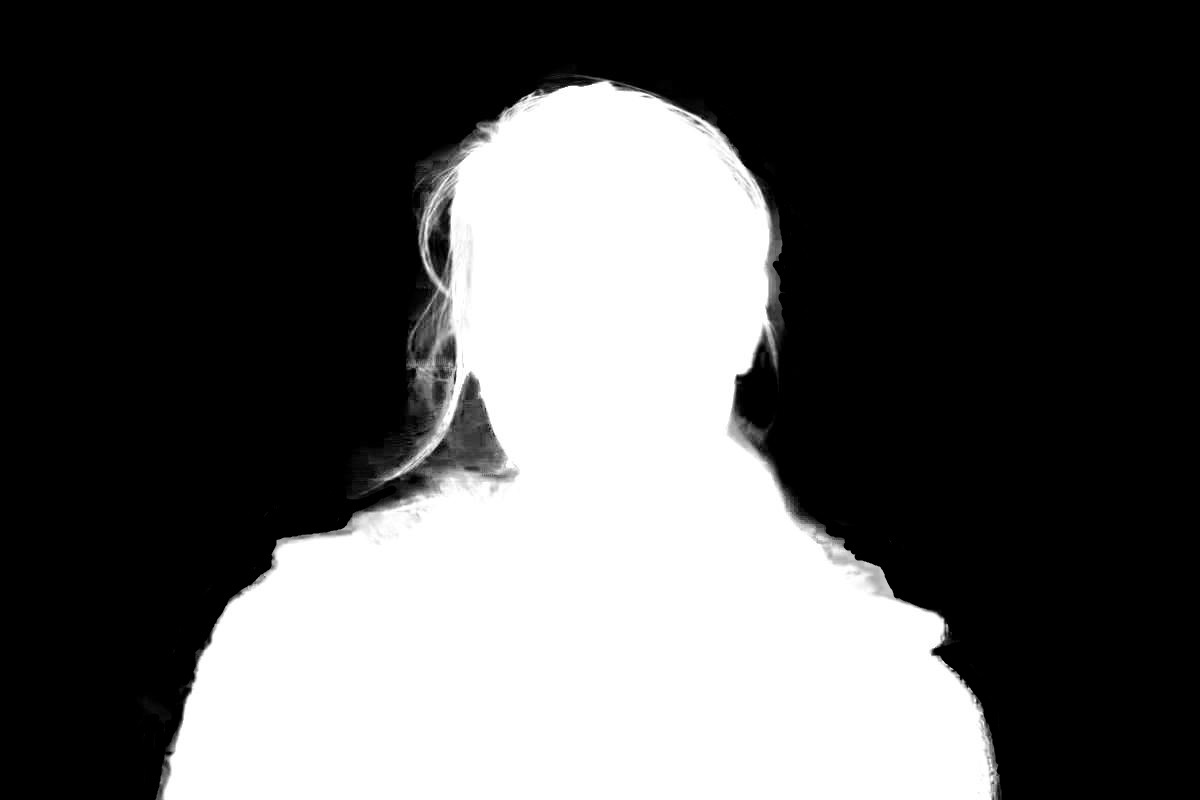} \\
			
			Input Image & Smart Scribbles & SS+Closed~\cite{Levin2007A} & SS+KNN~\cite{Chen2013KNN} & SS+DCNN~\cite{Cho2016Natural} & SS+IFM~\cite{Aksoy2017Designing} \\
			
			\includegraphics[scale=0.0668]{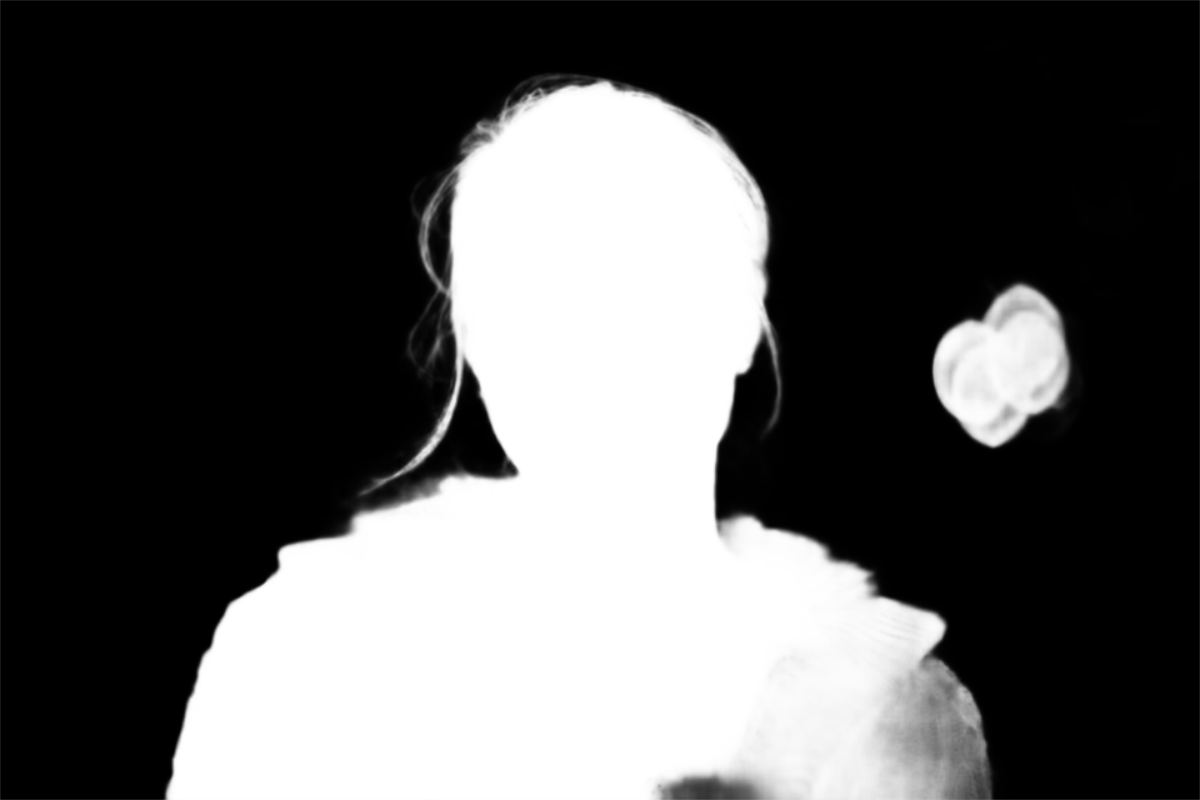} &
			\includegraphics[scale=0.244]{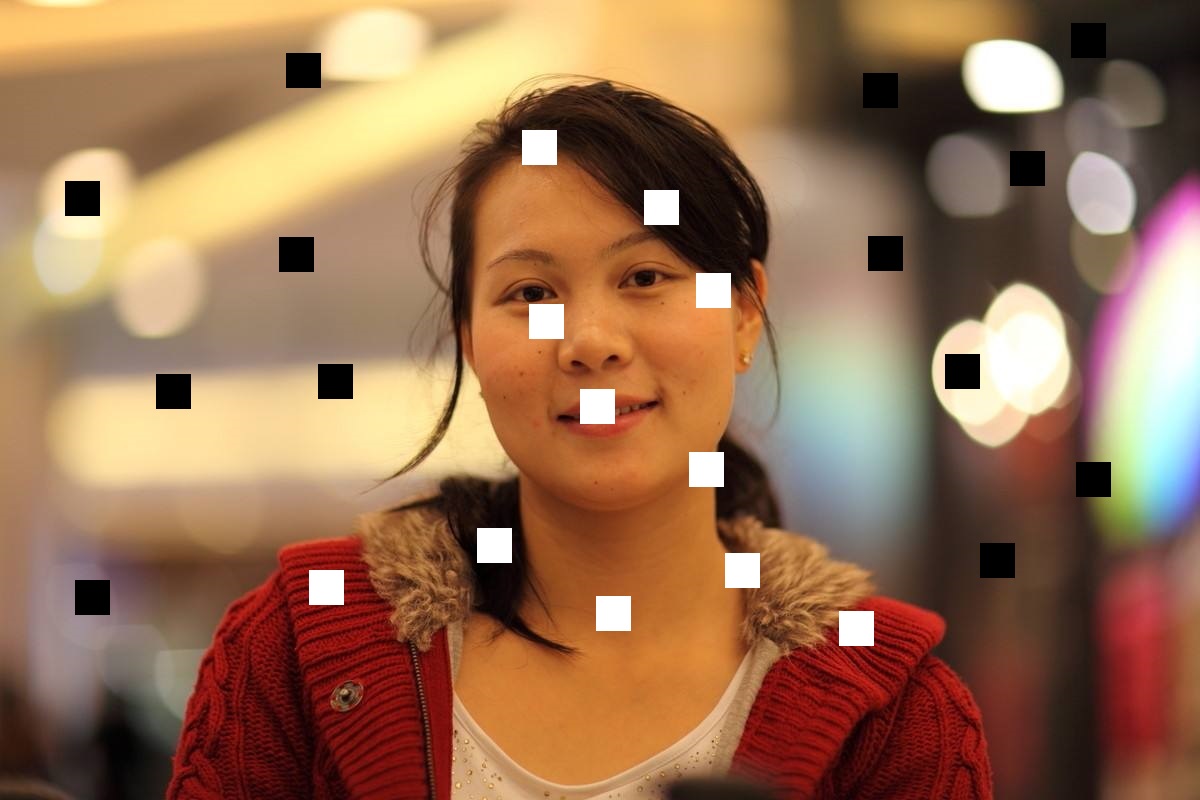} &
			\includegraphics[scale=0.0501]{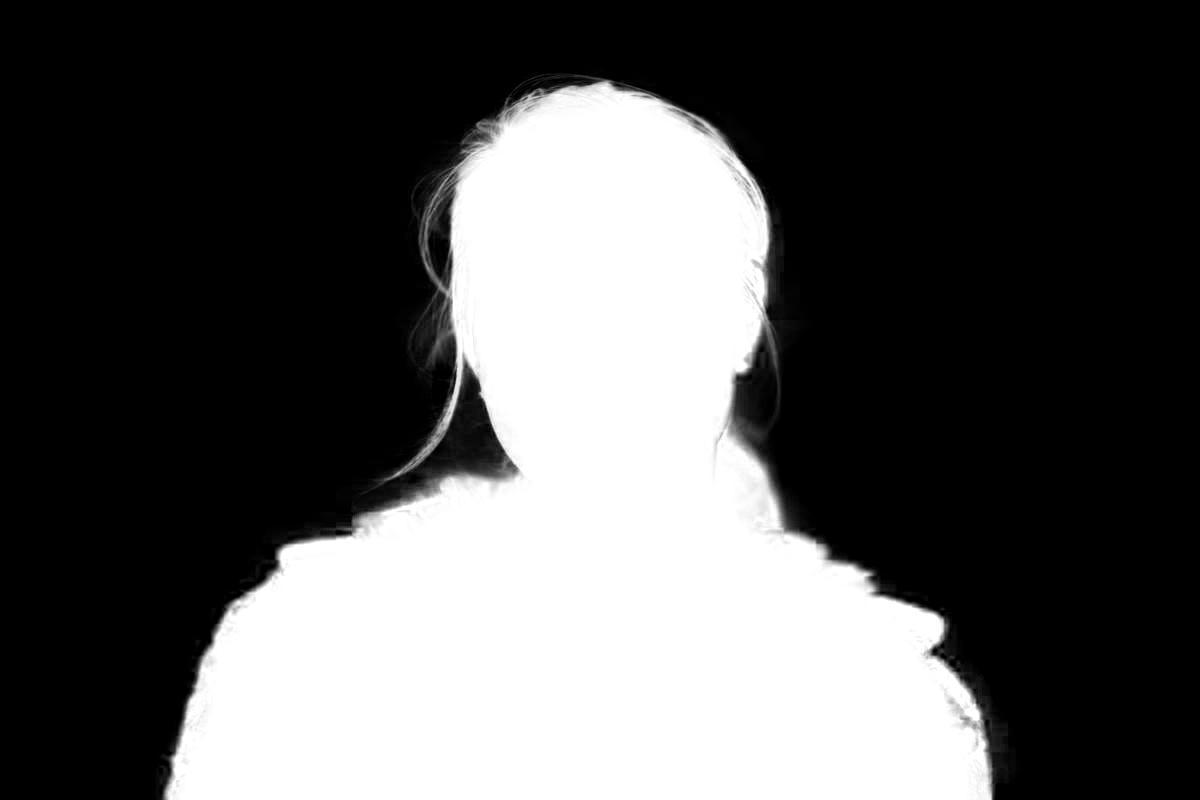} &
			\includegraphics[scale=0.0501]{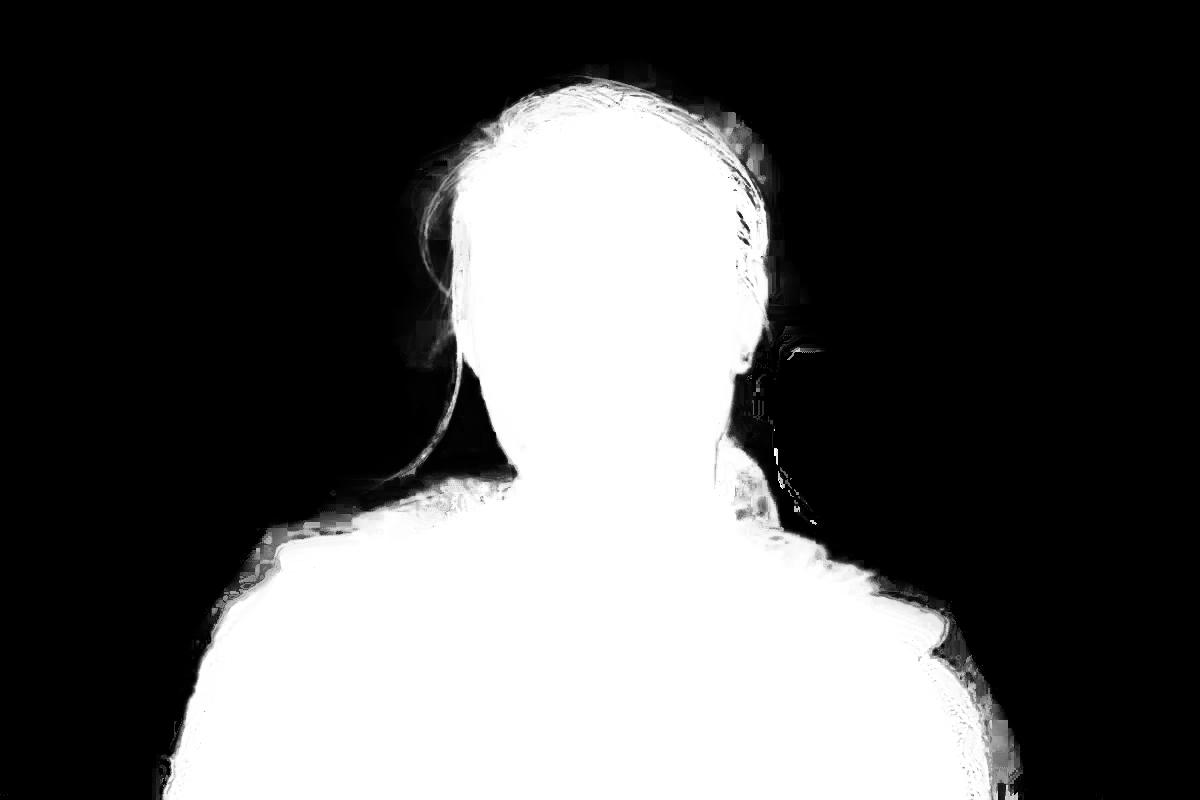} &
			\includegraphics[scale=0.0501]{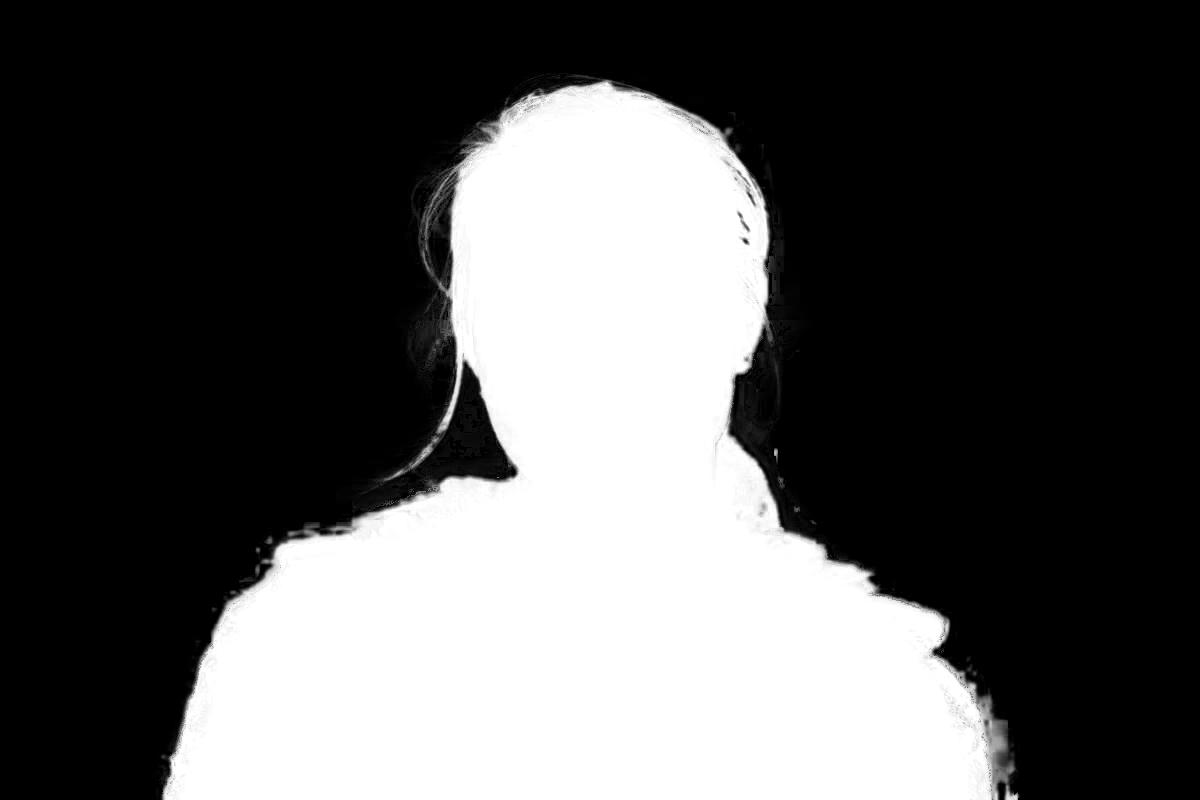} &
			\includegraphics[scale=0.0501]{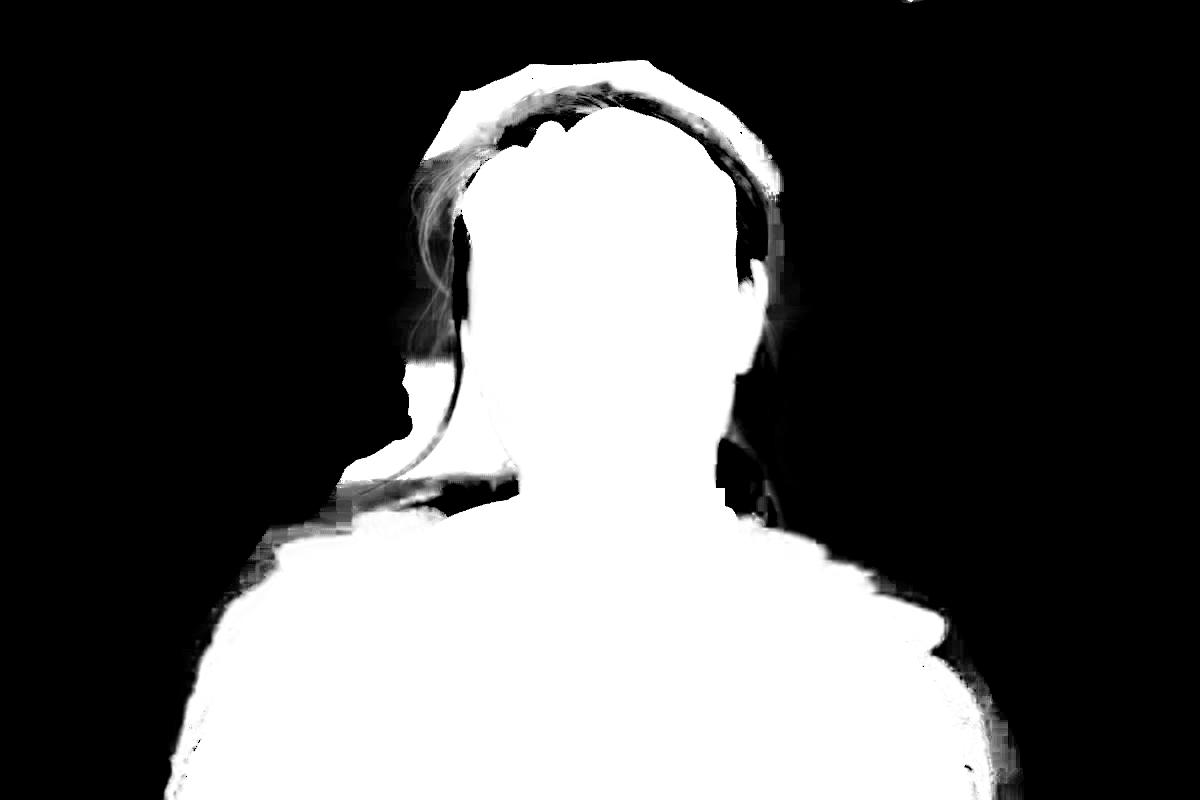} \\
			
			Late Fusion~\cite{Zhang_2019_CVPR} & AM~\cite{NIPS2018_7710} & AM+Closed~\cite{Levin2007A} & AM+KNN~\cite{Chen2013KNN} & AM+DCNN~\cite{Cho2016Natural} & AM+IFM~\cite{Aksoy2017Designing} \\
	\end{tabular}}
	\caption{The visual comparisons with Active Matting~\cite{NIPS2018_7710} and the Late Fusion~\cite{Zhang_2019_CVPR} on the real-world images. }
	\label{fig:visual_nips_2}
\end{figure*}

\subsection{Comparison with Active Matting on the real-world images}
\label{ssec:comWithAM}
Active Matting~\cite{NIPS2018_7710} (AM) can achieve competent alpha mattes with simple user interactions. The combination of CNNs and Reinforcement Learning in AM can effectively spread limited user information to the whole image to produce an accumulated trimap, and the corresponding alpha matte can be generated from an existing embedded matting algorithm. However, like other deep learning based matting networks, AM is also trained with artificial images, which significantly restricts the robustness and generalization of the network. We follow the matting model in AM to generate accumulated trimaps, then produce alpha mattes with existing matting methods. In this experiments, we feed the trimaps from \emph{smart scribbles} into existing matting methods to compare with AM on the real-world images. The visual comparisons are illustrated in Figure~\ref{fig:visual_nips_1} and Figure~\ref{fig:visual_nips_2}. The alpha mattes generated by \emph{smart scribbles} and AM are both from four identical existing matting algorithms: (ClosedForm~\cite{Levin2007A}, KNN~\cite{Chen2013KNN}, DCNN~\cite{Cho2016Natural} and IFM~\cite{Aksoy2017Designing}. The results of AM are coarse in diverse methods, even with more user interactions or relatively simple background. The reason for this is obvious: the overall model is fine tuned on the rendered images. Illumination, blurring, and bokeh in real-world pictures can greatly reduce the network performance obtained from training on artificial images.

Compared to AM, \emph{smart scribbles} can achieve alpha mattes in a more general and valid fashion. The proposed Markov and CNN propagation can spread limited scribbles to the whole image, and all operations are executed in superpixels-level of the input image, which means \emph{smart scribbles} is independent of large-scale dataset. Although the synthetic images supply new possibility for deep learning matting models training, the clear artifacts on them significantly reduce the robustness and versatility of matting network. \emph{Smart scribbles} can automatically adjust and adapt to the specific situation of the input image: 1) our informative region selection can effectively choose some representative regions (the ones with noise, blurring etc.); 2) users can draw few scribbles on suggested regions to separate the foreground, background and unknown; 3) the proposed two-phase propagation can effectually spread category labels to the whole image. Therefore, \emph{smart scribbles} can assign category labels for each superpixel precisely, though the input image is taken in a poor or specific situation. The first row of Figure~\ref{fig:visual_nips_1} demonstrates the robustness of our method, the bokeh and blurring in the input image did not degrade the performance of \emph{smart scribbles}. The hair of the woman is clearly visible and the complicated background is perfectly eliminated. 

\subsection{Scribbles Evaluations}
\label{ssec:comWithScribbles}
\begin{figure*}[t]
	\centering
	\includegraphics[scale=.468]{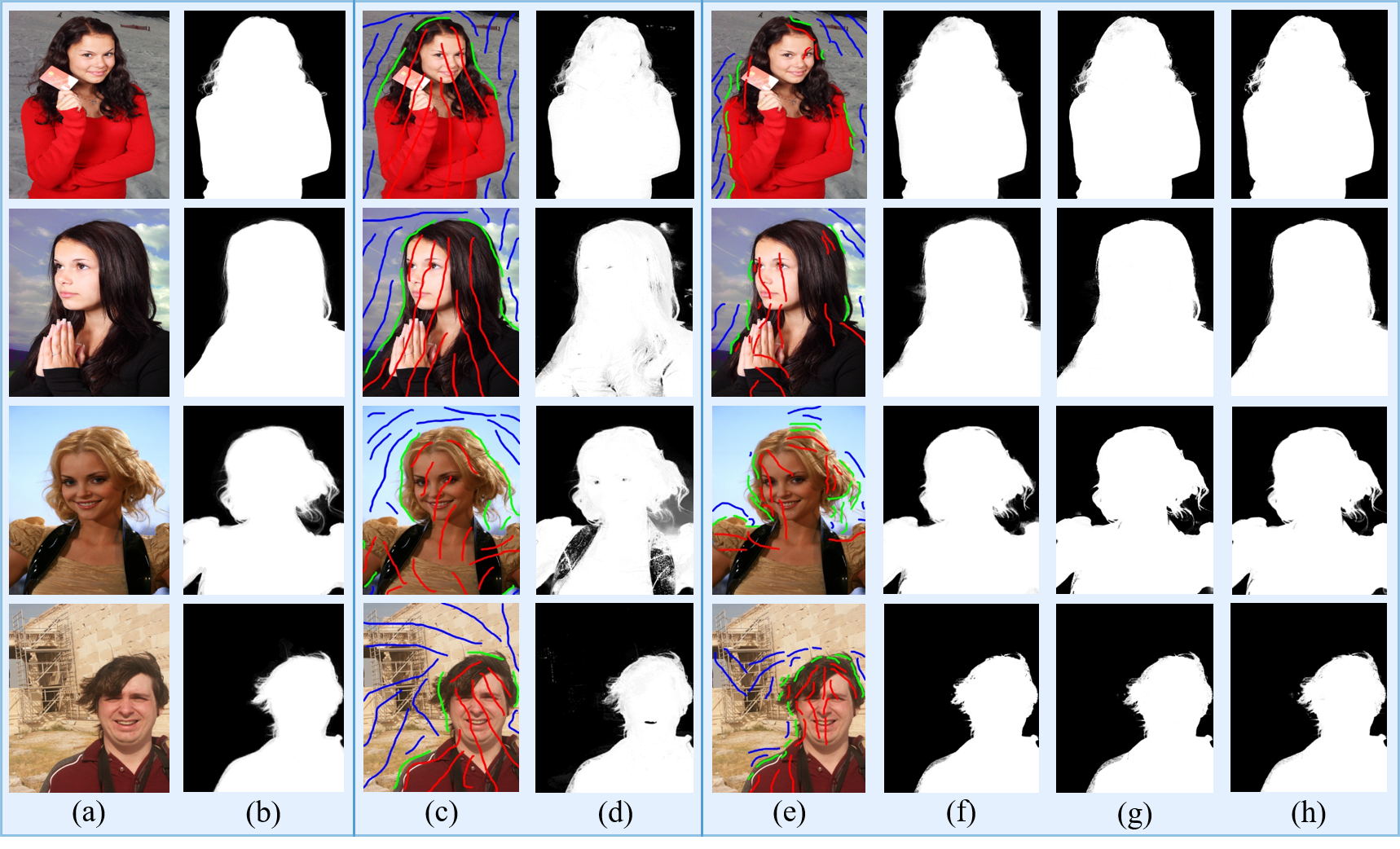}
	\caption{The visual comparisons with traditional scribbles. (a) The input images from the portrait testing dataset or the DIM~\cite{Xu2017Deep} dataset. (b) Ground truths. (c) Traditional scribbles. (d) Traditional scribbles + DCNN~\cite{Cho2016Natural}. (e) \emph{Smart scribbles}. (f) \emph{Smart scribbles} + KNN~\cite{Chen2013KNN}. (g) \emph{Smart scribbles} + DCNN~\cite{Cho2016Natural}. (h) \emph{Smart scribbles} + IFM~\cite{Aksoy2017Designing}. The alpha mattes produced by \emph{smart scribbles} have more complete outlines and abundant texture details.}
	\label{fig:comToscribbles}
\end{figure*}

Traditional scribbles usually require professional knowledge to draw essential labels. The common users may get poor results even if they draw many scribbles on the whole image, because some prior conditions of the matting algorithms are left out of consideration. Compared with traditional scribbles, \emph{smart scribbles} can generate better mattes with less scribbles. To demonstrate the superiority of \emph{smart scribbles}, we conduct this experiment on the matting benchmark~\cite{Rhemann2009A}, the portraits testing dataset~\cite{Shen2016Deep} with 300 images and the DIM dataset. Here we divide 20 inexperienced participators into two groups on average. One group draws traditional scribbles, and the other performs \emph{smart scribbles}. Both groups are only told the basic knowledge of separating foreground, background and unknown areas.  The alpha mattes are produced with diverse matting algorithms (ClosedForm~\cite{Levin2007A}, SharedMatting~\cite{Gastal2010Shared}, KNN~\cite{Chen2013KNN}, DCNN~\cite{Cho2016Natural} and IFM~\cite{Aksoy2017Designing}. The results are shown in Table~\ref{tab:scribbles comparison} with different datasets. The evaluation metric is RMSEs compared with ground truths and \emph{smart scribbles} achieves more than $40\%$ improvement over conventional scribbles methods. \emph{Smart scribbles} can obtain better alpha mattes with different matting algorithms. We attempt to generate alpha mattes with state-of-the-art matting algorithms (IFM~\cite{Aksoy2017Designing}, but the traditional scribbles achieve
\begin{table}[t]
	\caption{The RMSE comparisons with traditional scribbles on the portrait dataset~\cite{Cho2016Natural} and deep image matting (DIM) dataset~\cite{Xu2017Deep}.}
	\label{tab:scribbles comparison}
	\footnotesize
	\centering
	\setlength{\tabcolsep}{2.3mm}{
		\begin{tabular}{l|cc}
			\hline
			Methods \& datasets & Smart scribbles & Traditional scribbles \\
			\hline
			ClosedForm~\cite{Levin2007A}+Portraits & 0.1024 & 0.1806 \\
			SharedMatting~\cite{Gastal2010Shared}+Portraits & 0.0998 & 0.1778 \\
			KNN~\cite{Chen2013KNN}+Portraits & \textbf{0.0726} & \textbf{0.1599} \\
			DCNN~\cite{Cho2016Natural}+Portraits & 0.0870 & 0.1626 \\
			IFM~\cite{Aksoy2017Designing}+Portraits & 0.0733 & 0.5197 \\
			\hline
			ClosedForm~\cite{Levin2007A}+DIM & 0.1104 & 0.1616 \\
			KNN~\cite{Chen2013KNN}+DIM & 0.0859 & \textbf{0.1355} \\
			DCNN~\cite{Cho2016Natural}+DIM & 0.0974 & 0.1390 \\
			IFM~\cite{Aksoy2017Designing}+DIM & \textbf{0.0856} & 0.4257 \\
			\hline
	\end{tabular}}
\end{table}
poor results. The qualitative results are shown in Figure~\ref{fig:comToscribbles} and the preponderance of \emph{smart scribbles} is more obvious on visual outcomes. Compared to traditional scribbles, the alpha mattes produced by \emph{smart scribbles} have no extra background, and the boundary between foreground and background is clearly demarcated. Besides, some furry details can be detected more clearly by \emph{smart scribbles}.

\begin{figure}[t]
	\centering
	\includegraphics[scale=.35]{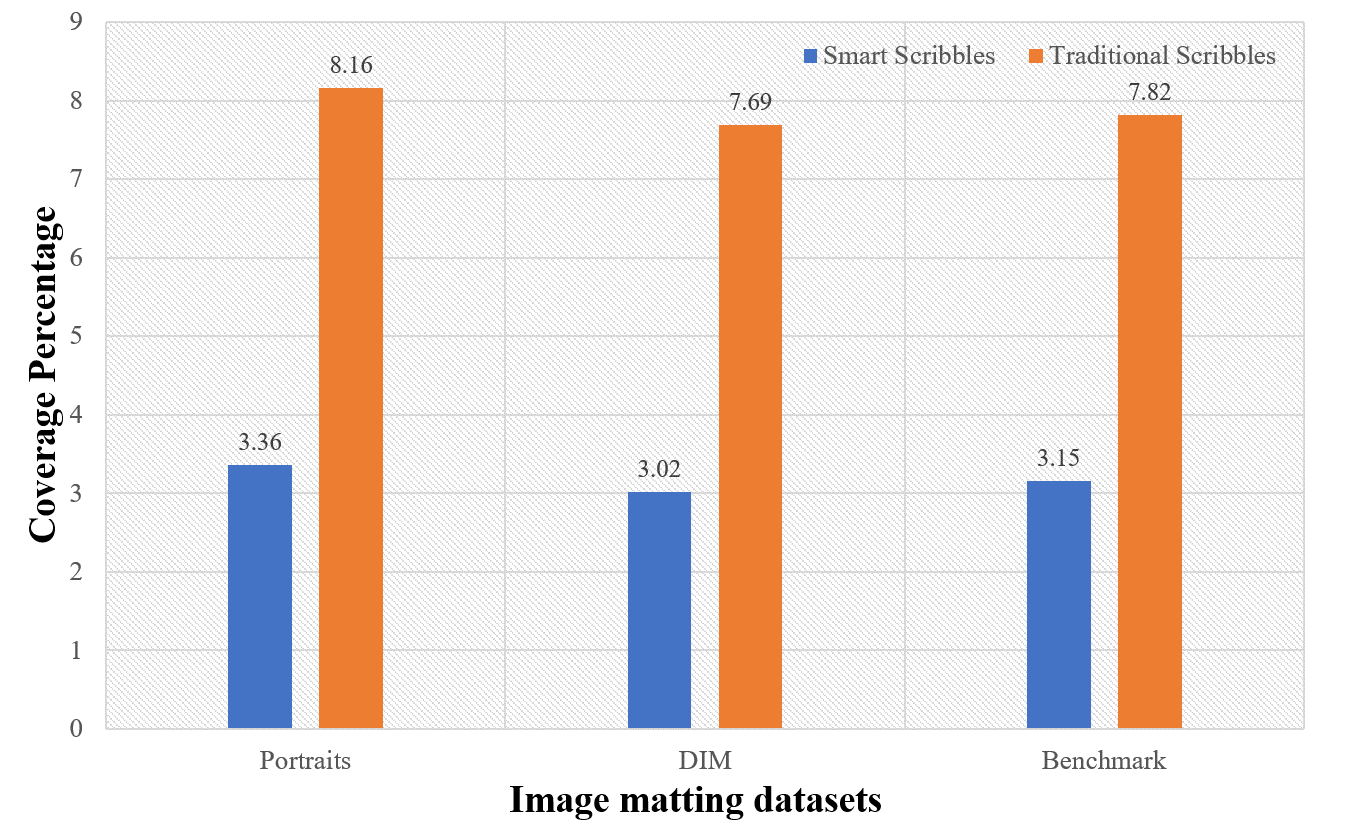}
	\caption{The summarized coverage percentage on diverse datasets, which is the percentage of the input image covered by scribbles. The smaller percentage of \emph{smart scribbles} indicates that it can produce better alpha mattes with less user interactions, as compared with traditional scribbles. }
	\label{fig:coverage}
\end{figure}

Figure~\ref{fig:coverage} reflects the quantitative comparison of additional user inputs in this experiment. The entry 'coverage percentage' indicates the percentage of the images covered by scribbles, which suggests how much labels information is provided through the user interactions. \emph{Smart scribbles} has lower 'coverage percentage', approximately $60\%$ reduction in the number of user scribbles on diverse datasets, indicating less user inputs getting above results. With informative regions selection, we can assure scribbles are drawn on crucial image areas. Limited user information can effectively spread to the whole image spatially and globally via Markov and CNN propagation. The above two points ensure that \emph{smart} scribbles can produce refined mattes with minor scribbles.

\subsection{Comparison with Trimaps}
\label{ssec:comWithGT}

\begin{table}[ht]
	\caption{The comparisons with artificial trimaps on the matting benchmark~\cite{Rhemann2009A}. The last row reflects the running time and the others are RMSEs. Full scratched means handcrafted trimaps, and grabcut trimaps are produced via executing Grabcut~\cite{Rother2004} iteratively. \emph{Smart scribbles} can approximate their poorer results (red marks), while taking much less time.}
	\label{tab:comTotrimap}
	\footnotesize
	\centering
	\setlength{\tabcolsep}{2.0mm}{
		\begin{tabular}{l|ccc}
			\hline
			Methods & Smart scribbles & Full scratched & Grabcut trimaps \\
			\hline
			Learning~\cite{Zheng2009Learning} & 0.0938 & None & \textcolor[rgb]{1.00,0.00,0.00}{0.0732} \\
			ClosedForm~\cite{Levin2007A} & 0.0915 & \textcolor[rgb]{1.00,0.00,0.00}{0.0700} & 0.0731 \\
			KNN~\cite{Chen2013KNN} & \textcolor[rgb]{1.00,0.00,0.00}{0.0751} & 0.0456 & 0.0554 \\
			DCNN~\cite{Cho2016Natural} & 0.0848 & 0.0528 & 0.0647 \\
			IFM~\cite{Aksoy2017Designing} & 0.0791 & 0.0631 & 0.0600 \\
			ThreeLayers~\cite{Li2017Three} & 0.0862 & None & 0.0638 \\
			\hline
			Average time & 41s & 8min & 198s \\
			\hline
	\end{tabular}}
\end{table}

\begin{figure*}[t]
	\centering
	\includegraphics[scale=.32]{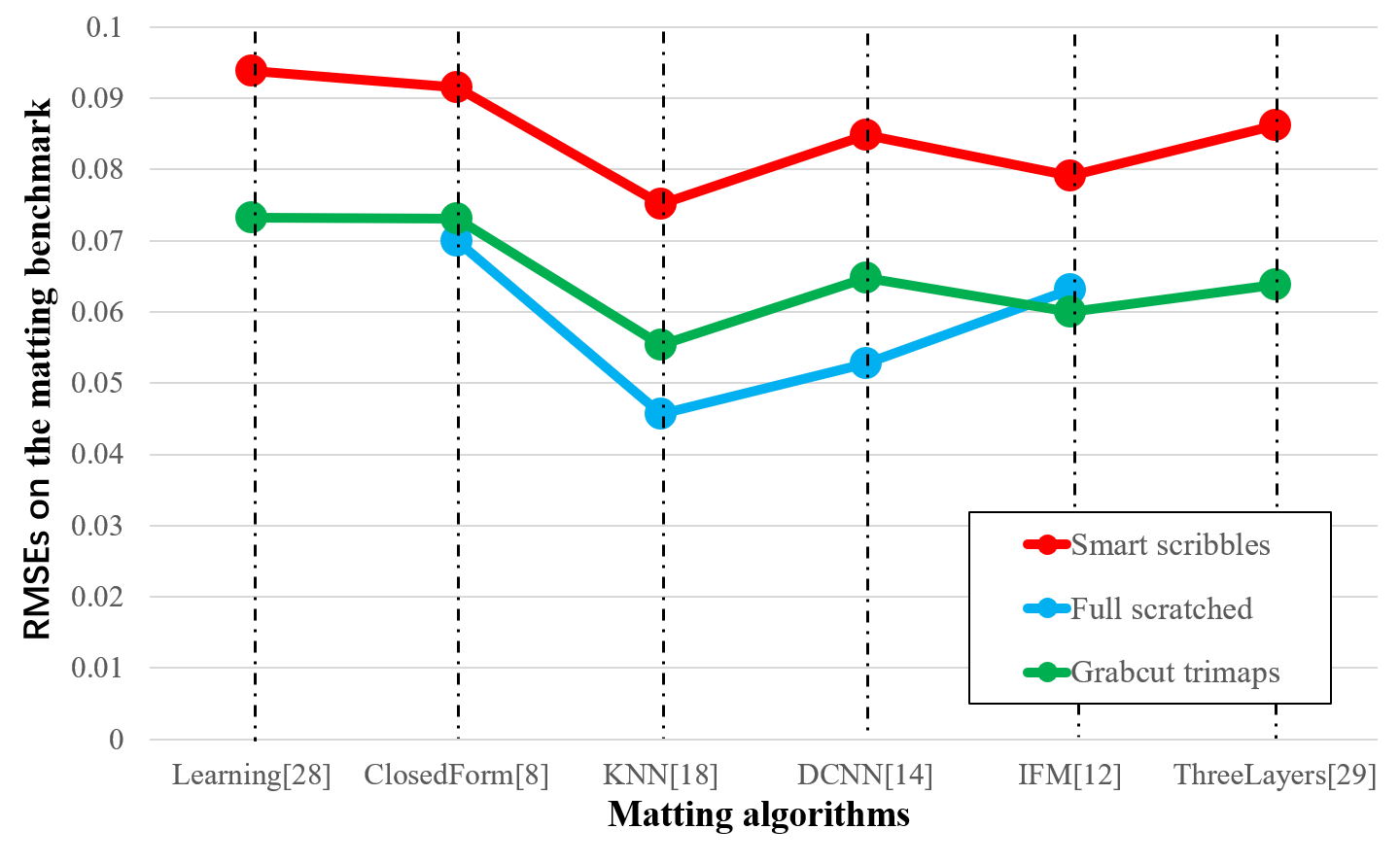}
	\caption{The histogram comparisons with artificial trimaps on the matting benchmark~\cite{Rhemann2009A}. Obviously, \emph{smart scribbles} can get close to their poorer results. Full scratched is unavailable to Learning~\cite{Zheng2009Learning} and ThreeLayers~\cite{Li2017Three} matting due to the presence of unspecified pixels. }
	\label{fig:artificial_trimaps}
\end{figure*}

\begin{figure*}[htp]
	\centering
	\setlength{\tabcolsep}{1pt}\small{
		\begin{tabular}{cccc}
			\includegraphics[scale=0.0961]{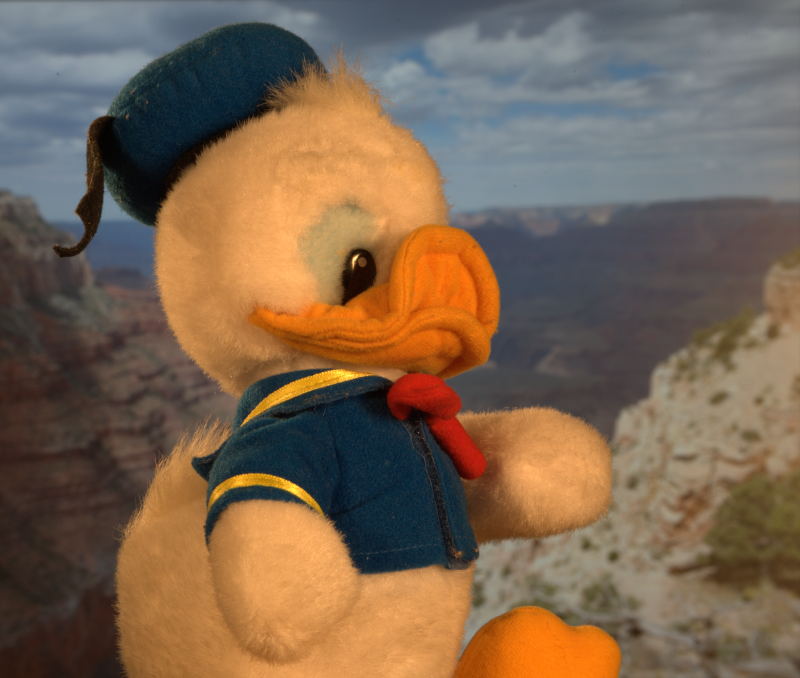} &
			\includegraphics[scale=0.1278]{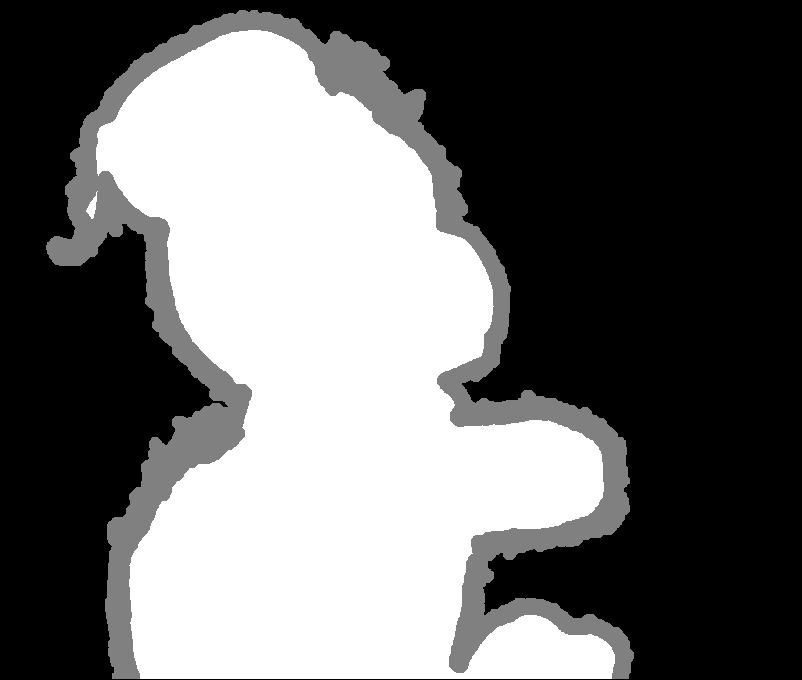} &
			\includegraphics[scale=0.1203]{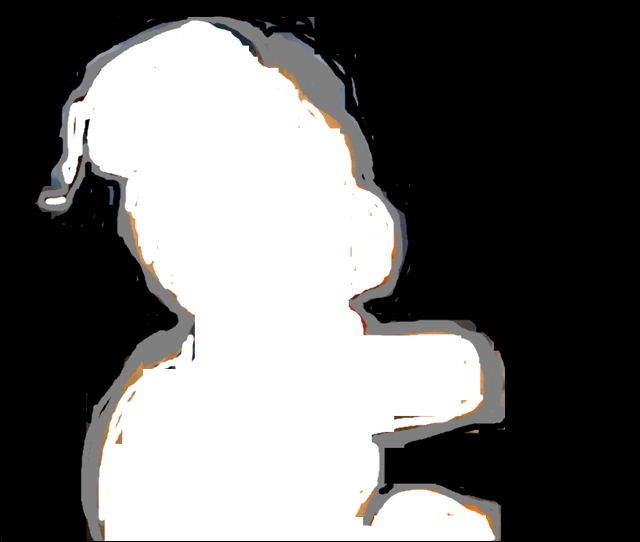} &
			\includegraphics[scale=0.1278]{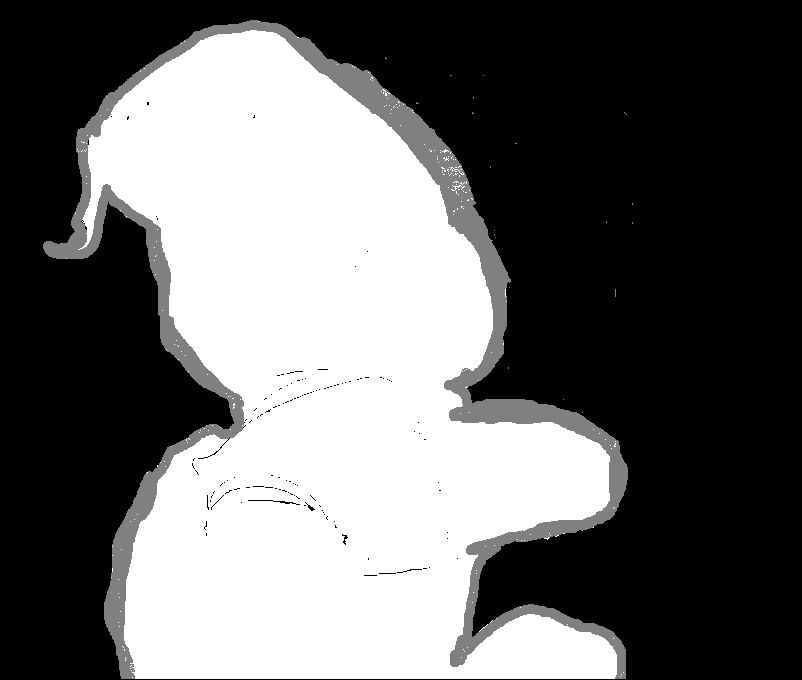} \\
			
			Input Image & Trimap GT & Full Scratched & Grabcut Trimap \\
	\end{tabular}}
	\caption{Diverse trimaps in our experiments. Full scratched and grabcut trimaps are both hand-crafted pixel-level trimaps. }
	\label{fig:diverse_trimaps}
\end{figure*}

Both trimaps and scribbles are classical assistant matte inputs, and scribbles can also be regarded as sparse trimaps. To demonstrate the generality and availability of \emph{smart scribbles}, here we show the comparisons with different forms of trimaps. Trimaps ground truths are accurate pixel-wise annotations and difficult to implement (generally generated by dilation or erosion from alpha mattes, impractical for user interactions based on RGB images), hence we utilize full scratched and Grabcut trimaps instead. Full scratched means drawing scribbles on the whole images according to the trimaps ground truths. Grabcut trimaps are achieved by executing Grabcut~\cite{Rother2004} iteratively. Specifically, users are first asked to draw scribbles and a bounding box for distinguishing the foreground and background, then we utilize Grabcut~\cite{Rother2004} to generate the corresponding trimaps. Such trimaps are employed to produce alpha mattes, and users continue to draw scribbles on the original images to improve these alpha mattes. Drawing scribbles and mattes generation are executing alternately, terminated when users are satisfied with the final alpha mattes. Both full scratched and Grabcut trimaps are approximately pixel-wise annotations, close to trimaps ground truths (Figure~\ref{fig:diverse_trimaps}). We evaluate our experiment on the matting benchmark~\cite{Rhemann2009A}. The matte RMSEs compared to full scratched and Grabcut trimaps are shown in Figure~\ref{fig:artificial_trimaps} (there is no numerical value on the line chart indicating that the input is not suitable for this method.

As Figure~\ref{fig:artificial_trimaps} shows, the mattes RMSEs of full scratched and Grabcut trimaps are apparently lower than \emph{smart scribbles}, suggesting that abundant user inputs can significantly improve the quality of final mattes. Nevertheless, the matte results generated by \emph{smart scribbles}, using certain algorithms (e.g. KNN~\cite{Chen2013KNN} and IFM~\cite{Aksoy2017Designing}), can get close to some poor results in full scratched or Grabcut trimaps. Compared to full scratched and Grabcut trimaps, \emph{smart scribbles} are more time-saving and user-friendly. We achieve the alpha mattes in Table~\ref{tab:comTotrimap} with average interaction time $41$s, while full scratched and Grabcut trimaps take $8$min and $198$s respectively. Besides, \emph{smart scribbles} only requires several scribbles from users to label different categories, in contrast, more relaxed for novice users.

\subsection{Comparison to Region Selection Baselines}
\label{ssec:baselines}
We have constructed three baselines in our experiments:

\begin{figure*}[t]
	\centering
	\includegraphics[scale=.408]{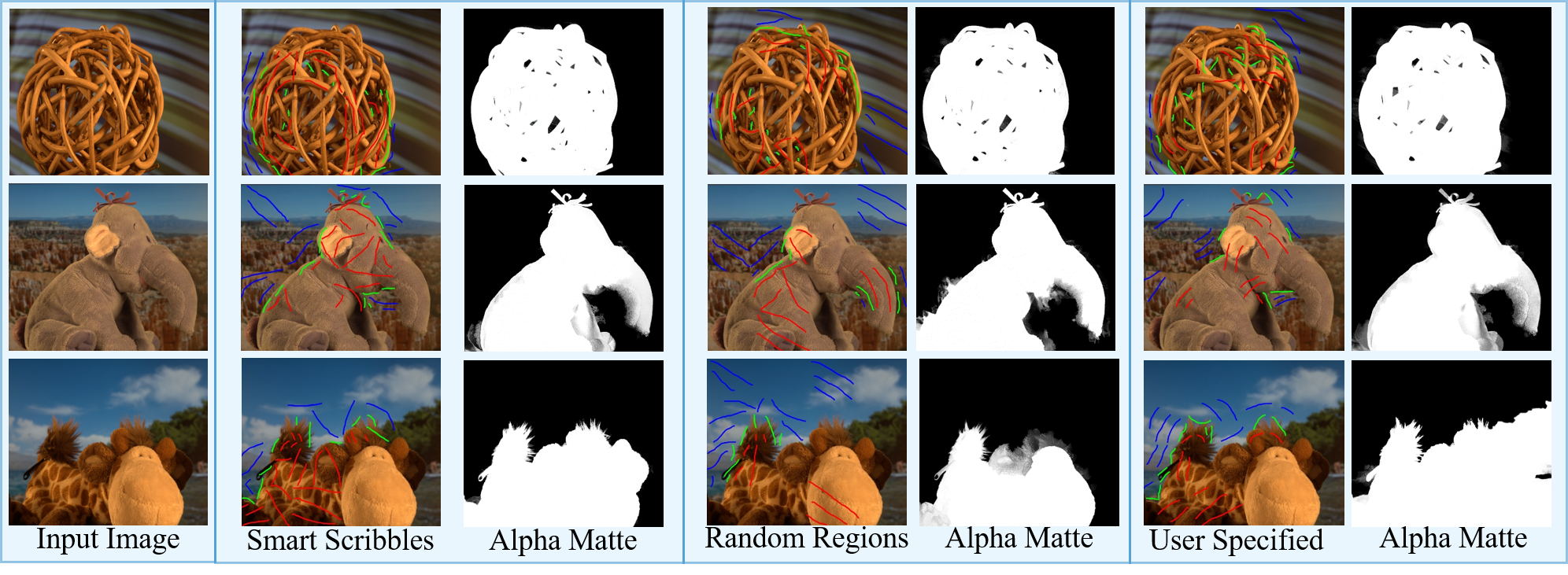}
	\caption{\emph{Smart scribbles} compared with randomly selected regions and user-specified regions. Both of them fail to take full advantage of limited scribbles, causing some details to be lost or even incomplete silhouettes in the final mattes. }
	\label{fig:baseline1}
\end{figure*}
\textbf{B1:How to select informative regions?} We propose \emph{information content} to select \emph{informative regions}, summarizing color, texture, labels information and object boundary. Here we compare the proposed region selection method with the other two. The first randomly select one region from the top 6 of information content per iteration, and the second way is for the users to specify regions based on their visual observation, which is to imitate the professional requirements for users in traditional scribbles. We term these two manners as random and users-specified regions respectively, and replace the \emph{information content} regions selection in our pipeline for comparisons.

Figure~\ref{fig:baseline1} shows the performance of different selection manners. We evaluate Root Mean Square Errors (RMSEs) of the alpha mattes produced by smart scribbles and ground truths on the matting benchmark training dataset, and the quantitative results are shown in Table~\ref{tab:summary_table}. Random regions manner (Figure~\ref{fig:baseline1}) specially fails in the thin structures (the furs of the elephant and monkey). User-specified manner give an inaccurate shape estimation (the last row of Figure~\ref{fig:baseline1}). In contrast, as proposed \emph{smart scribbles} consider the local affinity structure and the correlations across regions synthetically by a novel information content formulation, the global representative regions can be selected successfully. The \emph{smart scribbles} demonstrates enormous superiority in three aspects: the structure details (the third row in Figure~\ref{fig:baseline1}, the hair on the monkey), the shape completeness (the second row in Figure~\ref{fig:baseline1}, the abdomen of the elephant) and the lower RMSE in Table~\ref{tab:summary_table}.
\begin{table}[ht]
	\caption{The quantitative results of baseline1 and two ablation study (RMSEs). The first two rows compare different region selection strategies and the next two lines are propagation ablation results. The last four rows besides \emph{smart scribbles} reveal the significance of different information content entries. }
	\label{tab:summary_table}
	\footnotesize
	\centering
	\setlength{\tabcolsep}{3.2mm}{
		\begin{tabular}{l|ccc}
			\hline
			Methods & ClosedForm & KNN & DCNN \\
			\hline
			Random regions & 0.2010 & 0.1434 & 0.1608 \\
			User-specified regions & 0.1326 & 0.0942 & 0.1020 \\ 
			\hline
			Without Markov propagation & 0.2198 & 0.1717 & 0.1886 \\
			Without CNN propagation & 0.1934 & 0.0963 & 0.1234 \\
			\hline
			Without neighboring similarity & 0.1452 & 0.0946 & 0.1068 \\
			Without inner diversity & 0.1319 & 0.0940 & 0.0969  \\
			Without regions entropy & 0.1624 & 0.1065 & 0.1182 \\
			Without edge score & 0.2447 & 0.1766 & 0.1899 \\
			\hline
			Smart scribbles & \textbf{0.0915} & \textbf{0.0751} & \textbf{0.0848} \\
			\hline
	\end{tabular}}
\end{table}

\begin{figure*}[t]
	\centering
	\includegraphics[scale=.32]{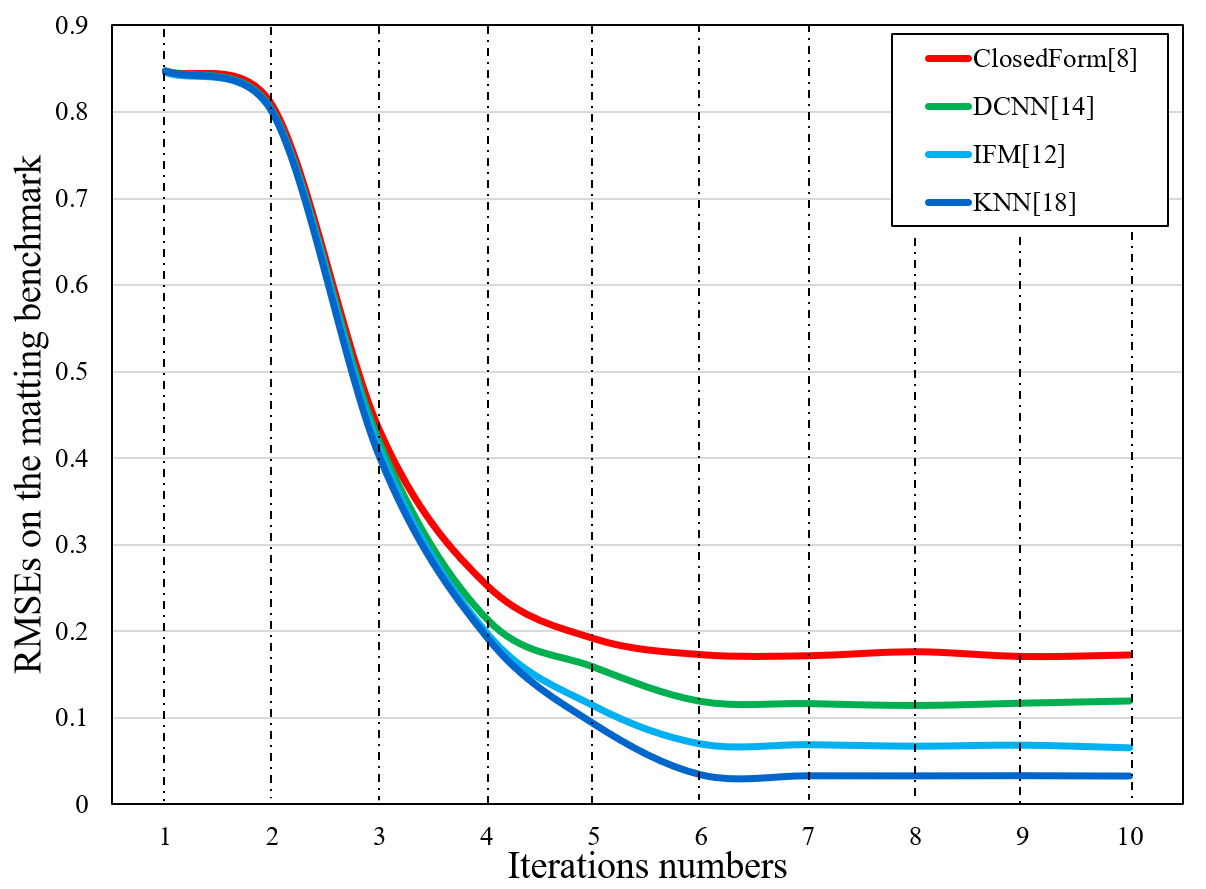}
	\caption{Compare the number of iterations. The RMSEs gradually descend as the number of iterations increases and become stable after $6$ times.}
	\label{fig:iter_num}
\end{figure*}
\textbf{B2:How many regions is essential?} This experiment is to verify the impact of iterations in the phase of Markov propagation. In Markov propagation, we produce a coarse trimap after 6 iterations. A large number of iterations can memorably improve the quality of immediate trimaps and final mattes. Nevertheless, the cumbersomeness of user interaction increases significantly as the number of iterations increases. We conduct our experiments with different iteration numbers to reflect the rationality of setting the number of iterations in \emph{smart scribbles}.

As illustrated in Figure~\ref{fig:iter_num}, here we employ diverse matting algorithms (ClosedForm~\cite{Levin2007A}, KNN~\cite{Chen2013KNN}, DCNN~\cite{Cho2016Natural}, IFM~\cite{Aksoy2017Designing}) to calculate alpha mattes, and the number of iterations is calculated from 1 to 10. In the initial two iterations, the two-phase propagation has insufficient labels information for reference, therefore the results are unsatisfactory and RMSEs are higher. RMSEs decline as the number of iterations increases, and the trend of descending levels off when the iterations times up comes to 6. More iterations can improve mattes slightly, but more users labors are involved, which goes against our intention of reducing user labors. To balance the trade-off between the quality and efficiency, we set the number of iterations as $6$ in practice.

\begin{figure*}[htp]
	\centering
	\setlength{\tabcolsep}{1pt}\small{
		\begin{tabular}{ccc}
			\includegraphics[scale=0.18]{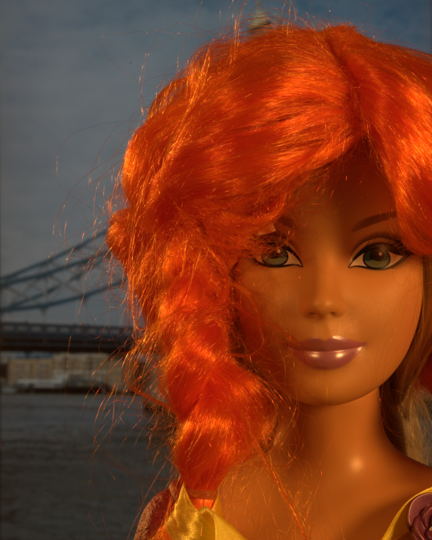} &
			\includegraphics[scale=0.18]{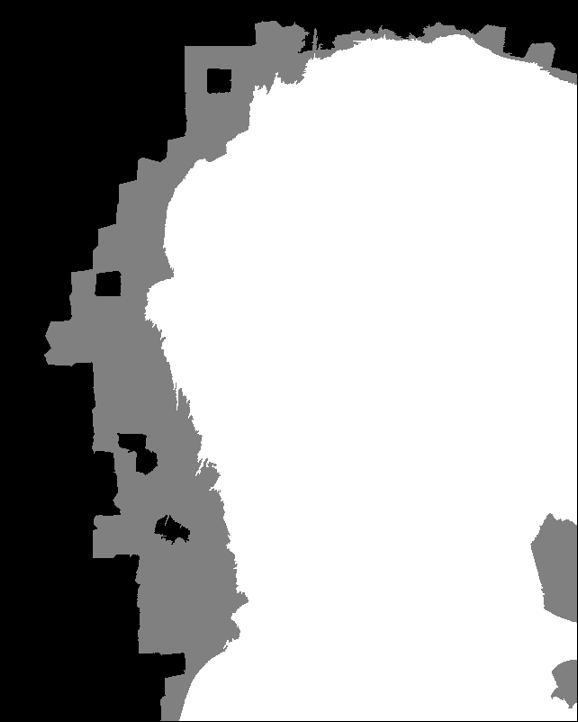} &
			\includegraphics[scale=0.18]{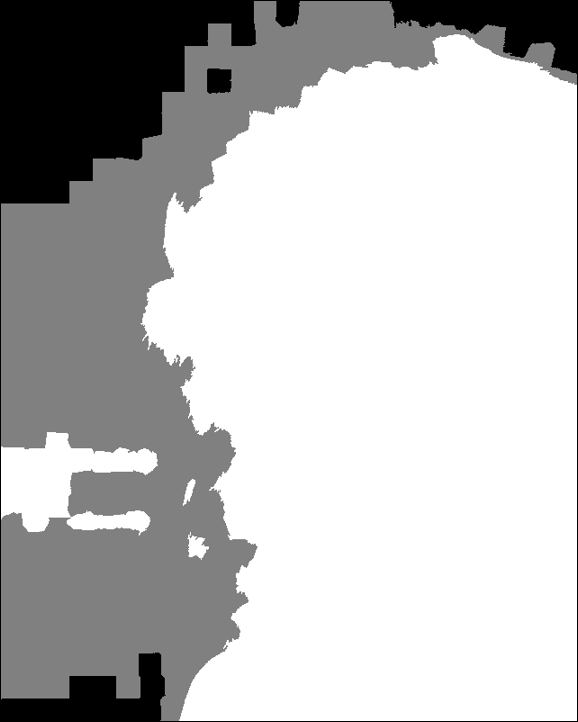} \\
			
			Input Image & With Iteration & Without Iteration \\
	\end{tabular}}
	\caption{Trimap comparison with/without iterative region selection.}
	\label{fig:iterations}
\end{figure*}

\textbf{B3:Why select regions iteratively?} For each region selection, we record users scribbles and perform a Markov propagation to update labels probabilities. For next selection, the entropy in the information content is recalculated according to new labels probabilities, while similarity, diversity and edge score remain. Here, we exclude the entropy from information content and select $N$ regions at a time for users to drawing scribbles. Then Markov propagation is executed once, followed by the CNN propagation. Here we display the trimaps comparisons for more intuitive (Figure~\ref{fig:iterations}). Selecting regions at once weakens the impact of Markov propagation on adjacent affinity, leading to local discontinuity, as shown in the left side of Figure~\ref{fig:iterations}.

\subsection{Ablation Study and Analysis}
\label{ssec:ablation}

\begin{figure*}[htp]
	\setlength{\tabcolsep}{1pt}\small{
		\begin{tabular}{ccccc}
			\includegraphics[scale=0.115]{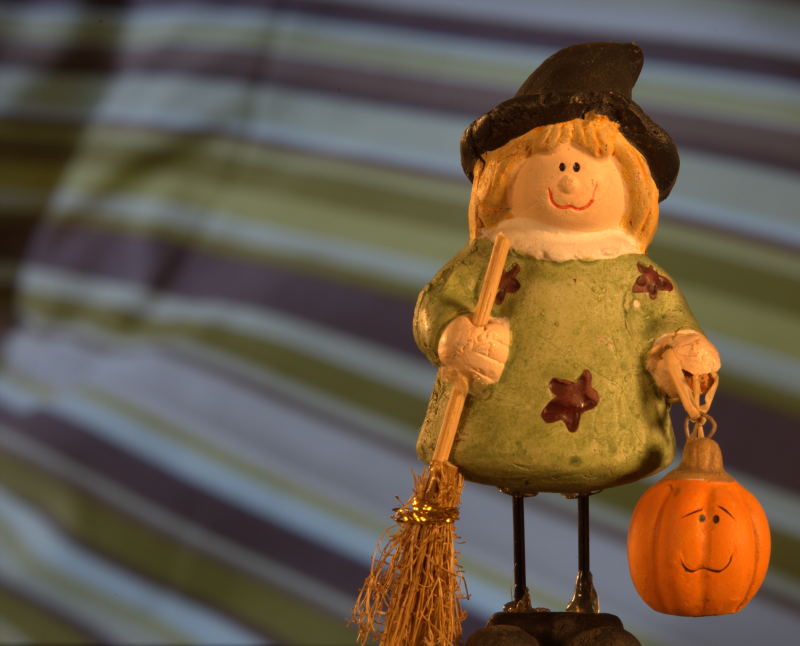} &
			\includegraphics[scale=0.115]{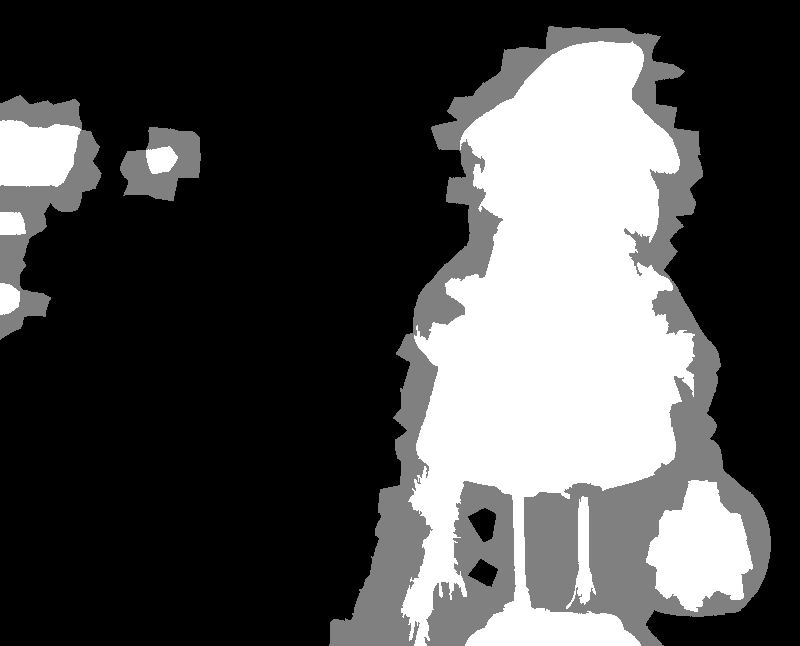} &
			\includegraphics[scale=0.115]{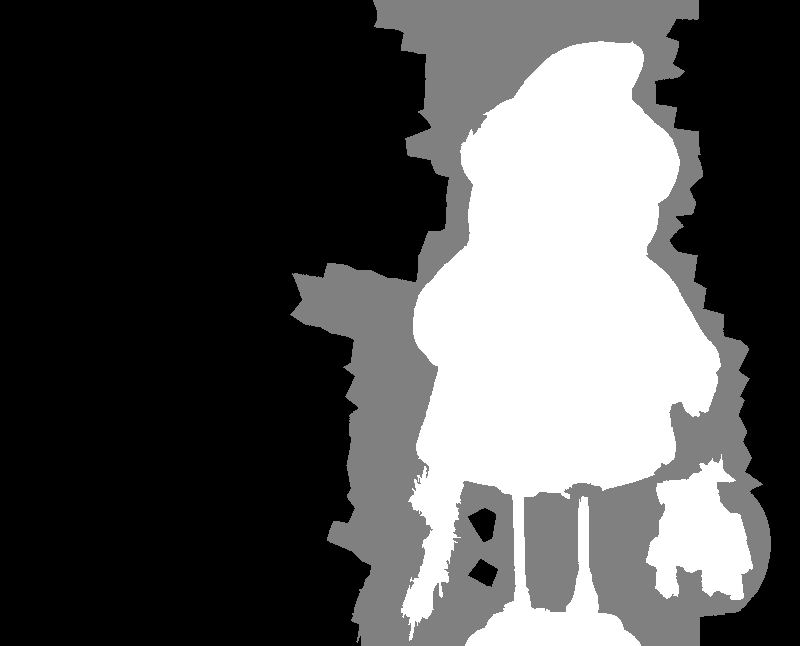} &
			\includegraphics[scale=0.115]{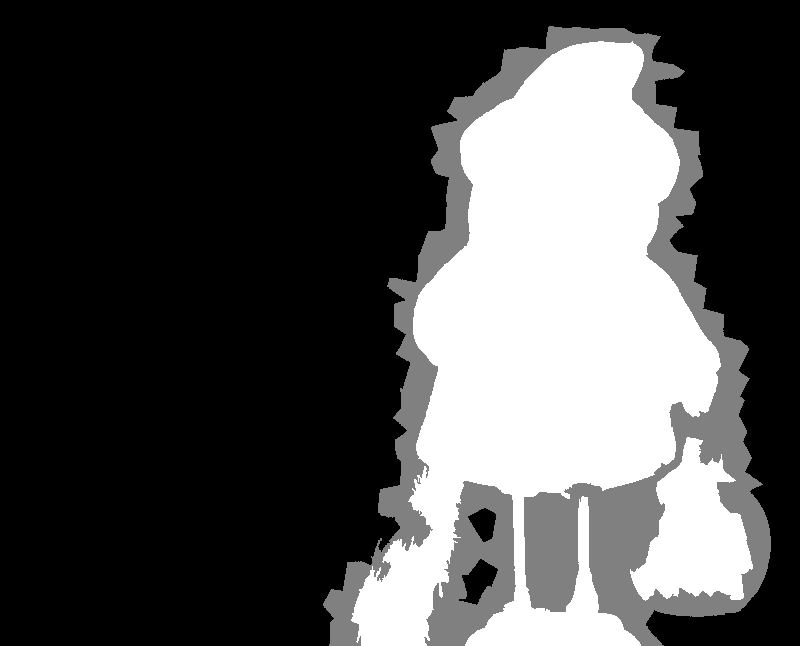} \\
			
			Input Image & Without Markov & Without CNN & Full Framework \\
	\end{tabular}}
	\caption{The qualitative results of the propagation ablation study. The proposed two propagation phases are complementary and the absence of either can possibly lead to prediction errors in the refined trimaps. }
	\label{fig:ablation}
\end{figure*}

\textbf{Propagation ablation.}Here we verify the effectiveness of Markov propagation and CNN propagation, where the two components are evaluated by removing the other one from \emph{smart scribbles}. We conduct ablation experiment on the matting benchmark training images to evaluate three kinds of framework: \emph{smart scribbles}, the framework without Markov propagation and the framework without CNN propagation. Here we adopt ClosedForm~\cite{Levin2007A}, KNN~\cite{Chen2013KNN} and DCNN~\cite{Cho2016Natural} for mattes generation. The quantitative results are summarized in Table~\ref{tab:summary_table} and the visual trimaps are shown in Figure~\ref{fig:ablation}. The framework without Markov propagation and the framework without CNN propagation both produced less competent results, inferior to our full framework. The absence of Markov propagation affects the continuity of spatial label distribution, while the CNN propagation resorts to the high-level features of the superpixels themselves, lacking the constrain of image context. The ablation experiment demonstrates that both propagation are essential in \emph{smart scribbles}.

\textbf{Information content entry ablation.} We summarize $4$ entries for information content formulation according to Equation~(\ref{eq:info}) and all them are essential for informative regions selection. Here we conduct an ablation experiment removing four entries in turn, to illuminate the significance of them. We calculate information content without neighboring similarity, without inner diversity, without entropy, without edge score respectively and produce corresponding mattes with DCNN~\cite{Cho2016Natural}. The quantitative RMSEs results are shown in Table~\ref{tab:summary_table} and the combination of these four terms leads to a minimum RMSE ($0.0848$). The absence of either entry can discount the final mattes: the removal of similarity or diversity lack of consideration for image color and texture; the entropy elimination has no regard for existing labels information; the edge score has the greatest impact due to its accurate identification of object edges.
\begin{figure*}[htp]
	\setlength{\tabcolsep}{1pt}\small{
		\begin{tabular}{ccccc}
			\includegraphics[scale=0.12]{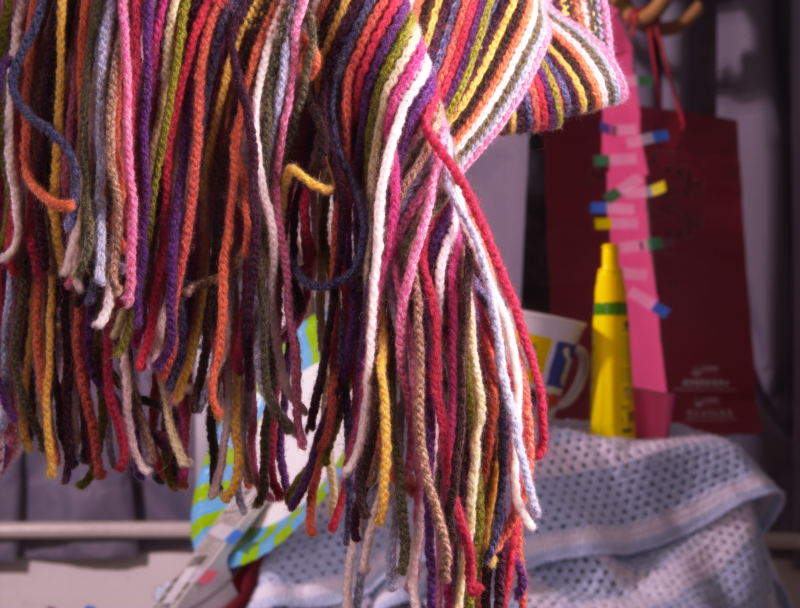} &
			\includegraphics[scale=0.12]{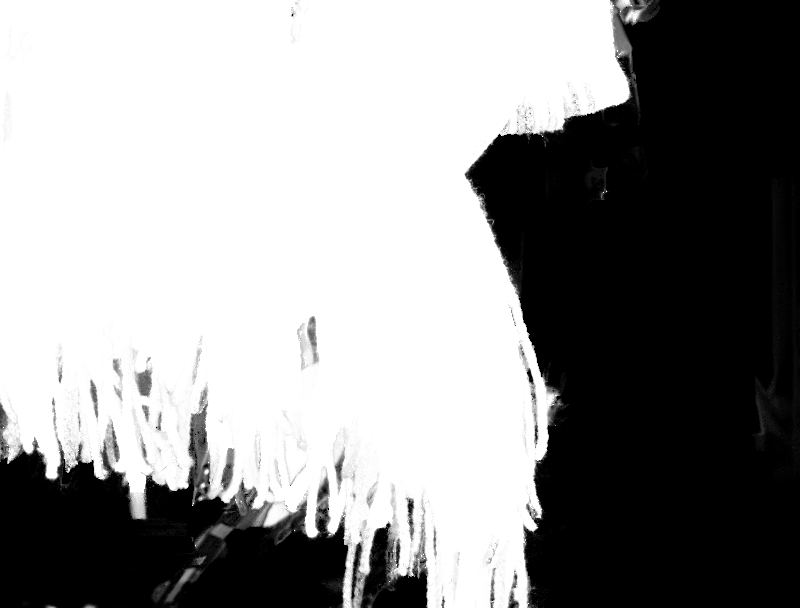} &
			\includegraphics[scale=0.12]{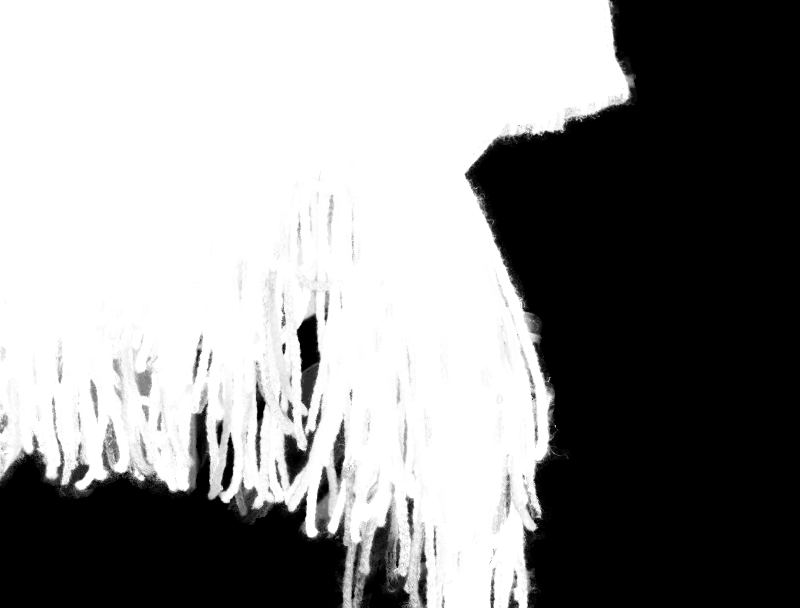} &
			\includegraphics[scale=0.12]{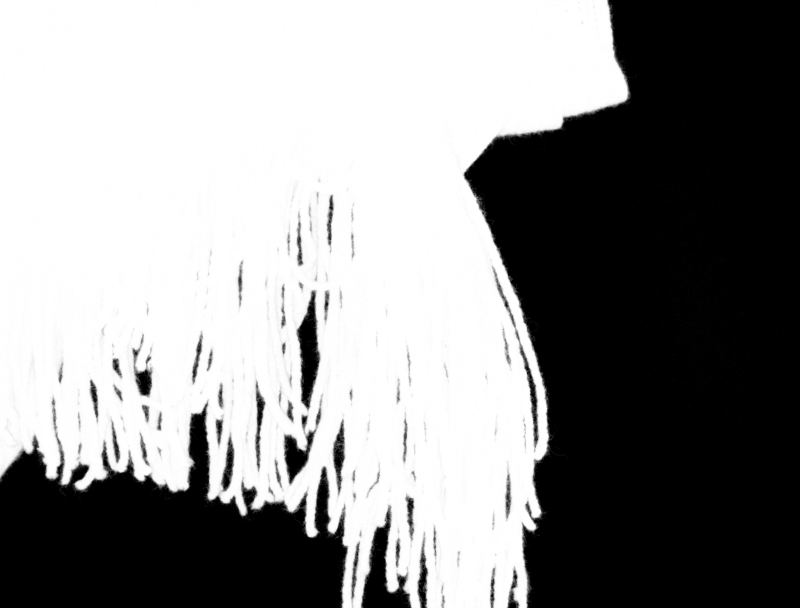} \\
			
			Input Image & Smart Scribbles & Trimap GT & Ground Truth \\
			
			\includegraphics[scale=0.12]{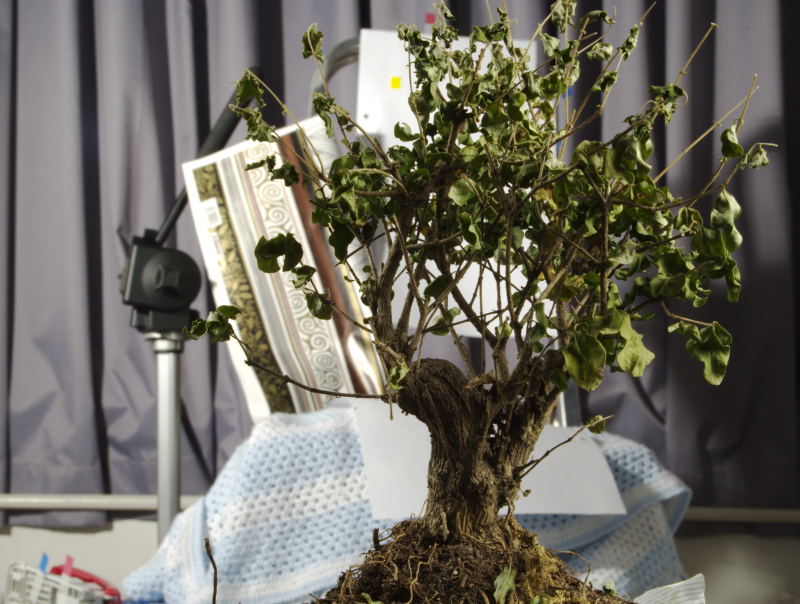} &
			\includegraphics[scale=0.16]{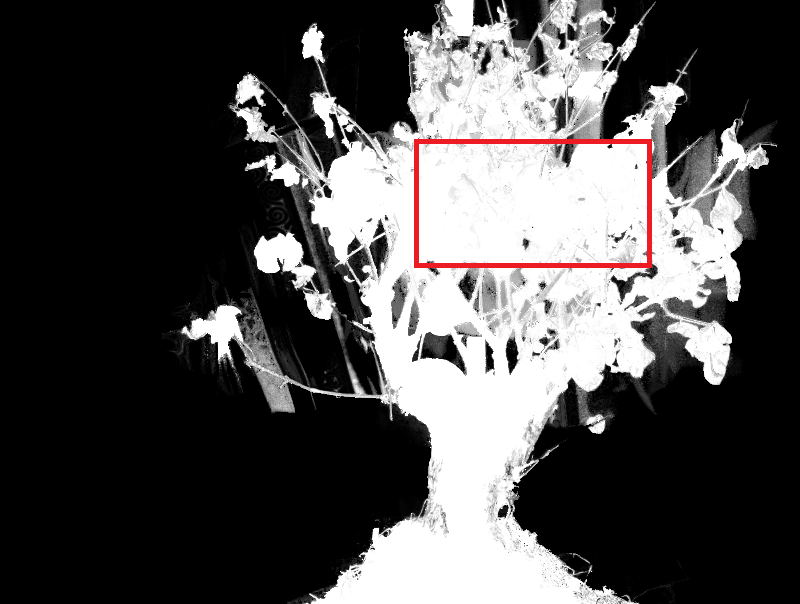} &
			\includegraphics[scale=0.12]{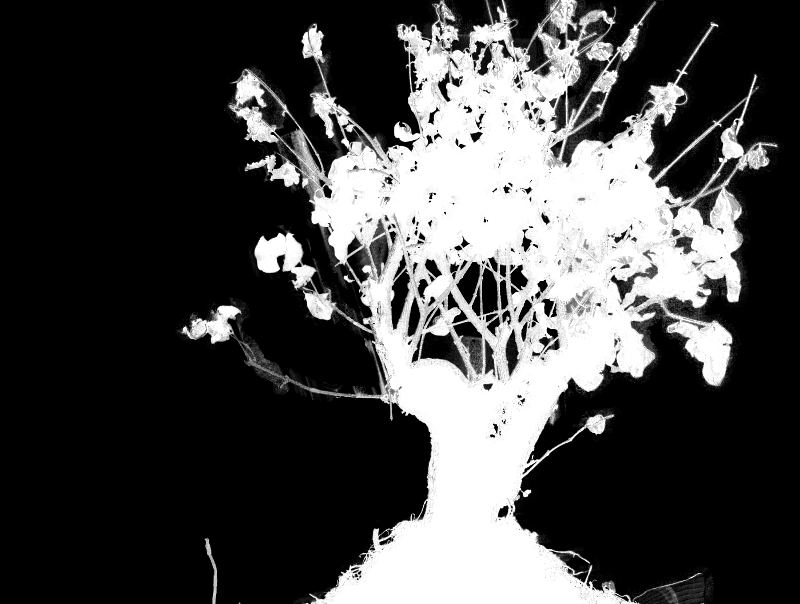} &
			\includegraphics[scale=0.12]{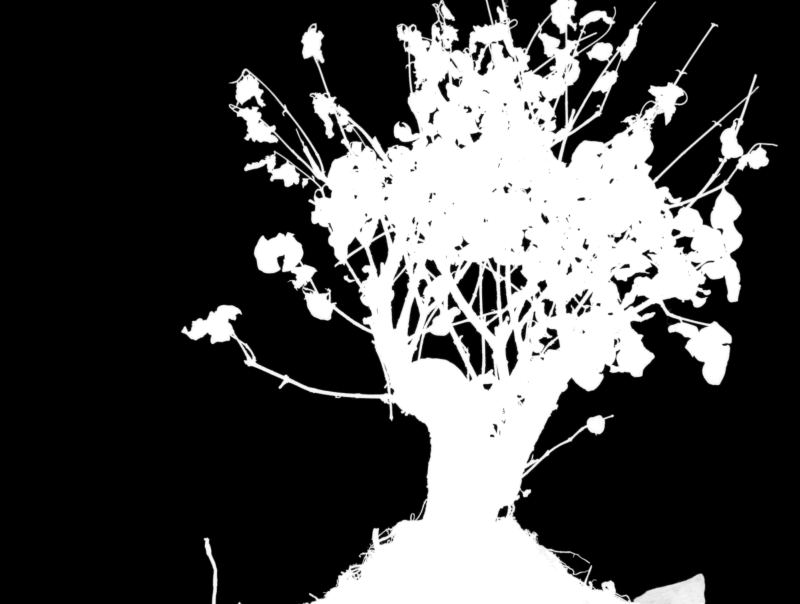} \\
			
			Input Image & Smart Scribbles & Trimap GT & Ground Truth \\
	\end{tabular}}
	\caption{The comparisons with trimap ground truths. Although \emph{smart scribbles} demonstrate complete contours and sophisticated texture details, compared to trimap GTs, there are still some local estimation errors. }
	\label{fig:comTotrigt}
\end{figure*}

\section{Conclusion and Future Work}
\label{sec:conclusion}

In this paper, we propose a new interactive framework for image matting, called \emph{smart scribbles}. We explore the principles of informative regions for matting, and an informative measurement strategy is presented for proposing regions for users labeling. It suggest informative regions for users to draw scribbles for labeling the foreground, background and unknown. A fine trimap can then be obtained by the proposed two-phase information propagation. Extensive experiments have proved the validity and universality of our framework and \emph{smart scribbles} can be applied to various matting algorithms.

Figure~\ref{fig:comTotrigt} shows a failure case of the proposed method. All alpha mattes in this example are generated using IFM~\cite{Aksoy2017Designing}. \emph{Smart scribbles} cannot handle regions where the foreground and background intersect severely, causing many details missing (Figure~\ref{fig:comTotrigt}). For future research, we aim to extend the proposed framework to a real-time editing system.

\begin{acks}
This work was supported in part by the National Natural Science Foundation of China under Grant 91748104, Grant 61972067, Grant 61632006, Grant U1811463, Grant U1908214, Grant 61751203, in part by the National Key Research and Development Program of China under Grant 2018AAA0102003, Grant 2018YFC0910506, in part by the Open Research Fund of Beijing Key Laboratory of Big Data Technology for Food Safety (Project No. BTBD-2018KF).
\end{acks}

\bibliographystyle{ACM-Reference-Format}
\bibliography{arxiv_sub}


\begin{thebibliography}{45}


\ifx \showCODEN    \undefined \def \showCODEN     #1{\unskip}     \fi
\ifx \showDOI      \undefined \def \showDOI       #1{#1}\fi
\ifx \showISBNx    \undefined \def \showISBNx     #1{\unskip}     \fi
\ifx \showISBNxiii \undefined \def \showISBNxiii  #1{\unskip}     \fi
\ifx \showISSN     \undefined \def \showISSN      #1{\unskip}     \fi
\ifx \showLCCN     \undefined \def \showLCCN      #1{\unskip}     \fi
\ifx \shownote     \undefined \def \shownote      #1{#1}          \fi
\ifx \showarticletitle \undefined \def \showarticletitle #1{#1}   \fi
\ifx \showURL      \undefined \def \showURL       {\relax}        \fi
\providecommand\bibfield[2]{#2}
\providecommand\bibinfo[2]{#2}
\providecommand\natexlab[1]{#1}
\providecommand\showeprint[2][]{arXiv:#2}

\bibitem[\protect\citeauthoryear{{Achanta}, {Shaji}, {Smith}, {Lucchi}, {Fua},
  and {Süsstrunk}}{{Achanta} et~al\mbox{.}}{2012}]%
        {achanta2012slic}
\bibfield{author}{\bibinfo{person}{R. {Achanta}}, \bibinfo{person}{A. {Shaji}},
  \bibinfo{person}{K. {Smith}}, \bibinfo{person}{A. {Lucchi}},
  \bibinfo{person}{P. {Fua}}, {and} \bibinfo{person}{S. {Süsstrunk}}.}
  \bibinfo{year}{2012}\natexlab{}.
\newblock \showarticletitle{SLIC Superpixels Compared to State-of-the-Art
  Superpixel Methods}.
\newblock \bibinfo{journal}{\emph{IEEE Transactions on Pattern Analysis and
  Machine Intelligence}} \bibinfo{volume}{34}, \bibinfo{number}{11}
  (\bibinfo{year}{2012}), \bibinfo{pages}{2274--2282}.
\newblock


\bibitem[\protect\citeauthoryear{{Aksoy}, {Aydin}, and {Pollefeys}}{{Aksoy}
  et~al\mbox{.}}{2017}]%
        {Aksoy2017Designing}
\bibfield{author}{\bibinfo{person}{Y. {Aksoy}}, \bibinfo{person}{T.~O.
  {Aydin}}, {and} \bibinfo{person}{M. {Pollefeys}}.}
  \bibinfo{year}{2017}\natexlab{}.
\newblock \showarticletitle{Designing Effective Inter-Pixel Information Flow
  for Natural Image Matting}. In \bibinfo{booktitle}{\emph{Proceedings of the
  IEEE Conference on Computer Vision and Pattern Recognition (CVPR'17)}}.
  \bibinfo{pages}{228--236}.
\newblock


\bibitem[\protect\citeauthoryear{{Badrinarayanan}, {Kendall}, and
  {Cipolla}}{{Badrinarayanan} et~al\mbox{.}}{2017}]%
        {Kendall2017}
\bibfield{author}{\bibinfo{person}{V. {Badrinarayanan}}, \bibinfo{person}{A.
  {Kendall}}, {and} \bibinfo{person}{R. {Cipolla}}.}
  \bibinfo{year}{2017}\natexlab{}.
\newblock \showarticletitle{SegNet: A Deep Convolutional Encoder-Decoder
  Architecture for Image Segmentation}.
\newblock \bibinfo{journal}{\emph{IEEE Transactions on Pattern Analysis and
  Machine Intelligence}} \bibinfo{volume}{39}, \bibinfo{number}{12}
  (\bibinfo{year}{2017}), \bibinfo{pages}{2481--2495}.
\newblock


\bibitem[\protect\citeauthoryear{{Cai}, {Zhang}, {Fan}, {Huang}, {Liu}, {Liu},
  {Liu}, {Wang}, and {Sun}}{{Cai} et~al\mbox{.}}{2019}]%
        {cai2019disentangled}
\bibfield{author}{\bibinfo{person}{S. {Cai}}, \bibinfo{person}{X. {Zhang}},
  \bibinfo{person}{H. {Fan}}, \bibinfo{person}{H. {Huang}}, \bibinfo{person}{J.
  {Liu}}, \bibinfo{person}{J. {Liu}}, \bibinfo{person}{J. {Liu}},
  \bibinfo{person}{J. {Wang}}, {and} \bibinfo{person}{J. {Sun}}.}
  \bibinfo{year}{2019}\natexlab{}.
\newblock \showarticletitle{Disentangled Image Matting}. In
  \bibinfo{booktitle}{\emph{Proceedings of the International Conference on
  Computer Vision (ICCV'19)}}. \bibinfo{pages}{8818--8827}.
\newblock


\bibitem[\protect\citeauthoryear{{Chen}, {Papandreou}, {Kokkinos}, {Murphy},
  and {Yuille}}{{Chen} et~al\mbox{.}}{2018}]%
        {Chen2014Semantic}
\bibfield{author}{\bibinfo{person}{L. {Chen}}, \bibinfo{person}{G.
  {Papandreou}}, \bibinfo{person}{I. {Kokkinos}}, \bibinfo{person}{K.
  {Murphy}}, {and} \bibinfo{person}{A.~L. {Yuille}}.}
  \bibinfo{year}{2018}\natexlab{}.
\newblock \showarticletitle{DeepLab: Semantic Image Segmentation with Deep
  Convolutional Nets, Atrous Convolution, and Fully Connected CRFs}.
\newblock \bibinfo{journal}{\emph{IEEE Transactions on Pattern Analysis and
  Machine Intelligence}} \bibinfo{volume}{40}, \bibinfo{number}{4}
  (\bibinfo{year}{2018}), \bibinfo{pages}{834--848}.
\newblock


\bibitem[\protect\citeauthoryear{Chen, Ge, Xu, Zhang, Yang, and Gai}{Chen
  et~al\mbox{.}}{2018}]%
        {Chen2018SHM}
\bibfield{author}{\bibinfo{person}{Quan Chen}, \bibinfo{person}{Tiezheng Ge},
  \bibinfo{person}{Yanyu Xu}, \bibinfo{person}{Zhiqiang Zhang},
  \bibinfo{person}{Xinxin Yang}, {and} \bibinfo{person}{Kun Gai}.}
  \bibinfo{year}{2018}\natexlab{}.
\newblock \showarticletitle{Semantic Human Matting}. In
  \bibinfo{booktitle}{\emph{Proceedings of the ACM International Conference on
  Multimedia (MM'18)}}. \bibinfo{pages}{618–626}.
\newblock


\bibitem[\protect\citeauthoryear{{Chen}, {Li}, and {Tang}}{{Chen}
  et~al\mbox{.}}{2013}]%
        {Chen2013KNN}
\bibfield{author}{\bibinfo{person}{Q. {Chen}}, \bibinfo{person}{D. {Li}}, {and}
  \bibinfo{person}{C. {Tang}}.} \bibinfo{year}{2013}\natexlab{}.
\newblock \showarticletitle{KNN Matting}.
\newblock \bibinfo{journal}{\emph{IEEE Transactions on Pattern Analysis and
  Machine Intelligence}} \bibinfo{volume}{35}, \bibinfo{number}{9}
  (\bibinfo{year}{2013}), \bibinfo{pages}{2175--2188}.
\newblock


\bibitem[\protect\citeauthoryear{Cho, Tai, and Kweon}{Cho
  et~al\mbox{.}}{2016}]%
        {Cho2016Natural}
\bibfield{author}{\bibinfo{person}{Donghyeon Cho}, \bibinfo{person}{Yu-Wing
  Tai}, {and} \bibinfo{person}{Inso Kweon}.} \bibinfo{year}{2016}\natexlab{}.
\newblock \showarticletitle{Natural Image Matting Using Deep Convolutional
  Neural Networks}. In \bibinfo{booktitle}{\emph{Proceedings of the European
  Conference on Computer Vision (ECCV'16)}}. \bibinfo{pages}{626--643}.
\newblock


\bibitem[\protect\citeauthoryear{Endo, Iizuka, Kanamori, and Mitani}{Endo
  et~al\mbox{.}}{2016}]%
        {Endo2016DeepProp}
\bibfield{author}{\bibinfo{person}{Yuki Endo}, \bibinfo{person}{Satoshi
  Iizuka}, \bibinfo{person}{Yoshihiro Kanamori}, {and} \bibinfo{person}{Jun
  Mitani}.} \bibinfo{year}{2016}\natexlab{}.
\newblock \showarticletitle{DeepProp: Extracting Deep Features from a Single
  Image for Edit Propagation}. In \bibinfo{booktitle}{\emph{Proceedings of the
  Annual Conference of the European Association for Computer Graphics
  (EG'16)}}. \bibinfo{pages}{189–201}.
\newblock


\bibitem[\protect\citeauthoryear{Feng, Liang, and Zhang}{Feng
  et~al\mbox{.}}{2016}]%
        {Feng2016A}
\bibfield{author}{\bibinfo{person}{Xiaoxue Feng}, \bibinfo{person}{Xiaohui
  Liang}, {and} \bibinfo{person}{Zili Zhang}.} \bibinfo{year}{2016}\natexlab{}.
\newblock \showarticletitle{A Cluster Sampling Method for Image Matting via
  Sparse Coding}. In \bibinfo{booktitle}{\emph{Proceedings of the European
  Conference on Computer Vision (ECCV'16)}}. \bibinfo{pages}{204--219}.
\newblock


\bibitem[\protect\citeauthoryear{Gasparini}{Gasparini}{1997}]%
        {Mauro1999Markov}
\bibfield{author}{\bibinfo{person}{Mauro Gasparini}.}
  \bibinfo{year}{1997}\natexlab{}.
\newblock \showarticletitle{Markov Chain Monte Carlo in Practice}.
\newblock \bibinfo{journal}{\emph{Technometrics}} \bibinfo{volume}{39},
  \bibinfo{number}{3} (\bibinfo{year}{1997}), \bibinfo{pages}{338--338}.
\newblock


\bibitem[\protect\citeauthoryear{Gastal and Oliveira}{Gastal and
  Oliveira}{2010}]%
        {Gastal2010Shared}
\bibfield{author}{\bibinfo{person}{Eduardo S.~L. Gastal} {and}
  \bibinfo{person}{Manuel~M. Oliveira}.} \bibinfo{year}{2010}\natexlab{}.
\newblock \showarticletitle{Shared Sampling for Real-Time Alpha Matting}.
\newblock \bibinfo{journal}{\emph{Computer Graphics Forum}}
  \bibinfo{volume}{29}, \bibinfo{number}{2} (\bibinfo{year}{2010}),
  \bibinfo{pages}{575--584}.
\newblock


\bibitem[\protect\citeauthoryear{Grady, Schiwietz, Aharon, and
  Westermann}{Grady et~al\mbox{.}}{2005}]%
        {Grady2005Random}
\bibfield{author}{\bibinfo{person}{Leo Grady}, \bibinfo{person}{Thomas
  Schiwietz}, \bibinfo{person}{Shmuel Aharon}, {and}
  \bibinfo{person}{R{\"u}diger Westermann}.} \bibinfo{year}{2005}\natexlab{}.
\newblock \showarticletitle{Random walks for interactive alpha-matting}. In
  \bibinfo{booktitle}{\emph{Proceedings of the Visualization, Imaging, and
  Image Processing (VIIP'05)}}. \bibinfo{pages}{423--429}.
\newblock


\bibitem[\protect\citeauthoryear{Guan, Chen, Liang, Ding, and Peng}{Guan
  et~al\mbox{.}}{2006}]%
        {Guan2006Easy}
\bibfield{author}{\bibinfo{person}{Yu Guan}, \bibinfo{person}{Wei Chen},
  \bibinfo{person}{Xiao Liang}, \bibinfo{person}{Zi’ang Ding}, {and}
  \bibinfo{person}{Qunsheng Peng}.} \bibinfo{year}{2006}\natexlab{}.
\newblock \showarticletitle{Easy Matting - A Stroke Based Approach for
  Continuous Image Matting}.
\newblock \bibinfo{journal}{\emph{Computer Graphics Forum}}
  \bibinfo{volume}{25}, \bibinfo{number}{3} (\bibinfo{year}{2006}),
  \bibinfo{pages}{567--576}.
\newblock


\bibitem[\protect\citeauthoryear{{Hou} and {Liu}}{{Hou} and {Liu}}{2019}]%
        {hou2019context}
\bibfield{author}{\bibinfo{person}{Q. {Hou}} {and} \bibinfo{person}{F. {Liu}}.}
  \bibinfo{year}{2019}\natexlab{}.
\newblock \showarticletitle{Context-Aware Image Matting for Simultaneous
  Foreground and Alpha Estimation}. In \bibinfo{booktitle}{\emph{Proceedings of
  the International Conference on Computer Vision (ICCV'19)}}.
  \bibinfo{pages}{4129--4138}.
\newblock


\bibitem[\protect\citeauthoryear{Karacan, Erdem, and Erdem}{Karacan
  et~al\mbox{.}}{2015}]%
        {Karacan2015Image}
\bibfield{author}{\bibinfo{person}{L. Karacan}, \bibinfo{person}{A. Erdem},
  {and} \bibinfo{person}{E. Erdem}.} \bibinfo{year}{2015}\natexlab{}.
\newblock \showarticletitle{Image Matting with KL-Divergence Based Sparse
  Sampling}. In \bibinfo{booktitle}{\emph{Proceedings of the International
  Conference on Computer Vision (ICCV'15)}}. \bibinfo{pages}{424--432}.
\newblock


\bibitem[\protect\citeauthoryear{Lee and Wu}{Lee and Wu}{2011}]%
        {Lee2011Nonlocal}
\bibfield{author}{\bibinfo{person}{P Lee} {and} \bibinfo{person}{Ying Wu}.}
  \bibinfo{year}{2011}\natexlab{}.
\newblock \showarticletitle{Nonlocal matting}. In
  \bibinfo{booktitle}{\emph{Proceedings of the IEEE Conference on Computer
  Vision and Pattern Recognition (CVPR'11)}}. \bibinfo{pages}{2193--2200}.
\newblock


\bibitem[\protect\citeauthoryear{Levin, Lischinski, and Weiss}{Levin
  et~al\mbox{.}}{2007}]%
        {Levin2007A}
\bibfield{author}{\bibinfo{person}{Anat Levin}, \bibinfo{person}{Dani
  Lischinski}, {and} \bibinfo{person}{Yair Weiss}.}
  \bibinfo{year}{2007}\natexlab{}.
\newblock \showarticletitle{A Closed-Form Solution to Natural Image Matting}.
\newblock \bibinfo{journal}{\emph{IEEE Transactions on Pattern Analysis and
  Machine Intelligence}} \bibinfo{volume}{30}, \bibinfo{number}{2}
  (\bibinfo{year}{2007}), \bibinfo{pages}{228--242}.
\newblock


\bibitem[\protect\citeauthoryear{Levin, Rav-Acha, and Lischinski}{Levin
  et~al\mbox{.}}{2008}]%
        {Levin2008Spectral}
\bibfield{author}{\bibinfo{person}{Anat Levin}, \bibinfo{person}{Alex
  Rav-Acha}, {and} \bibinfo{person}{Dani Lischinski}.}
  \bibinfo{year}{2008}\natexlab{}.
\newblock \showarticletitle{Spectral Matting}.
\newblock \bibinfo{journal}{\emph{IEEE Transactions on Pattern Analysis and
  Machine Intelligence}} \bibinfo{volume}{30}, \bibinfo{number}{10}
  (\bibinfo{year}{2008}), \bibinfo{pages}{1699--1712}.
\newblock


\bibitem[\protect\citeauthoryear{Li, Wang, Zhu, and Pi}{Li
  et~al\mbox{.}}{2017}]%
        {Li2017Three}
\bibfield{author}{\bibinfo{person}{Chao Li}, \bibinfo{person}{Ping Wang},
  \bibinfo{person}{Xiangyu Zhu}, {and} \bibinfo{person}{Huali Pi}.}
  \bibinfo{year}{2017}\natexlab{}.
\newblock \showarticletitle{Three-layer graph framework with the sumD feature
  for alpha matting}.
\newblock \bibinfo{journal}{\emph{Computer Vision and Image Understanding}}
  \bibinfo{volume}{162} (\bibinfo{year}{2017}), \bibinfo{pages}{34--45}.
\newblock


\bibitem[\protect\citeauthoryear{Long, Shelhamer, and Darrell}{Long
  et~al\mbox{.}}{2015}]%
        {Long2015Fully}
\bibfield{author}{\bibinfo{person}{Jonathan Long}, \bibinfo{person}{Evan
  Shelhamer}, {and} \bibinfo{person}{Trevor Darrell}.}
  \bibinfo{year}{2015}\natexlab{}.
\newblock \showarticletitle{Fully convolutional networks for semantic
  segmentation}. In \bibinfo{booktitle}{\emph{Proceedings of the IEEE
  Conference on Computer Vision and Pattern Recognition (CVPR'15)}}.
  \bibinfo{pages}{3431--3440}.
\newblock


\bibitem[\protect\citeauthoryear{{Lu}, {Dai}, {Shen}, and {Xu}}{{Lu}
  et~al\mbox{.}}{2019}]%
        {hao2019indexnet}
\bibfield{author}{\bibinfo{person}{H. {Lu}}, \bibinfo{person}{Y. {Dai}},
  \bibinfo{person}{C. {Shen}}, {and} \bibinfo{person}{S. {Xu}}.}
  \bibinfo{year}{2019}\natexlab{}.
\newblock \showarticletitle{Indices Matter: Learning to Index for Deep Image
  Matting}. In \bibinfo{booktitle}{\emph{Proceedings of the International
  Conference on Computer Vision (ICCV'19)}}. \bibinfo{pages}{3265--3274}.
\newblock


\bibitem[\protect\citeauthoryear{Lutz, Amplianitis, and Smolic}{Lutz
  et~al\mbox{.}}{2018}]%
        {lutz2018alphagan}
\bibfield{author}{\bibinfo{person}{Sebastian Lutz},
  \bibinfo{person}{Konstantinos Amplianitis}, {and} \bibinfo{person}{Aljoscha
  Smolic}.} \bibinfo{year}{2018}\natexlab{}.
\newblock \showarticletitle{AlphaGAN: Generative adversarial networks for
  natural image matting}. In \bibinfo{booktitle}{\emph{Proceedings of the
  British Machine Vision Conference (BMVC'18)}}. \bibinfo{pages}{259}.
\newblock


\bibitem[\protect\citeauthoryear{Qiao, Liu, Yang, Zhou, Xu, Zhang, and
  Wei}{Qiao et~al\mbox{.}}{2020}]%
        {Qiao_2020_CVPR}
\bibfield{author}{\bibinfo{person}{Yu Qiao}, \bibinfo{person}{Yuhao Liu},
  \bibinfo{person}{Xin Yang}, \bibinfo{person}{Dongsheng Zhou},
  \bibinfo{person}{Mingliang Xu}, \bibinfo{person}{Qiang Zhang}, {and}
  \bibinfo{person}{Xiaopeng Wei}.} \bibinfo{year}{2020}\natexlab{}.
\newblock \showarticletitle{Attention-Guided Hierarchical Structure Aggregation
  for Image Matting}. In \bibinfo{booktitle}{\emph{Proceedings of the IEEE
  Conference on Computer Vision and Pattern Recognition (CVPR'20)}}.
\newblock


\bibitem[\protect\citeauthoryear{Rhemann and Rother}{Rhemann and
  Rother}{2011}]%
        {Rhemann2011A}
\bibfield{author}{\bibinfo{person}{C. Rhemann} {and} \bibinfo{person}{C.
  Rother}.} \bibinfo{year}{2011}\natexlab{}.
\newblock \showarticletitle{A global sampling method for alpha matting}. In
  \bibinfo{booktitle}{\emph{Proceedings of the IEEE Conference on Computer
  Vision and Pattern Recognition (CVPR'11)}}. \bibinfo{pages}{2049--2056}.
\newblock


\bibitem[\protect\citeauthoryear{Rhemann, Rother, Wang, Gelautz, Kohli, and
  Rott}{Rhemann et~al\mbox{.}}{2009}]%
        {Rhemann2009A}
\bibfield{author}{\bibinfo{person}{C. Rhemann}, \bibinfo{person}{C. Rother},
  \bibinfo{person}{Jue Wang}, \bibinfo{person}{M. Gelautz}, \bibinfo{person}{P.
  Kohli}, {and} \bibinfo{person}{P. Rott}.} \bibinfo{year}{2009}\natexlab{}.
\newblock \showarticletitle{A perceptually motivated online benchmark for image
  matting}. In \bibinfo{booktitle}{\emph{Proceedings of the IEEE Conference on
  Computer Vision and Pattern Recognition (CVPR'09)}}.
  \bibinfo{pages}{1826--1833}.
\newblock


\bibitem[\protect\citeauthoryear{Rother, Kolmogorov, and Blake}{Rother
  et~al\mbox{.}}{2004}]%
        {Rother2004}
\bibfield{author}{\bibinfo{person}{Carsten Rother}, \bibinfo{person}{Vladimir
  Kolmogorov}, {and} \bibinfo{person}{Andrew Blake}.}
  \bibinfo{year}{2004}\natexlab{}.
\newblock \showarticletitle{“GrabCut”: Interactive Foreground Extraction
  Using Iterated Graph Cuts}.
\newblock \bibinfo{journal}{\emph{ACM Transactions on Graphics}}
  \bibinfo{volume}{23}, \bibinfo{number}{3} (\bibinfo{year}{2004}),
  \bibinfo{pages}{309–314}.
\newblock


\bibitem[\protect\citeauthoryear{Shahrian, Rajan, Price, and Cohen}{Shahrian
  et~al\mbox{.}}{2013}]%
        {Shahrian2013Improving}
\bibfield{author}{\bibinfo{person}{Ehsan Shahrian}, \bibinfo{person}{Deepu
  Rajan}, \bibinfo{person}{Brian Price}, {and} \bibinfo{person}{Scott Cohen}.}
  \bibinfo{year}{2013}\natexlab{}.
\newblock \showarticletitle{Improving Image Matting Using Comprehensive
  Sampling Sets}. In \bibinfo{booktitle}{\emph{Proceedings of the IEEE
  Conference on Computer Vision and Pattern Recognition (CVPR'13)}}.
  \bibinfo{pages}{636--643}.
\newblock


\bibitem[\protect\citeauthoryear{{Shelhamer}, {Long}, and
  {Darrell}}{{Shelhamer} et~al\mbox{.}}{2017}]%
        {Shelhamer2017}
\bibfield{author}{\bibinfo{person}{E. {Shelhamer}}, \bibinfo{person}{J.
  {Long}}, {and} \bibinfo{person}{T. {Darrell}}.}
  \bibinfo{year}{2017}\natexlab{}.
\newblock \showarticletitle{Fully Convolutional Networks for Semantic
  Segmentation}.
\newblock \bibinfo{journal}{\emph{IEEE Transactions on Pattern Analysis and
  Machine Intelligence}} \bibinfo{volume}{39}, \bibinfo{number}{4}
  (\bibinfo{year}{2017}), \bibinfo{pages}{640--651}.
\newblock


\bibitem[\protect\citeauthoryear{Shen, Tao, Gao, Zhou, and Jia}{Shen
  et~al\mbox{.}}{2016}]%
        {Shen2016Deep}
\bibfield{author}{\bibinfo{person}{Xiaoyong Shen}, \bibinfo{person}{Xin Tao},
  \bibinfo{person}{Hongyun Gao}, \bibinfo{person}{Chao Zhou}, {and}
  \bibinfo{person}{Jiaya Jia}.} \bibinfo{year}{2016}\natexlab{}.
\newblock \showarticletitle{Deep Automatic Portrait Matting}. In
  \bibinfo{booktitle}{\emph{Proceedings of the European Conference on Computer
  Vision (ECCV'16)}}. \bibinfo{pages}{92--107}.
\newblock


\bibitem[\protect\citeauthoryear{Sun, Jia, Tang, and Shum}{Sun
  et~al\mbox{.}}{2004}]%
        {Sun2004Poisson}
\bibfield{author}{\bibinfo{person}{Jian Sun}, \bibinfo{person}{Jiaya Jia},
  \bibinfo{person}{Chi~Keung Tang}, {and} \bibinfo{person}{Heung~Yeung Shum}.}
  \bibinfo{year}{2004}\natexlab{}.
\newblock \showarticletitle{Poisson matting}.
\newblock \bibinfo{journal}{\emph{ACM Transactions on Graphics}}
  \bibinfo{volume}{23}, \bibinfo{number}{3} (\bibinfo{year}{2004}),
  \bibinfo{pages}{315--321}.
\newblock


\bibitem[\protect\citeauthoryear{Sun, Lu, and Liu}{Sun et~al\mbox{.}}{2015}]%
        {Sun2015Saliency}
\bibfield{author}{\bibinfo{person}{J. Sun}, \bibinfo{person}{H. Lu}, {and}
  \bibinfo{person}{X. Liu}.} \bibinfo{year}{2015}\natexlab{}.
\newblock \showarticletitle{Saliency region detection based on Markov
  absorption probabilities}.
\newblock \bibinfo{journal}{\emph{IEEE Transactions on Image Processing}}
  \bibinfo{volume}{24}, \bibinfo{number}{5} (\bibinfo{year}{2015}),
  \bibinfo{pages}{1639--1649}.
\newblock


\bibitem[\protect\citeauthoryear{{Tang}, {Aksoy}, {Oztireli}, {Gross}, and
  {Aydin}}{{Tang} et~al\mbox{.}}{2019}]%
        {Tang_2019_CVPR}
\bibfield{author}{\bibinfo{person}{J. {Tang}}, \bibinfo{person}{Y. {Aksoy}},
  \bibinfo{person}{C. {Oztireli}}, \bibinfo{person}{M. {Gross}}, {and}
  \bibinfo{person}{T.~O. {Aydin}}.} \bibinfo{year}{2019}\natexlab{}.
\newblock \showarticletitle{Learning-Based Sampling for Natural Image Matting}.
  In \bibinfo{booktitle}{\emph{Proceedings of the IEEE Conference on Computer
  Vision and Pattern Recognition (CVPR'19)}}. \bibinfo{pages}{3050--3058}.
\newblock


\bibitem[\protect\citeauthoryear{Wang and Cohen}{Wang and Cohen}{2005}]%
        {Wang2005An}
\bibfield{author}{\bibinfo{person}{Jue Wang} {and} \bibinfo{person}{Michael~F.
  Cohen}.} \bibinfo{year}{2005}\natexlab{}.
\newblock \showarticletitle{An Iterative Optimization Approach for Unified
  Image Segmentation and Matting}. In \bibinfo{booktitle}{\emph{Proceedings of
  the International Conference on Computer Vision (ICCV'05)}}.
  \bibinfo{pages}{936--943}.
\newblock


\bibitem[\protect\citeauthoryear{Wang and Cohen}{Wang and Cohen}{2007}]%
        {Wang2007Optimized}
\bibfield{author}{\bibinfo{person}{Jue Wang} {and} \bibinfo{person}{Michael~F.
  Cohen}.} \bibinfo{year}{2007}\natexlab{}.
\newblock \showarticletitle{Optimized Color Sampling for Robust Matting}. In
  \bibinfo{booktitle}{\emph{Proceedings of the IEEE Conference on Computer
  Vision and Pattern Recognition (CVPR'07)}}. \bibinfo{pages}{1--8}.
\newblock


\bibitem[\protect\citeauthoryear{Wang, Liu, Li, Yan, and Lu}{Wang
  et~al\mbox{.}}{2016}]%
        {Wang2016}
\bibfield{author}{\bibinfo{person}{Yuhang Wang}, \bibinfo{person}{Jing Liu},
  \bibinfo{person}{Yong Li}, \bibinfo{person}{Junjie Yan}, {and}
  \bibinfo{person}{Hanqing Lu}.} \bibinfo{year}{2016}\natexlab{}.
\newblock \showarticletitle{Objectness-Aware Semantic Segmentation}. In
  \bibinfo{booktitle}{\emph{Proceedings of the ACM International Conference on
  Multimedia (MM'16)}}. \bibinfo{pages}{307–311}.
\newblock


\bibitem[\protect\citeauthoryear{Xu, Wang, Yang, He, Zhang, Yin, Wei, and
  Lau}{Xu et~al\mbox{.}}{2018}]%
        {xu2018efficient}
\bibfield{author}{\bibinfo{person}{Ke Xu}, \bibinfo{person}{Xin Wang},
  \bibinfo{person}{Xin Yang}, \bibinfo{person}{Shengfeng He},
  \bibinfo{person}{Qiang Zhang}, \bibinfo{person}{Baocai Yin},
  \bibinfo{person}{Xiaopeng Wei}, {and} \bibinfo{person}{Rynson~WH Lau}.}
  \bibinfo{year}{2018}\natexlab{}.
\newblock \showarticletitle{Efficient image super-resolution integration}.
\newblock \bibinfo{journal}{\emph{The Visual Computer}} \bibinfo{volume}{34},
  \bibinfo{number}{6-8} (\bibinfo{year}{2018}), \bibinfo{pages}{1065--1076}.
\newblock


\bibitem[\protect\citeauthoryear{Xu, Price, Cohen, and Huang}{Xu
  et~al\mbox{.}}{2017}]%
        {Xu2017Deep}
\bibfield{author}{\bibinfo{person}{Ning Xu}, \bibinfo{person}{Brian Price},
  \bibinfo{person}{Scott Cohen}, {and} \bibinfo{person}{Thomas Huang}.}
  \bibinfo{year}{2017}\natexlab{}.
\newblock \showarticletitle{Deep Image Matting}. In
  \bibinfo{booktitle}{\emph{Proceedings of the IEEE Conference on Computer
  Vision and Pattern Recognition (CVPR'17)}}. \bibinfo{pages}{311--320}.
\newblock


\bibitem[\protect\citeauthoryear{{Yang}, {Mei}, {Zhang}, {Xu}, {Yin}, {Zhang},
  and {Wei}}{{Yang} et~al\mbox{.}}{2019}]%
        {yang2019drfn}
\bibfield{author}{\bibinfo{person}{Xin {Yang}}, \bibinfo{person}{Haiyang
  {Mei}}, \bibinfo{person}{Jiqing {Zhang}}, \bibinfo{person}{Ke {Xu}},
  \bibinfo{person}{Baocai {Yin}}, \bibinfo{person}{Qiang {Zhang}}, {and}
  \bibinfo{person}{Xiaopeng {Wei}}.} \bibinfo{year}{2019}\natexlab{}.
\newblock \showarticletitle{DRFN: Deep Recurrent Fusion Network for
  Single-Image Super-Resolution With Large Factors}.
\newblock \bibinfo{journal}{\emph{IEEE Transactions on Multimedia}}
  \bibinfo{volume}{21}, \bibinfo{number}{2} (\bibinfo{year}{2019}),
  \bibinfo{pages}{328--337}.
\newblock


\bibitem[\protect\citeauthoryear{Yang, Xu, Chen, He, Yin, and Lau}{Yang
  et~al\mbox{.}}{2018}]%
        {NIPS2018_7710}
\bibfield{author}{\bibinfo{person}{Xin Yang}, \bibinfo{person}{Ke Xu},
  \bibinfo{person}{Shaozhe Chen}, \bibinfo{person}{Shengfeng He},
  \bibinfo{person}{Baocai~Yin Yin}, {and} \bibinfo{person}{Rynson Lau}.}
  \bibinfo{year}{2018}\natexlab{}.
\newblock \showarticletitle{Active Matting}.
\newblock In \bibinfo{booktitle}{\emph{Proceedings of the International
  Conference on Neural Information Processing Systems (NeurIPS'18)}}.
  \bibinfo{pages}{4590--4600}.
\newblock


\bibitem[\protect\citeauthoryear{{Yung-Yu Chuang}, {Curless}, {Salesin}, and
  {Szeliski}}{{Yung-Yu Chuang} et~al\mbox{.}}{2001}]%
        {Chuang2003A}
\bibfield{author}{\bibinfo{person}{{Yung-Yu Chuang}}, \bibinfo{person}{B.
  {Curless}}, \bibinfo{person}{D.~H. {Salesin}}, {and} \bibinfo{person}{R.
  {Szeliski}}.} \bibinfo{year}{2001}\natexlab{}.
\newblock \showarticletitle{A Bayesian approach to digital matting}. In
  \bibinfo{booktitle}{\emph{Proceedings of the IEEE Conference on Computer
  Vision and Pattern Recognition (CVPR'01)}}. \bibinfo{pages}{II--II}.
\newblock


\bibitem[\protect\citeauthoryear{Zhang, Long, Wang, Yang, Mei, and Yin}{Zhang
  et~al\mbox{.}}{2020}]%
        {zhang2020multi}
\bibfield{author}{\bibinfo{person}{Jiqing Zhang}, \bibinfo{person}{Chengjiang
  Long}, \bibinfo{person}{Yuxin Wang}, \bibinfo{person}{Xin Yang},
  \bibinfo{person}{Haiyang Mei}, {and} \bibinfo{person}{Baocai Yin}.}
  \bibinfo{year}{2020}\natexlab{}.
\newblock \showarticletitle{Multi-Context And Enhanced Reconstruction Network
  For Single Image Super Resolution}. In \bibinfo{booktitle}{\emph{Proceedings
  of the IEEE International Conference on Multimedia and Expo (ICME'20)}}.
  \bibinfo{pages}{1--6}.
\newblock


\bibitem[\protect\citeauthoryear{{Zhang}, {Gong}, {Fan}, {Ren}, {Huang}, {Bao},
  and {Xu}}{{Zhang} et~al\mbox{.}}{2019}]%
        {Zhang_2019_CVPR}
\bibfield{author}{\bibinfo{person}{Y. {Zhang}}, \bibinfo{person}{L. {Gong}},
  \bibinfo{person}{L. {Fan}}, \bibinfo{person}{P. {Ren}}, \bibinfo{person}{Q.
  {Huang}}, \bibinfo{person}{H. {Bao}}, {and} \bibinfo{person}{W. {Xu}}.}
  \bibinfo{year}{2019}\natexlab{}.
\newblock \showarticletitle{A Late Fusion CNN for Digital Matting}. In
  \bibinfo{booktitle}{\emph{Proceedings of the IEEE Conference on Computer
  Vision and Pattern Recognition (CVPR'19)}}. \bibinfo{pages}{7461--7470}.
\newblock


\bibitem[\protect\citeauthoryear{Zheng and Kambhamettu}{Zheng and
  Kambhamettu}{2009}]%
        {Zheng2009Learning}
\bibfield{author}{\bibinfo{person}{Yuanjie Zheng} {and}
  \bibinfo{person}{Chandra Kambhamettu}.} \bibinfo{year}{2009}\natexlab{}.
\newblock \showarticletitle{Learning based digital matting}. In
  \bibinfo{booktitle}{\emph{Proceedings of the International Conference on
  Computer Vision (ICCV'09)}}. \bibinfo{pages}{889--896}.
\newblock


\bibitem[\protect\citeauthoryear{Zitnick and Dollar}{Zitnick and
  Dollar}{2014}]%
        {Zitnick2014Edge}
\bibfield{author}{\bibinfo{person}{C.~Lawrence Zitnick} {and}
  \bibinfo{person}{Piotr Dollar}.} \bibinfo{year}{2014}\natexlab{}.
\newblock \showarticletitle{Edge Boxes: Locating Object Proposals from Edges}.
  In \bibinfo{booktitle}{\emph{Proceedings of the European Conference on
  Computer Vision (ECCV'14)}}. \bibinfo{pages}{391--405}.
\newblock


\end{thebibliography}

\appendix

\end{document}